\documentclass[journal]{IEEEtran}
\usepackage{cite}
\usepackage{amsmath,amssymb,amsfonts}
\usepackage{algorithmic}
\usepackage{graphicx}
\usepackage{textcomp}
\usepackage{epstopdf}
\usepackage{bm}
\usepackage{makecell}
\usepackage{multirow}
\usepackage{array}
\usepackage{booktabs}
\usepackage{color}
\usepackage{float}

\ifCLASSINFOpdf

\else

\fi

\begin{document}

\title{Deep Dual-resolution Networks for Real-time and Accurate Semantic Segmentation of Road Scenes}

\author{Yuanduo Hong, Huihui Pan, Weichao Sun,~\IEEEmembership{Senior Member,~IEEE}, Yisong Jia 
\thanks{The authors are with Research Institute of Intelligent Control and Systems,
Harbin Institute of Technology, Harbin 150001, China (e-mail: huihuipan@hit.edu.cn)}}

%
%

\markboth{Journal of \LaTeX\ Class Files,~Vol.~14, No.~8, August~2015}%
{Shell \MakeLowercase{\textit{et al.}}: Bare Demo of IEEEtran.cls for IEEE Journals}
%



\maketitle

\begin{abstract}
Semantic segmentation is a key technology for autonomous vehicles to understand the surrounding scenes. The appealing performances of contemporary models usually come at the expense of heavy computations and lengthy inference time, which is intolerable for self-driving. Using light-weight architectures (encoder-decoder or two-pathway) or reasoning on low-resolution images, recent methods realize very fast scene parsing, even running at more than 100 FPS on a single 1080Ti GPU. However, there is still a significant gap in performance between these real-time methods and the models based on dilation backbones. To tackle this problem, we proposed a family of efficient backbones specially designed for real-time semantic segmentation. The proposed deep dual-resolution networks (DDRNets) are composed of two deep branches between which multiple bilateral fusions are performed. Additionally, we design a new contextual information extractor named Deep Aggregation Pyramid Pooling Module (DAPPM) to enlarge effective receptive fields and fuse multi-scale context based on low-resolution feature maps. Our method achieves a new state-of-the-art trade-off between accuracy and speed on both Cityscapes and CamVid dataset. In particular, on a single 2080Ti GPU, DDRNet-23-slim yields 77.4$\%$ mIoU at 102 FPS on Cityscapes test set and 74.7$\%$ mIoU at 230 FPS on CamVid test set. With widely used test augmentation, our method is superior to most state-of-the-art models and requires much less computation. Codes and trained models are available online.

\end{abstract}


%
\IEEEpeerreviewmaketitle

\section{Introduction}
\begin{figure}[!t]
\centerline{\includegraphics[width=\columnwidth]{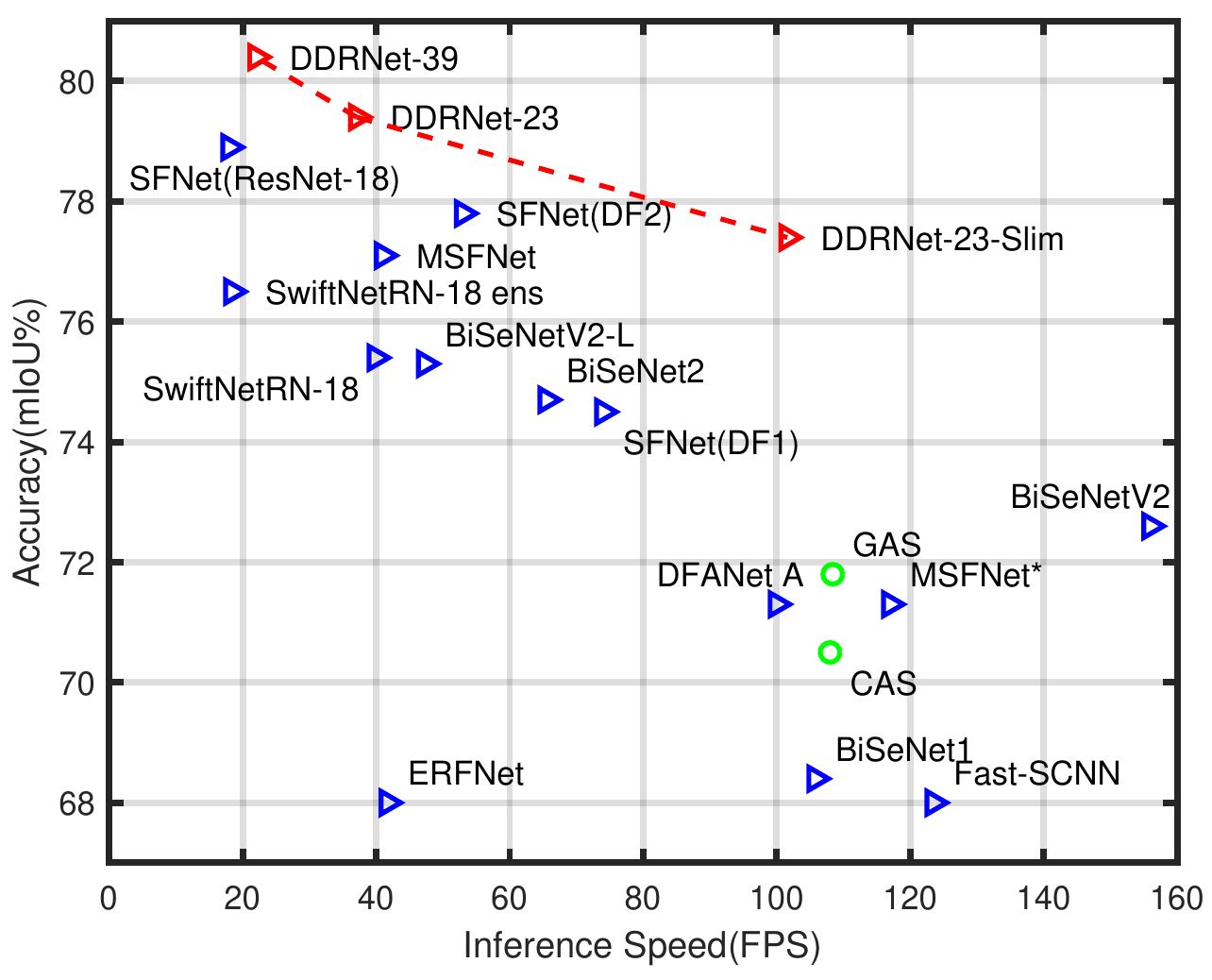}}
\caption{A comparison of speed-accuracy trade-off on Cityscapes test set. The red triangles indicate our methods while blue triangles represent other methods. Green circles represent architecture search methods.}
\label{fig4}
\end{figure}

\IEEEPARstart{S}{emantic} segmentation is a fundamental task in which each pixel of the input image should be assigned to the corresponding label\cite{8006236,8760555,9126262}. It plays a vital role in many practical applications, such as medical image segmentation, navigation of autonomous vehicles, and robots\cite{8264783,romera2017erfnet}. With the rise of deep learning technologies, convolutional neural networks are applied to image segmentation and greatly outperform traditional methods based on handcrafted features. Since the fully convolutional network (FCN)\cite{long2015fully} was proposed to handle semantic segmentation problems, a series of novel networks have been proposed. DeepLab\cite{chen2014semantic} eliminates some of downsampling in ResNet to maintain high resolution and utilizes convolutions with large dilations\cite{mallat1999wavelet} to enlarge receptive fields. From then on, dilated convolutions based backbones with context extraction modules have become the standard layout widely used in a variety of methods, including DeepLabV2\cite{chen2017deeplab}, DeepLabV3\cite{chen2017rethinking}, PSPNet\cite{zhao2017pyramid}, and DenseASPP\cite{yang2018denseaspp}.

Since semantic segmentation is a kind of dense prediction task, neural networks need to output high-resolution feature maps of large receptive fields to produce satisfactory results, which is computationally expensive. This problem is especially critical for scene parsing of autonomous driving which requires enforcement on very large images to cover a wide field of view. Therefore, the above methods are very time-consuming in the inference stage and can not be directly deployed on the actual autonomous vehicles. They can not even process an image in one second because of utilizing multi-scale test to improve accuracy.

With the ever-increasing demand for mobile device deployment, real-time segmentation algorithms\cite{paszke2016enet,9040271,zhao2018icnet,mehta2018espnet,9032321} are getting more and more attention. DFANet\cite{li2019dfanet} employs deeply multi-scale feature aggregation and lightweight depthwise separable convolutions, achieving 71.3$\%$ test mIoU with 100 FPS. Different from the encoder-decoder paradigm,
authors in \cite{yu2018bisenet} propose a novel bilateral network composed of a spatial path and a context path. Specially, the spatial path utilizes three relatively wide 3$\times$3 convolutional layers to capture spatial details, and the context path is a compact pre-trained backbone for extracting contextual information. Such bilateral methods including \cite{poudel2019fast} achieved higher inference speed than encoder-decoder structures at that time.

Recently, some competitive real-time methods aiming at semantic segmentation of road scenes are proposed. These methods can be divided into two categories. One utilizes the GPU-efficient backbones, especially ResNet-18\cite{orsic2019defense,hu2020real,li2020semantic}. The other develops complex lightweight encoders trained from scratch, one of which, BiSeNetV2\cite{yu2020bisenet} hits a new peak in terms of real-time performance, achieving 72.6$\%$ test mIoU at 156 FPS on Cityscapes. However, except for \cite{li2020semantic} using extra training data, these recent works do not show the potential for higher quality results. Some of them suffer from a lack of scalability due to deliberately designed architectures and tuned hyper-parameters. Additionally, ResNet-18 is of little advantage given the prosperity of more powerful backbones.

In this paper, we propose the dual-resolution networks with deep high-resolution representation for real-time semantic segmentation of high-resolution images, especially road-driving images. Our DDRNets start from one trunk and then divide into two parallel deep branches with different resolutions. One deep branch generates relatively high-resolution feature maps and the other extracts rich semantic information through multiple downsampling operations. Multiple bilateral connections are bridged between two branches to achieve efficient information fusion. Besides, we propose a novel module named DAPPM which inputs low-resolution feature maps, extracts multi-scale context information, and merges them in a cascaded way. Before training on semantic segmentation dataset, the dual-resolution networks are trained on ImageNet following common paradigms.

According to extensive experimental results on three popular benchmarks ($i.e.$, Cityscapes, CamVid, and COCOStuff), DDRNets attain an excellent balance between segmentation accuracy and inference speed. Our method achieves new state-of-the-art accuracy on both Cityscapes and CamVid compared with other real-time algorithms, without attention mechanism and extra bells or whistles. With standard test augmentation, DDRNet is comparable to state-of-the-art models and requires much fewer computing resources. We also report statistically relevant performances and conduct ablation experiments to analyze the effect of architecture improvements and standard training tricks.

The main contributions are summarized as follows:
\begin{itemize}
  \item A family of novel bilateral networks with deep dual-resolution branches and multiple bilateral fusions is proposed for real-time semantic segmentation as efficient backbones.
  \item A novel module is designed to harvest rich context information by combining feature aggregation with pyramid pooling. When executed on low-resolution feature maps, it leads to little increase in inference time.
  \item Our method achieves a new state-of-the-art trade-off between accuracy and speed with the 2080Ti, 77.4$\%$ mIoU at 102 FPS on Cityscapes test set and 74.7$\%$ mIoU at 230 FPS on CamVid test set. To our best knowledge, we are the first to achieve 80.4$\%$ mIoU in nearly real time (22 FPS) on Cityscapes only using fine annotations.

\end{itemize}
\section{Related Work}

In recent years, dilation convolutions based methods have boosted the performance of semantic segmentation under many challenging scenes. And pioneering works explore more possibilities for lightweight architectures such as the encoder-decoder and the two-pathway. In addition, contextual information is proved to be very crucial for scene parsing tasks. In this section, we group the related works into three categories, \emph{i.e.}, high-performance semantic segmentation, real-time semantic segmentation, and context extraction modules.

\begin{figure*}[!t]
\centerline{\includegraphics[width=\textwidth]{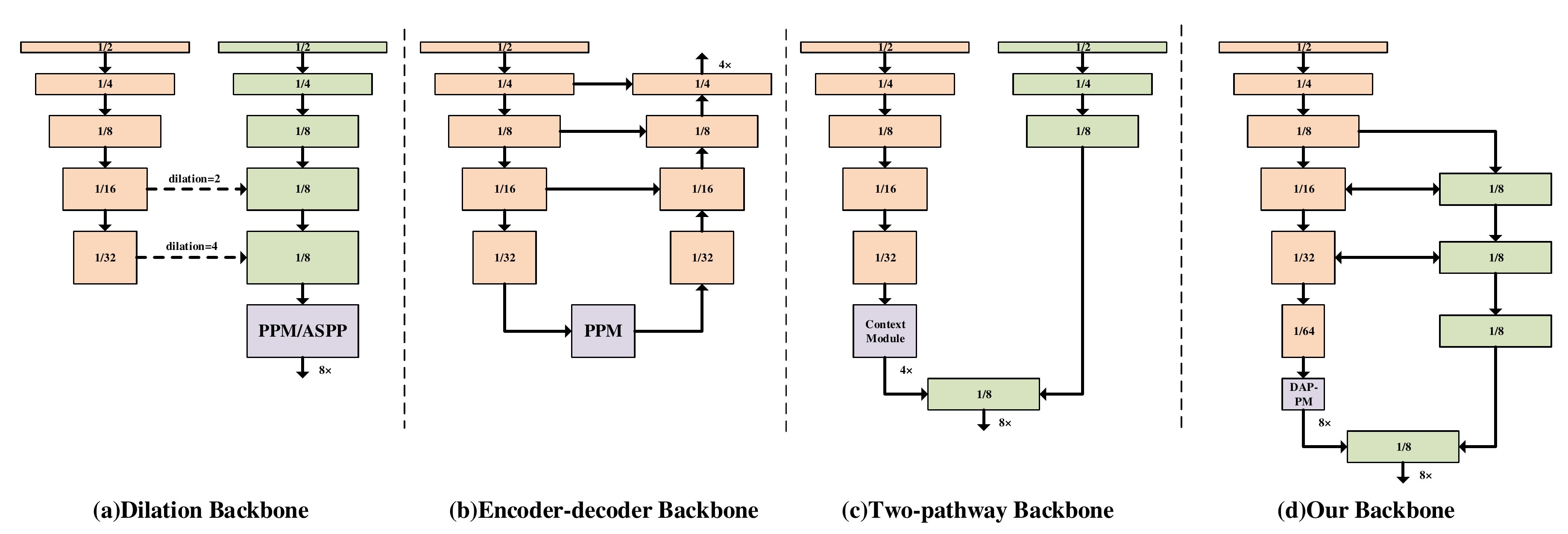}}
\caption{A comparison about dilation methods, encoder-decoder methods, two-pathway methods and our deep dual-resolution network.}
\label{fig6}
\end{figure*}

\subsection{High-performance Semantic Segmentation}

The output of the last layer of a common encoder can not be used directly to predict the segmentation masks due to the lack of spatial details. And effective receptive fields will be too small to learn high-level semantic information if only getting rid of downsampling of  the classification backbones. An acceptable strategy is to utilize dilated convolutions to set up the long-range connection between pixels while removing the last two downsampling layers\cite{chen2017rethinking,zhao2017pyramid}, as shown in Fig. \ref{fig6} (a). However, it also poses new challenges to real-time inference due to the exponential growth of high-resolution feature-map dimensions and inadequate optimization of dilated convolution implementation. There is a fact that most state-of-the-art models are built on dilation backbones and are therefore largely unqualified for scene parsing of self-driving.

Some works attempt to explore the substitute of the standard dilation backbones. Authors of DeepLabv3plus\cite{chen2018encoder} propose a simple decoder that fuses upsampled feature maps with low-level feature maps. It alleviates the requirement for high-resolution feature maps generated directly from dilated convolutions. DeepLabv3plus can achieve competitive results though the output stride of the encoder is set to 16. HRNet\cite{sun2019high} highlights the deep high-resolution representations and reflects higher efficiency than dilation backbones. We find that the higher computational efficiency and inference speed of HRNet owe to its much thinner high-resolution information flows. Taking HRNetV2-W48 as an example, the dimensions of 1/4 resolution and 1/8 resolution features are 48 and 96, respectively, which are much smaller than those of pre-trained ResNets\cite{he2016deep} with dilation convolutions. Although high-resolution branches of HRNet are much thinner, they can be greatly enhanced by parallel low-resolution branches and repeated multi-scale fusions.

Our work begins with the deep, thin, high-resolution representations and puts forward more compact architectures. They maintain high-resolution representations and extract high-level contextual information simultaneously through two concise trunks.

\subsection{Real-time Semantic Segmentation}
Almost all real-time semantic segmentation models employ two basic methods: encoder-decoder methods and two-pathway methods. Lightweight encoders which play an significant role in both methods are also discussed.
\subsubsection{Encoder-decoder Architecture}
Compared to dilated convolution based models, encoder-decoder architectures intuitively cost less computation and inference time.
The encoder is usually a deep network with repeated spatial reductions to extract contextual information and the decoder restores the resolution by interpolation or transposed convolution\cite{zeiler2010deconvolutional} to complete dense predictions, as shown in Fig. \ref{fig6} (b). Specially, the encoder can be a lightweight backbone pre-trained on ImageNet or an efficient variant trained from scratch like ERFNet\cite{romera2017erfnet} and ESPNet\cite{mehta2018espnet}. SwiftNet\cite{orsic2019defense} defends the advantage of pre-training encoders on ImageNet and leverages lightweight lateral connections to assist with upsampling. Authors in \cite{si2019real} propose a strategy of multiply spatial fusion and class boundary supervision. FANet \cite{hu2020real} achieves a good trade-off between speed and accuracy with fast attention modules and extra downsampling throughout the network. SFNet\cite{li2020semantic} delivers a Flow Alignment Module (FAM) to align feature maps of adjacent levels for better fusion.

\subsubsection{Two-pathway Architecture}

The encoder-decoder architecture reduces computational effort, but due to the loss of some information during repeated downsampling, it can not be completely recovered by unsampling, which impairs the accuracy of semantic segmentation. The two-pathway architecture is proposed in order to alleviate this problem\cite{yu2018bisenet}, as shown in Fig. \ref{fig6} (c). In addition to one pathway for extracting semantic information, the other shallow pathway of high resolution provides rich spatial details as a supplement. To further improve the accuracy, BiSeNetV2\cite{yu2020bisenet} uses global average pooling for context embedding and proposes attention based feature fusion. The two pathways in BiSeNetV1$\&$V2 are initially separate while the two branches in Fast-SCNN\cite{poudel2019fast} share the learning to downsample module. CABiNet\cite{kumaar2020cabinet} adopts the overall architecture of Fast-SCNN but uses the MobileNetV3\cite{howard2019searching} as the context branch. 

Other than existing two-pathway methods, the deep and thin high-resolution branch of DDRNets enables multiple feature fusions and sufficient ImageNet pre-training while guaranteeing the inference efficiency. Our method can be easily scaled to achieve higher accuracy (above 80$\%$ mIoU on Cityscapes). 

\subsubsection{Lightweight Encoders}

There are many computationally efficient backbones can be used as the encoder, such as MobileNet\cite{howard2017mobilenets}, ShuffleNet\cite{zhang2018shufflenet} and small version of Xception\cite{chollet2017xception}. MobileNet replaces standard convolutions with depthwise separable convolutions to reduce parameters and computation. The strong regularization effect of depthwise separable convolutions is alleviated by inverted residual blocks in MobileNetV2\cite{sandler2018mobilenetv2}. ShuffleNet utilizes the compactness of grouped convolutions and proposes a channel shuffle operation to facilitate information fusion between different groups. However, these networks contain numerous depthwise separable convolutions which can not be efficiently implemented with the existing GPU architecture. Therefore, although the FLOPs of ResNet-18\cite{he2016deep} is about six times of MobileNetV2 1.0$\times$, inference speed of the former is higher than the latter on single 1080Ti GPU\cite{orsic2019defense}. However, the existing lightweight backbones may be suboptimal for semantic segmentation because they are usually overly tuned for image classification.


\subsection{Context Extraction Modules}

Another key to semantic segmentation is how to capture richer contextual information. Atrous Spatial Pyramid Pooling (ASPP)\cite{chen2017deeplab} consists of parallel atrous convolutional layers with different rates which can attend to multi-scale contextual information. Pyramid Pooling Module (PPM)\cite{zhao2017pyramid} in PSPNet is more computationally efficient than ASPP by implementing pyramid pooling ahead of convolutional layers. Unlike the local nature of convolutional kernels, self-attention mechanism is good at capturing global dependencies. In this way, Dual Attention Network (DANet)\cite{fu2019dual} takes advantage of both position attention and channel attention to further improve feature representation. Object Context Network (OCNet)\cite{yuan2018ocnet} utilizes self-attention mechanism to explore object context which is defined as a set of pixels belonging to the same object category. Authors of CCNet\cite{huang2019ccnet} raise criss-cross attention to improve the efficiency of memory usage and computation. However, these context extraction modules are designed and performed for high-resolution feature maps, too time consuming for lightweight models. Taking the low-resolution feature maps as input, we strengthen the PPM module with more scales and deep feature aggregation. When appended to the end of the low-resolution branch, the proposed module outperforms the PPM and Base-OC module in OCNet.

\section{Method}

\begin{table*}[]
\caption{The architectures of DDRNet-23-slim and DDRNet-39 for ImageNet. `conv4$\times r$' denotes that conv4 is repeated $r$ times. For DDRNet-23-slim, $r = 1$ and for DDRNet-39, $r = 2$.}
\label{tab:3}
\begin{tabular}{p{40pt}<{\centering}|p{60pt}<{\centering}|p{85pt}<{\centering}|p{85pt}<{\centering}|p{85pt}<{\centering}|p{85pt}<{\centering}}
\toprule
stage           &output  & \multicolumn{2}{c|}{DDRNet-23-slim}& \multicolumn{2}{c}{DDRNet-39}      \\ \hline
conv1           &$112\times112$&\multicolumn{2}{c|}{$3\times3, 32$, stride 2}& \multicolumn{2}{c}{$3\times3, 64$, stride 2}         \\ \hline
\multirow{3}{*}{conv2}&\multirow{3}{*}{$56\times56$} & \multicolumn{2}{c|}{$3\times3, 32$, stride 2}& \multicolumn{2}{c}{$3\times3, 64$, stride 2}         \\ \cline{3-6}
                    &   & \multicolumn{2}{c|}{$\left[\begin{aligned}&3 \times 3, 32\\&3 \times 3, 32\end{aligned}\right] \times 2$}
                        & \multicolumn{2}{c}{$\left[\begin{aligned}&3 \times 3, 64\\&3 \times 3, 64\end{aligned}\right] \times 3$}\\ \hline
conv3           &$28\times28$       & \multicolumn{2}{c|}{$\left[\begin{aligned}&3 \times 3, 64\\&3 \times 3, 64\end{aligned}\right] \times 2$}& \multicolumn{2}{c}{$\left[\begin{aligned}&3 \times 3, 128\\&3 \times 3, 128\end{aligned}\right] \times 4$}\\\hline
\multirow{2}{*}{conv4$\times r$} & \multirow{2}{*}{$14\times14, 28\times28$}& $\left[\begin{aligned}&3 \times 3, 128\\&3 \times 3, 128\end{aligned}\right] \times 2$                  &$\left[\begin{aligned}&3 \times 3, 64\\&3 \times 3, 64\end{aligned}\right] \times 2$ & $\left[\begin{aligned}&3 \times 3, 256\\&3 \times 3, 256\end{aligned}\right] \times 3$                  &$\left[\begin{aligned}&3 \times 3, 128\\&3 \times 3, 128\end{aligned}\right] \times 3$                  \\ \cline{3-6}
                       & & \multicolumn{2}{c|}{Bilateral fusion} & \multicolumn{2}{c}{Bilateral fusion}                                    \\ \hline
\multirow{6}{*}{conv5\_1} & \multirow{6}{*}{$7\times7, 28\times28$}& $\left[\begin{aligned}&3 \times 3, 256\\&3 \times 3, 256\end{aligned}\right] \times 2$                  &$\left[\begin{aligned}&3 \times 3, 64\\&3 \times 3, 64\end{aligned}\right] \times 2$ & $\left[\begin{aligned}&3 \times 3, 512\\&3 \times 3, 512\end{aligned}\right] \times 3$                  &$\left[\begin{aligned}&3 \times 3, 128\\&3 \times 3, 128\end{aligned}\right] \times 3$                  \\ \cline{3-6}
                       & & \multicolumn{2}{c|}{Bilateral fusion} & \multicolumn{2}{c}{Bilateral fusion}                                      \\ \cline{3-6}
                       & & $\left[\begin{aligned}&1 \times 1, 256\\&3 \times 3, 256\\&1 \times 1, 512\end{aligned}\right] \times 1$                  & $\left[\begin{aligned}&1 \times 1, 64\\&3 \times 3, 64\\&1 \times 1, 128\end{aligned}\right] \times 1$ & $\left[\begin{aligned}&1 \times 1, 512\\&3 \times 3, 512\\&1 \times 1, 1024\end{aligned}\right] \times 1$                  &$\left[\begin{aligned}&1 \times 1, 128\\&3 \times 3, 128\\&1 \times 1, 256\end{aligned}\right] \times 1$                  \\ \hline
\multirow{2}{*}{conv5\_2}& \multirow{2}{*}{$7\times7$}& \multicolumn{2}{c|}{High-to-low fusion} & \multicolumn{2}{c}{High-to-low fusion}                                      \\ \cline{3-6}
                       & & \multicolumn{2}{c|}{$1\times1, 1024$} & \multicolumn{2}{c}{$1\times1, 2048$}                 \\ \hline
                       &\multirow{2}{*}{$1\times1$} & \multicolumn{2}{c|}{$7\times7$ global average pool} & \multicolumn{2}{c}{$7\times7$ global average pool}                                      \\ \cline{3-6}
                       & & \multicolumn{2}{c|}{1000-d fc, softmax} & \multicolumn{2}{c}{1000-d fc, softmax}                 \\ \bottomrule
\end{tabular}
\end{table*}

In this section, the whole pipeline is described, which consists of two main components: the Deep Dual-resolution Network and the Deep Aggregation Pyramid Pooling Module.

\subsection{Deep Dual-resolution Network}

For convenience, we can add an additional high-resolution branch to the widely used classification backbone such as ResNets. To achieve a trade-off between resolution and inference speed, we let the high-resolution branch create feature maps whose resolution is 1/8 of the input image resolution. Therefore, the high-resolution branch is appended to the end of the conv3 stage. Note that the high-resolution branch does not contain any downsampling operations and has a one-to-one correspondence with the low-resolution branch to form deep high-resolution representations. Then multiple bilateral feature fusions can be performed at different stages to fully fuse the spatial information and semantic information.

\begin{figure}[H]
\centerline{\includegraphics[width=\columnwidth]{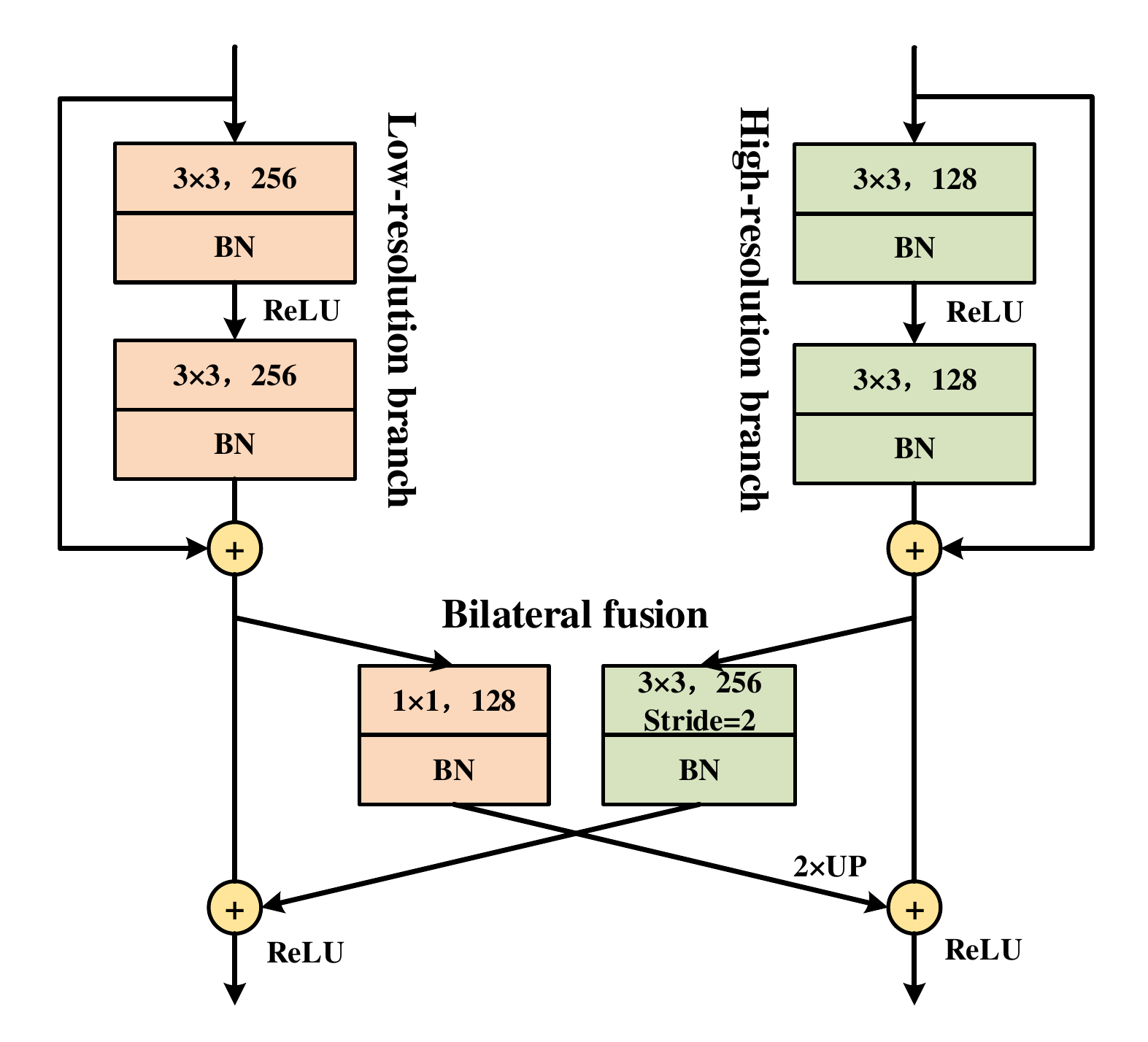}}
\caption{The details of bilateral fusion in DDRNet. Point-wise summation is implemented before ReLU.}
\label{fig5}
\end{figure}

The detailed architectures of DDRNet-23-slim and DDRNet-39 are shown in Table \ref{tab:3}. We modify the input stem of the original ResNet, replacing one 7$\times$7 convolutional layer with two sequential 3$\times$3 convolutional layers. Residual basic blocks are utilized to construct the trunk and the subsequent two branches. To expand the output dimension, one bottleneck block is added at the end of each branch. The bilateral fusion includes fusing the high-resolution branch into the low-resolution branch (high-to-low fusion) and fusing the low-resolution into the high-resolution branch (low-to-high fusion). For high-to-low fusion, high-resolution feature maps are downsampled by a sequence of 3$\times$3 convolutions with a stride of 2 prior to pointwise summation. For low-to-high resolution, low-resolution feature maps are first compressed by a 1$\times$1 convolution and then upsampled with bilinear interpolation. Fig. \ref{fig5} shows how bilateral fusion is implemented. The $i$-th high-resolution feature maps $X_{Hi}$ and low-resolution feature maps $X_{Li}$ can be written as:
\begin{equation}
\left\{
\begin{aligned}
&X_{Hi} = R(F_H(X_{H(i-1)}) + T_{L-H}(F_L(X_{L(i-1)})))  \\
&X_{Li} = R(F_L(X_{L(i-1)}) + T_{H-L}(F_H(X_{H(i-1)})))
\end{aligned}
\right.
\label{eq1}
\end{equation}
where $F_H$ and $F_L$ correspond to the sequence of residual basic blocks with high resolution and low resolution, $T_{L-H}$ and $T_{H-L}$ refer to the low-to-high and high-to-low transformer, $R$ denotes the ReLU function.

We totally construct four dual-resolution networks of different depths and widths. DDRNet-23 is twice as wide as DDRNet-23-slim and DDRNet-39 1.5$\times$ is also a wider version of DDRNet-39.

\begin{figure*}
\centerline{\includegraphics[width=\textwidth]{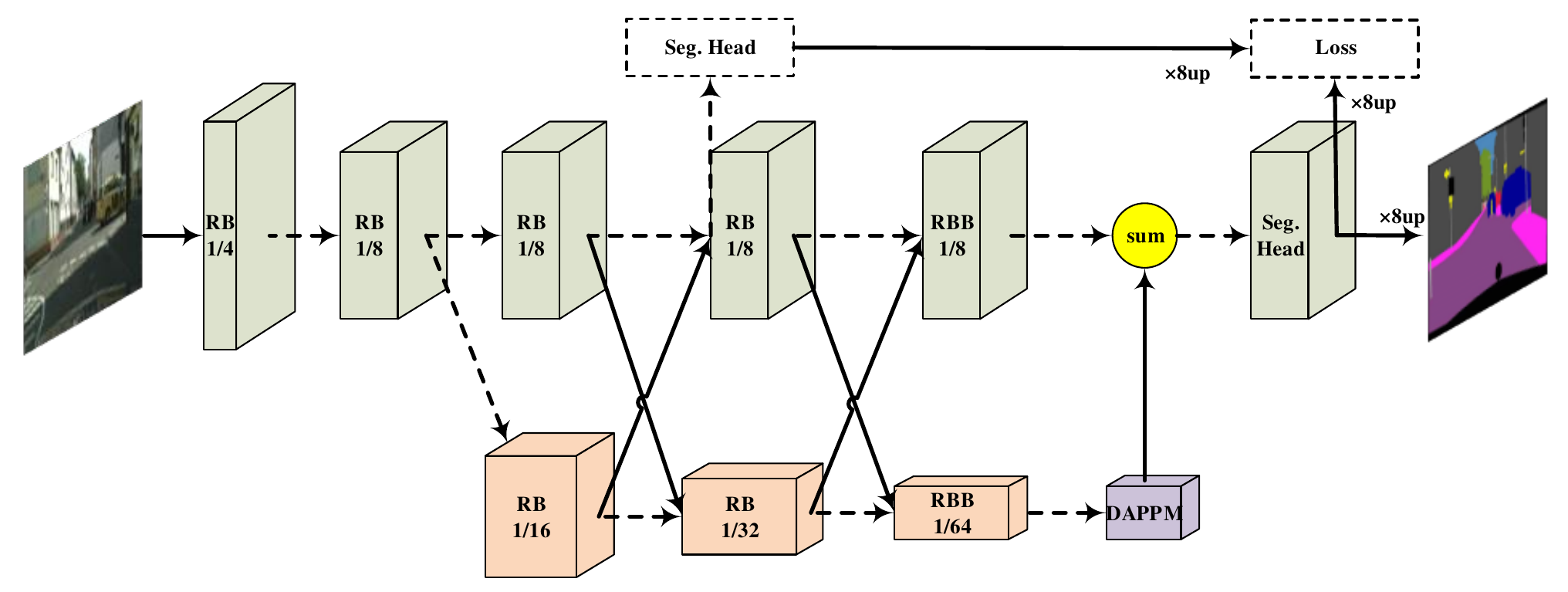}}
\caption{The overview of DDRNets on semantic segmentation. ``RB'' denotes sequential residual basic blocks. ``RBB'' denotes the single residual bottleneck block. ``DAPPM'' denotes the Deep Aggregation Pyramid Pooling Module. ``Seg. Head'' denotes the segmentation head. Black solid lines denote information paths with data processing (including upsampling and downsampling) and black dashed lines denote information paths without data processing. ``sum'' denotes the pointwise summation. Dashed boxes denote the components which are discarded in the inference stage.}
\label{fig1}
\end{figure*}

\subsection{Deep Aggregation Pyramid Pooling Module}
\label{deep}

\begin{figure}[!t]
\centerline{\includegraphics[width=\columnwidth]{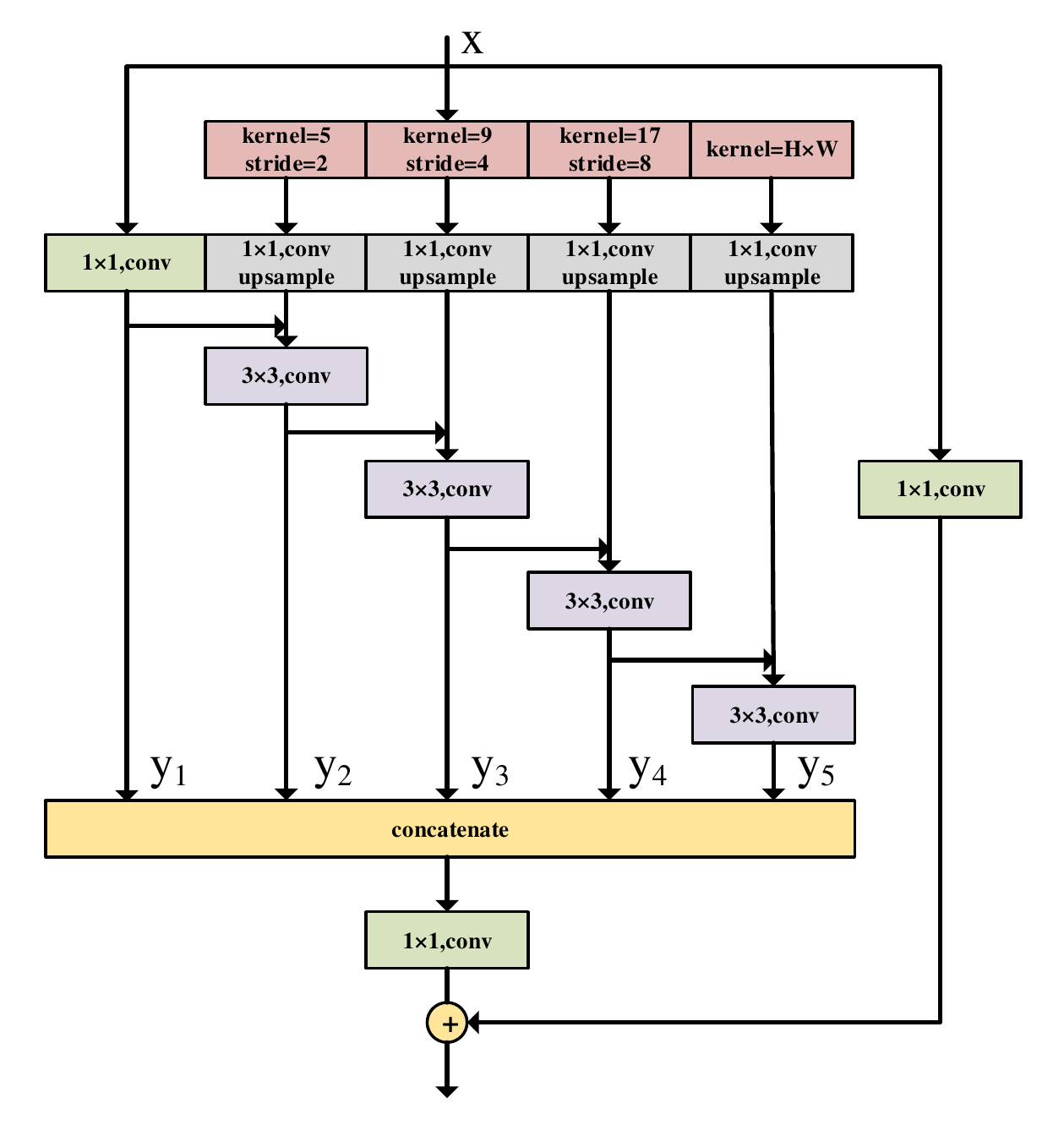}}
\caption{The detailed architecture of Deep Aggregation Pyramid Pooling Module. The number of multi-scale branches can be adjusted according to input resolution.}
\label{fig2}
\end{figure}

Here, we propose a novel module to further extract contextual information from low-resolution feature maps. Fig. \ref{fig2} shows the interior structure of the DAPPM. Taking feature maps of 1/64 image resolution as input, large pooling kernels with exponential strides are performed to generate feature maps of 1/128, 1/256, 1/512 image resolution. Input feature maps and image-level information generated by global average pooling are also utilized. We argue that it is inadequate to blend all the multi-scale contextual information by a single 3$\times$3 or 1$\times$1 convolution. Inspired by Res2Net\cite{gao2019res2net}, we first upsample the feature maps and then uses more 3$\times$3 convolutions to fuse contextual information of different scales in a hierarchial-residual way. Considering an input $x$, each scale $y_i$ can be written as:
\begin{equation}
y_i=\left\{
\begin{aligned}
&C_{1\times1}(x), & i = 1 ; \\
&C_{3\times3}(U(C_{1\times1}(P_{2^i+1,2^{i-1}}(x)))+y_{i-1}), &1<i<n ; \\
&C_{3\times3}(U(C_{1\times1}(P_{global}(x)))+y_{i-1}), &i=n .
\end{aligned}
\right.
\label{eq2}
\end{equation}
where $C_{1\times1}$ is $1\times1$ convolution, $C_{3\times3}$ is $3\times3$ convolution, $U$ denotes upsampling operation, $P_{j,k}$ denotes the pool layer of which kernel size is $j$ and stride is $k$, $P_{global}$ denotes the global average pooling.
In the end, all feature maps are concatenated and compressed using a 1$\times$1 convolution. Besides, a 1$\times1$ projection shortcut is added for easy optimization. Similar to SPP in SwiftNet\cite{orsic2019defense}, DAPPM is implemented with the sequence BN-ReLU-Conv.

Inside a DAPPM, contexts extracted by larger pooling kernels are integrated with deeper information flow, and multi-scale nature is formed by integrating different depths with different sizes of pooling kernels. Table \ref{tab:6} shows that DAPPM is able to provide much richer context than PPM. Although DAPPM contains more convolution layers and more complex fusion strategy, it hardly affects the inference speed because the input resolution is only 1/64 of the image resolution. For example, with a 1024$\times$1024 image, the maximum resolution of feature maps is 16$\times$16.

\begin{table}[]
\caption{Considering an input image of 1024$\times$1024, the generated context sizes of PPM and DAPPM are listed}
\label{tab:6}
\begin{tabular}{p{50pt}p{82pt}<{\centering}p{82pt}<{\centering}}
\toprule
                              & PPM                               & DAPPM  \\ \midrule
\multirow{5}{*}{Output scale} & \multirow{5}{*}{[16, 6, 3, 2, 1]} & [16] \\
                              &                                   & [16, 8] \\
                              &                                   & [16, 8, 4] \\
                              &                                   & [16, 8, 4, 2] \\
                              &                                   & [16, 8, 4, 2, 1] \\ \bottomrule
\end{tabular}
\end{table}

\subsection{Overall Architecture for Semantic Segmentation}

An overview of our method is depicted in Fig. \ref{fig1}. Some changes are made to the dual-resolution network to accommodate the semantic segmentation task. First, the stride of 3$\times$3 convolution in the RBB of the low-resolution branch is set to 2 to further downsample. Then, a DAPPM is added to the output of the low-resolution branch, which extracts rich contextual information from the high-level feature maps of 1/64 image resolution. Besides, the last high-to-low fusion is replaced by low-to-high fusion implemented by bilinear interpolation and summation fusion. At last, we devise a simple segmentation head consisting of a 3$\times$3 convolutional layer followed by a 1$\times$1 convolutional layer. The computational load of the segmentation head can be adjusted by changing the output dimension of the 3$\times$3 convolutional layer. We set it to 64 for DDRNet-23-slim, 128 for DDRNet-23, and 256 for DDRNet-39. Note that except for the segmentation head and the DAPPM module, all the modules have been pre-trained on ImageNet.

\subsection{Deep Supervision}

Extra supervision during the training stage can ease the optimization of deep convolutional neural networks (DCNNs). In PSPNet, an auxiliary loss is added to supervise the output of res4\_22 block of ResNet-101 and the corresponding weight is set to 0.4 according to experimental results\cite{zhao2017pyramid}. BiSeNetV2\cite{yu2020bisenet} proposes a booster training strategy in which extra segmentation heads are added at the end of each stage of the semantic branch. However, it needs numerous experiments to find the optimal weights to balance each loss, and leads to a non-negligible increase in training memory. To acquire better results, SFNet\cite{li2020semantic} utilizes a similar strategy named Cascaded Deeply Supervised Learning. In this paper, we only adopt simple extra supervision for a fair comparison with most of the methods. We add the auxiliary loss as shown in Fig. \ref{fig1} and set the weight to 0.4 following the PSPNet. The auxiliary segmentation head is discarded during the testing stage. The final loss which is the weighted sum of cross-entropy loss can be expressed as:
\begin{equation}
L_f = L_n + \alpha L_a
\label{eq3}
\end{equation}
where $L_f$, $L_n$, $L_a$ represents the final loss, normal loss, auxiliary loss respectively and $\alpha$ denotes the weight of auxiliary loss, which is 0.4 in this paper.

\section{Experiments}

\subsection{Datasets}

Cityscapes\cite{cordts2016cityscapes} is one of the well-known datasets focusing on urban street scene parsing. It contains 2975 finely annotated images for training, 500 images for validation, and 1525 images for testing. We do not use extra 20000 coarsely labeled images during training. There is a total of 19 classes available for the semantic segmentation task. The resolution of images is 2048$\times$1024, which is challenging for real-time semantic segmentation.

CamVid\cite{brostow2009semantic} consists of 701 densely annotated frames and the resolution of each frame is 960$\times$720. It comprises 367 images for training, 101 images for validation, and 233 images for testing. We merge the train set and validation set for training and evaluate our models on the test set using 11 classes following previous works\cite{yu2018bisenet,orsic2019defense,li2019dfanet}.

COCOStuff\cite{caesar2018coco} provides 10$K$ complex images with dense annotations of 182 categories, including 91 thing and 91 stuff classes. Note that 11 of the thing classes do not have any segmentation annotations. We follow the split in \cite{caesar2018coco} (9$K$ for training and 1$K$ for testing) for a fair comparison.

\subsection{Train Setting}

Before finetuning on semantic segmentation tasks, the dual-resolution networks are trained on ImageNet\cite{russakovsky2015imagenet} following the same data augmentation strategy as previous works\cite{he2016deep},\cite{xie2017aggregated}. All the models are trained with an input resolution of 224$\times$224, a batch size of 256, and 100 epochs on four 2080Ti GPUs. The initial learning rate is set to 0.1 and is reduced by 10 times at epoch 30, 60, and 90. We train all the networks using SGD with a weight decay of 0.0001 and a Nesterov momentum of 0.9. Top-1 errors on ImageNet validation set are shown in Table \ref{tab:5}. Although the efficiency of DDRNet is not superior to many advanced lightweight backbones which are elaborately designed on ImageNet, it still achieves start-of-the-art results on semantic segmentation benchmarks considering a speed trade-off. The training settings of Cityscapes, CamVid, and COCOStuff are introduced as follows.

\begin{table}[]
\caption{Top-1 error rates, parameter size and GFLOPs of four scaled-up DDRNets}
\label{tab:5}
\begin{tabular}{p{70pt}p{33pt}<{\centering}p{55pt}<{\centering}p{45pt}<{\centering}}
\toprule
Model                    & top-1 err.           & Params.              & GFLOPs        \\ \midrule
DDRNet-23-slim           & 29.8                 & 7.57M                & 0.98G               \\ \midrule
DDRNet-23                & 24.1                 & 28.22M               & 3.88G               \\ \midrule
DDRNet-39                & 22.7                 & 40.13M               & 6.95G               \\ \midrule
DDRNet-39 1.5$\times$    & 21.6                 & 76.86M               & 14.85G              \\ \bottomrule
\end{tabular}
\end{table}

\begin{table*}[]
\caption{Accuracy and speed comparison on Cityscapes. We report results on both val set and test set. Since inference speed of different models is measured under different conditions, the corresponding GPU models and input resolutions are reported. Our GFLOPs calculation adopts 2048$\times$1024 image as input. The corresponding speed is measured using TensorRT acceleration if the method is marked with \dag}
\label{tab:1}
\begin{tabular}{p{80pt}p{43pt}<{\centering}p{43pt}<{\centering}p{50pt}<{\centering}p{50pt}<{\centering}p{50pt}<{\centering}p{50pt}<{\centering}p{50pt}<{\centering}}
\toprule
\multirow{2}{*}{Model} & \multicolumn{2}{c}{MIoU} & \multirow{2}{*}{Speed (FPS)} & \multirow{2}{*}{GPU} & \multirow{2}{*}{Resolution} & \multirow{2}{*}{GFLOPs} & \multirow{2}{*}{Params} \\ \cmidrule{2-3}
                       & val         & test       &                      &                      &                             &                         &                         \\ \midrule
SegNet\cite{badrinarayanan2017segnet}& -           & 57         & 16.7                 & TitanX               & 640$\times$360              & 286                        & 29.5M                   \\
ENet\cite{paszke2016enet}  & -           & 57         & 135.4                & TitanX               & 640$\times$360              & 3.8                        & 0.4M                    \\
SQ\cite{treml2016speeding} & -           & 59.8       & 16.7                 & TitanX               & 2048$\times$1024            & 270                        & -                       \\
ICNet\cite{zhao2018icnet}  & -           & 69.5       & 30                   & TitanX M             & 2048$\times$1024            & 28.3                       & 26.5M                   \\
ESPNet\cite{mehta2018espnet}& -           & 60.3       & 113                  & TitanX               & 1024$\times$512             & -                       & 0.4M                    \\
ERFNet\cite{romera2017erfnet}& 70.0        & 68.0       & 41.7                 & TitanX M             & 1024$\times$512             & 27.7                    & 20M                     \\ \midrule
Fast-SCNN\cite{poudel2019fast}& 68.6        & 68.0       & 123.5                & TitanXp              & 2048$\times$1024            & -                        & 1.1M                    \\
DFANet A\cite{li2019dfanet}& -           & 71.3       & 100                  & TitanX               & 1024$\times$1024            & 3.4                        & 7.8M                    \\
DFANet B\cite{li2019dfanet}& -           & 67.1       & 120                  & TitanX               & 1024$\times$1024            & 2.1                        & 4.8M                    \\ \midrule
SwiftNetRN-18\cite{orsic2019defense}& 75.5        & 75.4       & 39.9                 & GTX 1080Ti           & 2048$\times$1024            & 104.0                        & 11.8M                   \\
SwiftNetRN-18 ens\cite{orsic2019defense}& -        & 76.5      & 18.4                & GTX 1080Ti           & 2048$\times$1024            & 218.0                        & 24.7M                   \\ \midrule
BiSeNet1\cite{yu2018bisenet}& 69.0        & 68.4       & 105.8                & GTX 1080Ti           & 1536$\times$768             & 14.8                        & 5.8M                    \\
BiSeNet2\cite{yu2018bisenet}& 74.8        & 74.7       & 65.5                 & GTX 1080Ti           & 1536$\times$768             & 55.3                        & 49M                     \\
BiSeNetV2\dag\cite{yu2020bisenet}& 73.4        & 72.6      &$\textbf{156}$        & GTX 1080Ti           & 1024$\times$512             & 21.1                        & -                       \\
BiSeNetV2-L\dag\cite{yu2020bisenet}& 75.8        & 75.3       & 47.3              & GTX 1080Ti           & 1024$\times$512             & 118.5                        & -                       \\ \midrule
CAS\cite{zhang2019customizable}    & 71.6        & 70.5       & 108                  & TitanXp           & 1536$\times$768            & -                        & -                       \\
GAS\cite{lin2020graph}             & 72.4         & 71.8       & 108.4                & TitanXp           & 1537$\times$769            & -                        & -                       \\ \midrule
SFNet(DF1)\cite{li2020semantic}& -            & 74.5       & 74               & GTX 1080Ti           & 2048$\times$1024             & -                        & 9.03M                       \\
SFNet(DF2)\cite{li2020semantic}& -            & 77.8       & 53               & GTX 1080Ti           & 2048$\times$1024             & -                        & 10.53M                       \\
SFNet(ResNet-18)\cite{li2020semantic}& -      &78.9     & 18       & GTX 1080Ti           & 2048$\times$1024             & 247                        & 12.87M                       \\ \midrule
MSFNet*\cite{si2019real}& -          & 71.3       & 117                  & GTX 2080Ti           & 1024$\times$512            & 24.2                        & -                       \\
MSFNet\cite{si2019real}& -           & 77.1       & 41                   & GTX 2080Ti           & 2048$\times$1024            & 96.8                        & -                       \\ \midrule
CABiNet\cite{kumaar2020cabinet}& 76.6      &75.9     & 76.5       & GTX 2080Ti           & 2048$\times$1024             & 12.0                        & 2.64M                       \\ \midrule
DDRNet-23-Slim& $\textbf{77.8}$(77.3$\pm$0.4)    & 77.4     & 101.6         & GTX 2080Ti           & 2048$\times$1024         & 36.3                        & 5.7M                   \\
DDRNet-23     & $\textbf{79.5}$(79.1$\pm$0.3)    & 79.4  & 37.1  & GTX 2080Ti           & 2048$\times$1024            & 143.1                        & 20.1M                   \\
DDRNet-39     & -             & $\textbf{80.4}$  & 22.0  & GTX 2080Ti           & 2048$\times$1024            & 281.2                        & 32.3M                   \\ \bottomrule
\end{tabular}
\end{table*}

\subsubsection{Cityscapes}

Following \cite{9052469}, we use the SGD optimizer with the initial learning rate of 0.01, the momentum of 0.9, and the weight decay of 0.0005. We adopt the ploy learning policy with the power of 0.9 to drop the learning rate and implement the data augmented method including random cropping images, random scaling in the range of 0.5 to 2.0, and random horizontal flipping. Images are randomly cropped into 1024$\times$1024 for training following \cite{li2019dfanet},\cite{si2019real},\cite{li2020semantic}. All the models are trained with 484 epochs (about 120$K$ iterations), a batch size of 12, and syncBN on four 2080Ti GPUs. For the models evaluated on the test server, we feed images from train and val set simultaneously during training. For a fair comparison with \cite{yu2020bisenet} and \cite{li2020semantic}, online hard example mining (OHEM)\cite{shrivastava2016training} is also used.

\subsubsection{CamVid}

We set the initial learning rate to 0.001 and train all the models for 968 epochs. Images are randomly cropped into 960$\times$720 for training following \cite{li2019dfanet}. All the models are trained on a single GPU and other training details are identical to those for Cityscapes. When employing Cityscapes pre-train, we finetune the models for 200 epochs.

\subsubsection{COCOStuff}

The initial learning rate is 0.001 and the total training epochs are 110. We resize the short side of the images to 640 before data augmentation. The crop size is 640$\times$640, as same as that of BiSeNetV2\cite{yu2020bisenet}. Other training details are identical to those for Cityscapes while the weight decay is 0.0001. In the inference phase, we fix the image resolution to 640$\times$640.

\subsection{Measure of Inference Speed and Accuracy}

The inference speed is measured on a single GTX 2080Ti GPU by setting the batch size to 1, with CUDA 10.0, CUDNN 7.6, and PyTorch 1.3. Similar to MSFNet and SwiftNet, we exclude batch normalization layers after convolutional layers because they can be integrated into convolutions during inference. We use the protocols established by \cite{chen2019fasterseg} for a fair comparison (image size: 2048$\times$1024 for Cityscapes, 960$\times$720 for CamVid, and 640$\times$640 for COCOStuff).

Following ResNet\cite{he2016deep}, we report the best results, average results, and standard deviations of four trials except for the cityscapes test set of which accuracy is provided by the official server.

\subsection{Speed and Accuracy Comparisons}

\subsubsection{Cityscapes}

As can be observed from Table \ref{tab:1} and Fig. \ref{fig4}, our method achieves a new state-of-the-art trade-off between real-time and high accuracy. Specially, DDRNet-23-slim (our smallest model) achieves 77.4$\%$ mIoU on the test set at 102 FPS. It outperforms DFANet A and MSFNet* by 6.1$\%$ mIoU with similar inference speed, and reasons approximately 2.5 times as fast as MSFNet. Besides, it runs 40$\%$ faster than the smallest SFNet and achieves a 2.9$\%$ mIoU gain on the test set. It is worth noting that our method also towers over those methods based on architecture search for real-time semantic segmentation such as CAS\cite{zhang2019customizable} and GAS\cite{lin2020graph} with similar inference speed. For the wider model, DDRNet-23 achieves the overall best accuracy among the published real-time methods in Table \ref{tab:1}, attaining 79.4$\%$ mIoU at 37 FPS. DDRNet-23 has a performance gain of 0.5$\%$ over SFNet (ResNet-18) and runs twice as fast as it.

\begin{figure*}
\centering
\begin{minipage}{1.6in}
\includegraphics[width=1.6in]{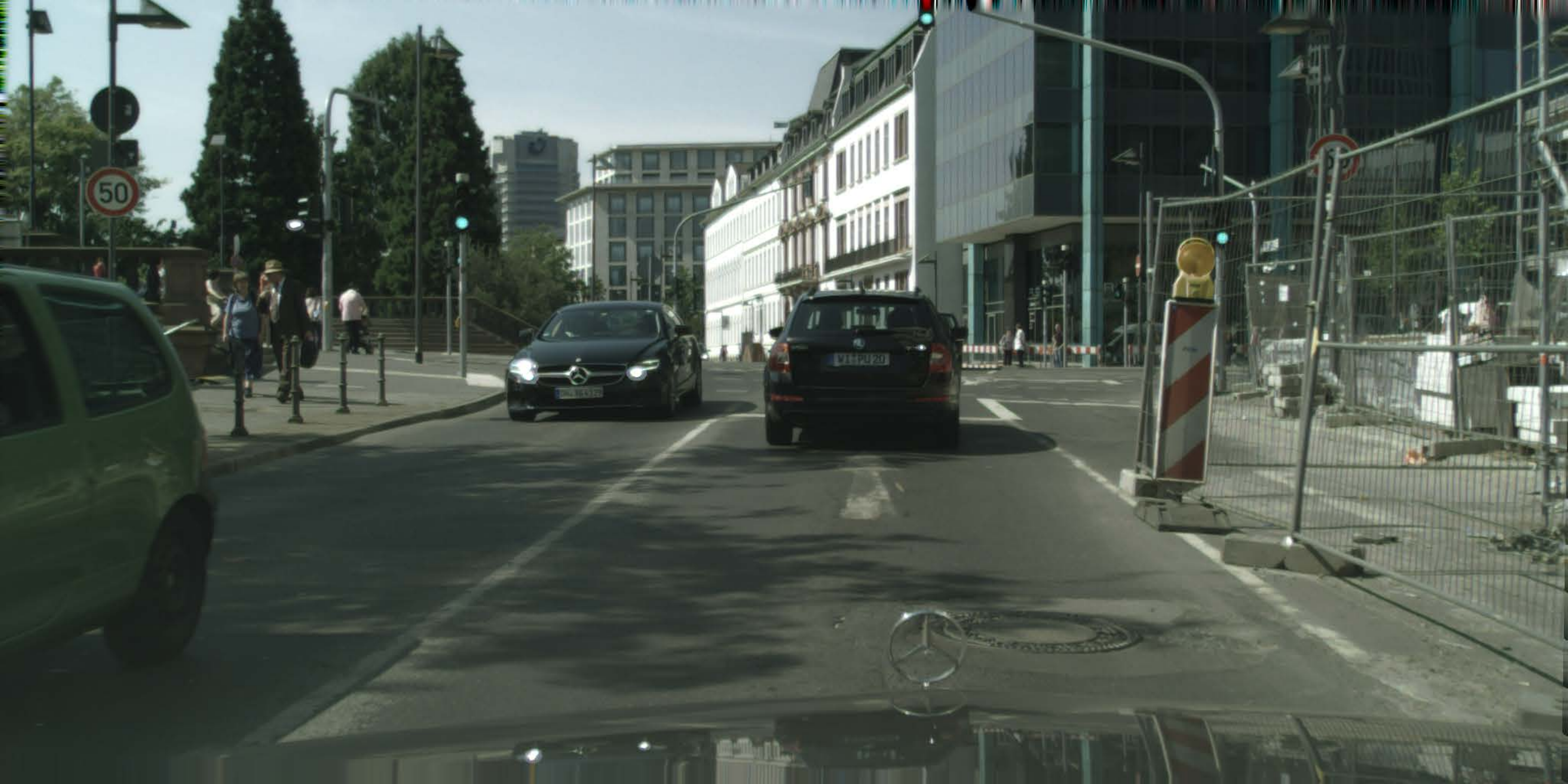}
\end{minipage}
\begin{minipage}{1.6in}
\includegraphics[width=1.6in]{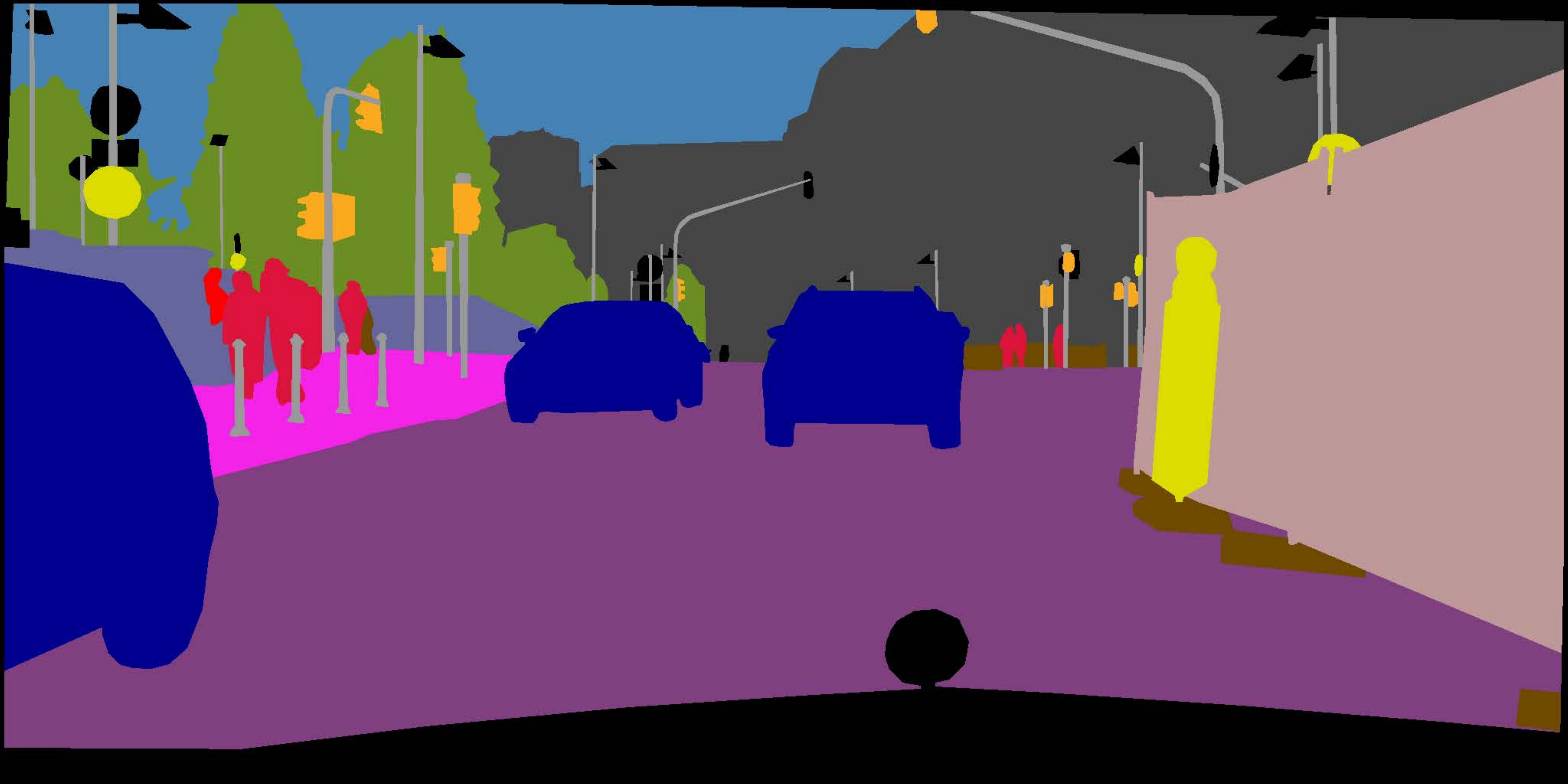}
\end{minipage}
\begin{minipage}{1.6in}
\includegraphics[width=1.6in]{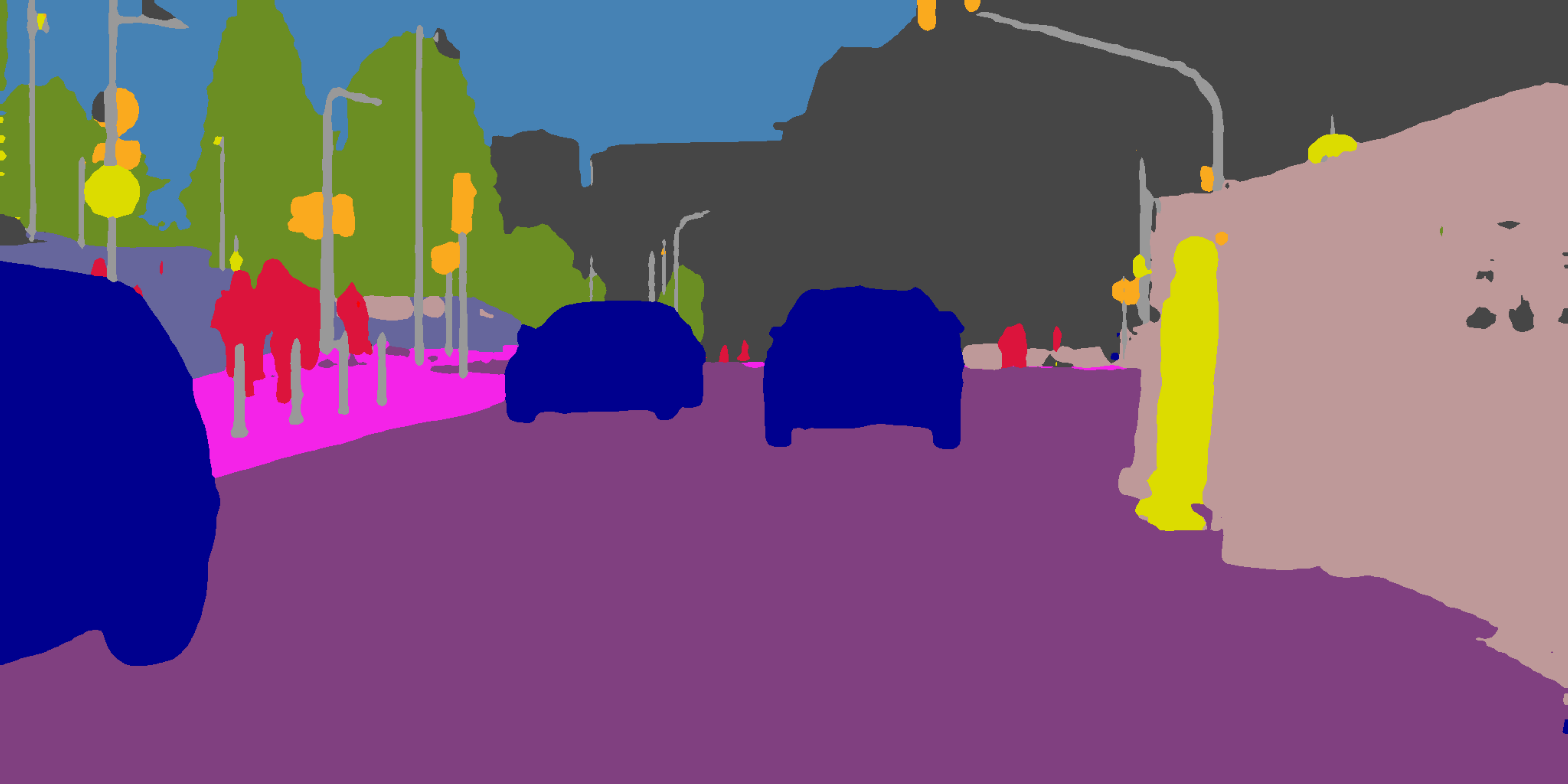}
\end{minipage}
\vspace{0.3em}
\begin{minipage}{1.6in}
\includegraphics[width=1.6in]{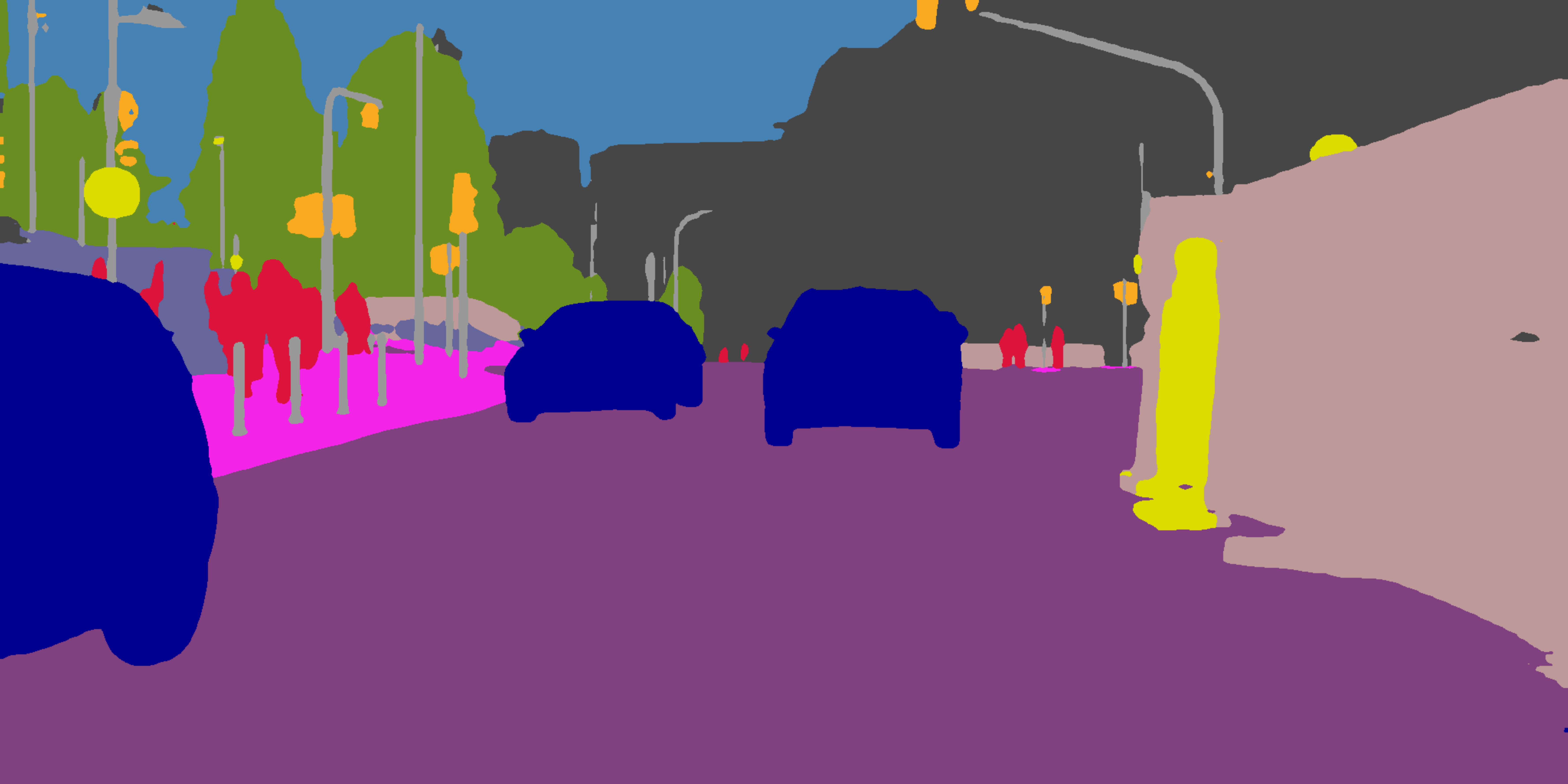}
\end{minipage}
\vspace{0.3em}
\begin{minipage}{1.6in}
\includegraphics[width=1.6in]{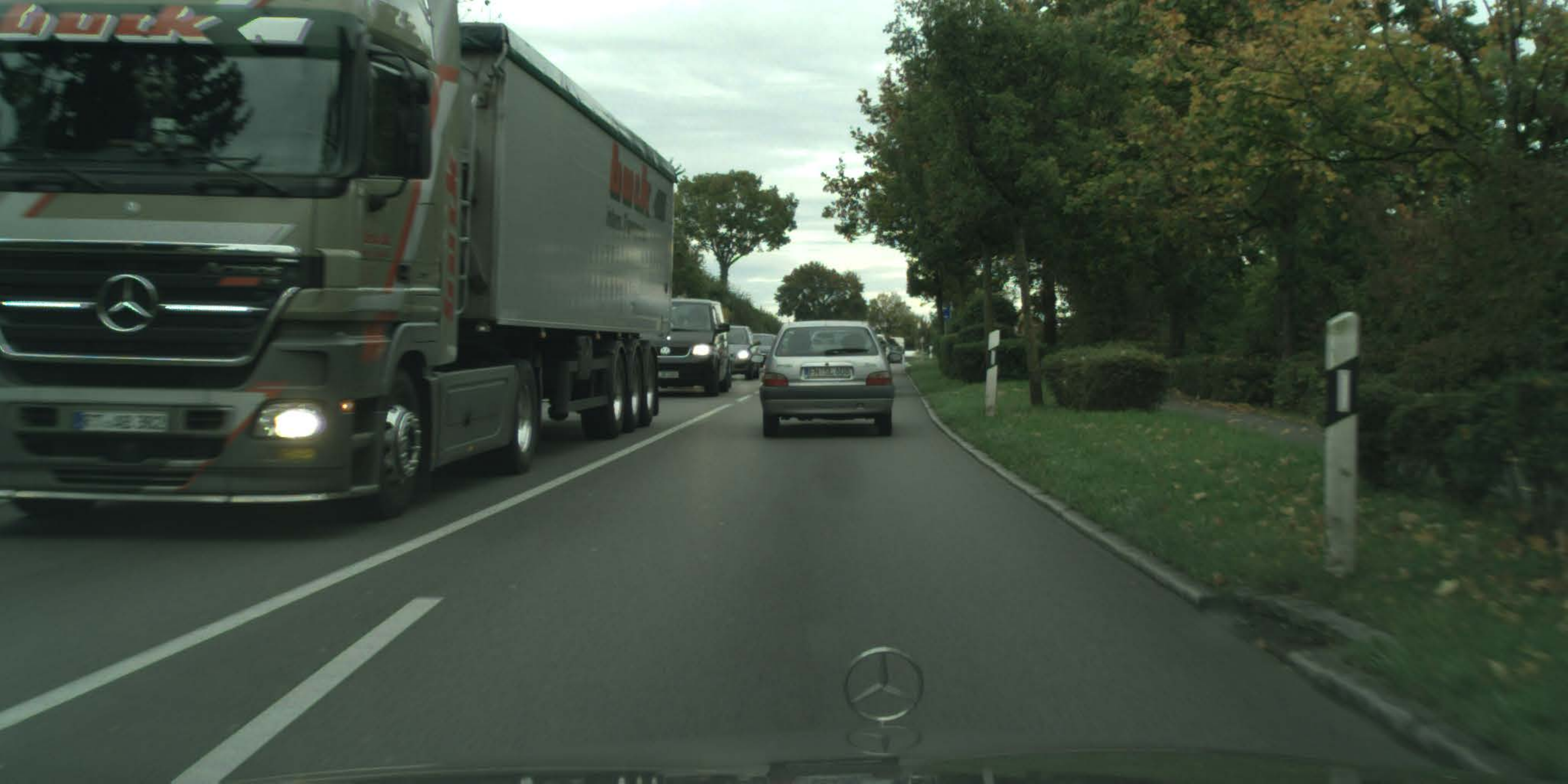}
\end{minipage}
\begin{minipage}{1.6in}
\includegraphics[width=1.6in]{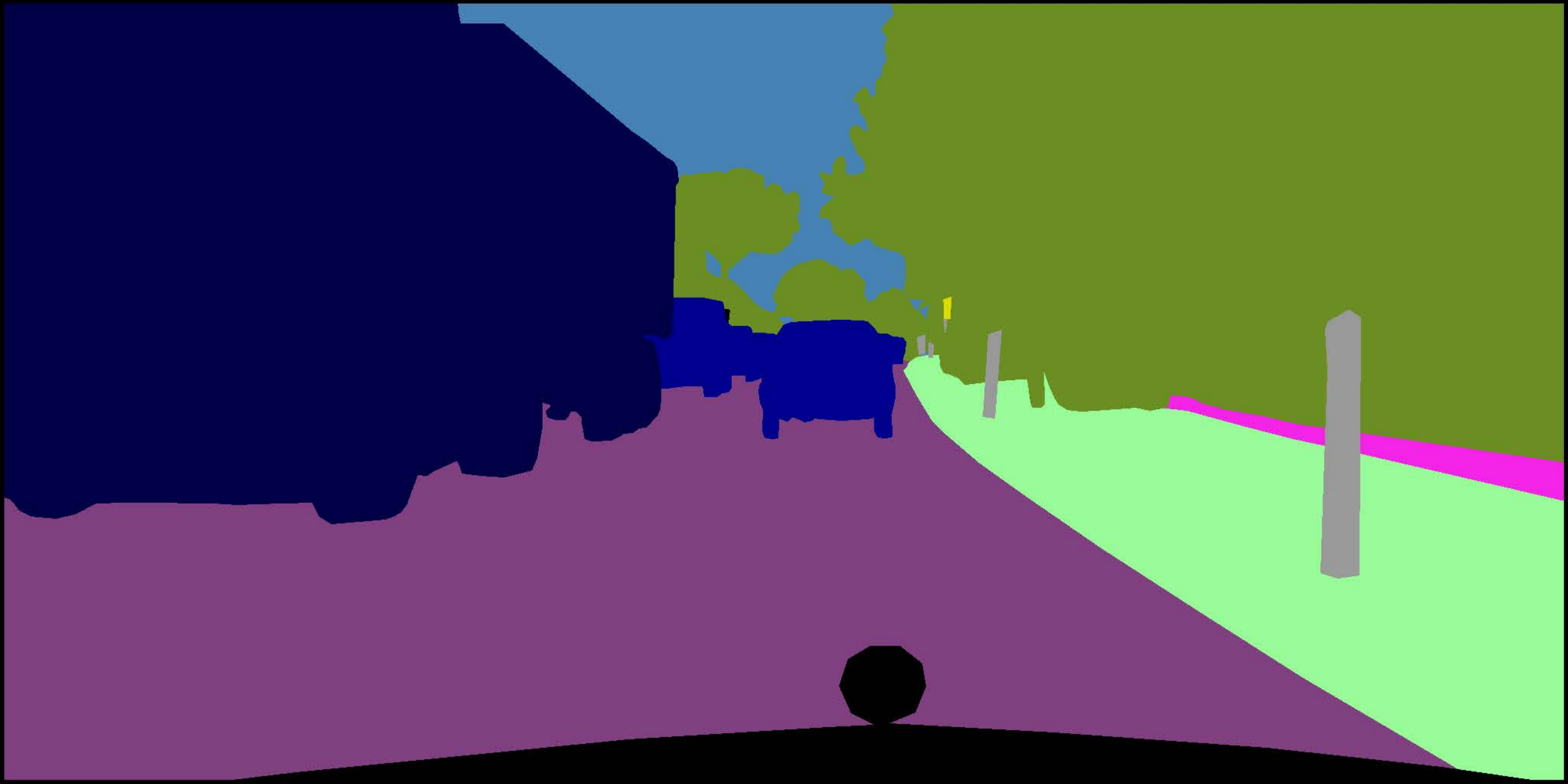}
\end{minipage}
\begin{minipage}{1.6in}
\includegraphics[width=1.6in]{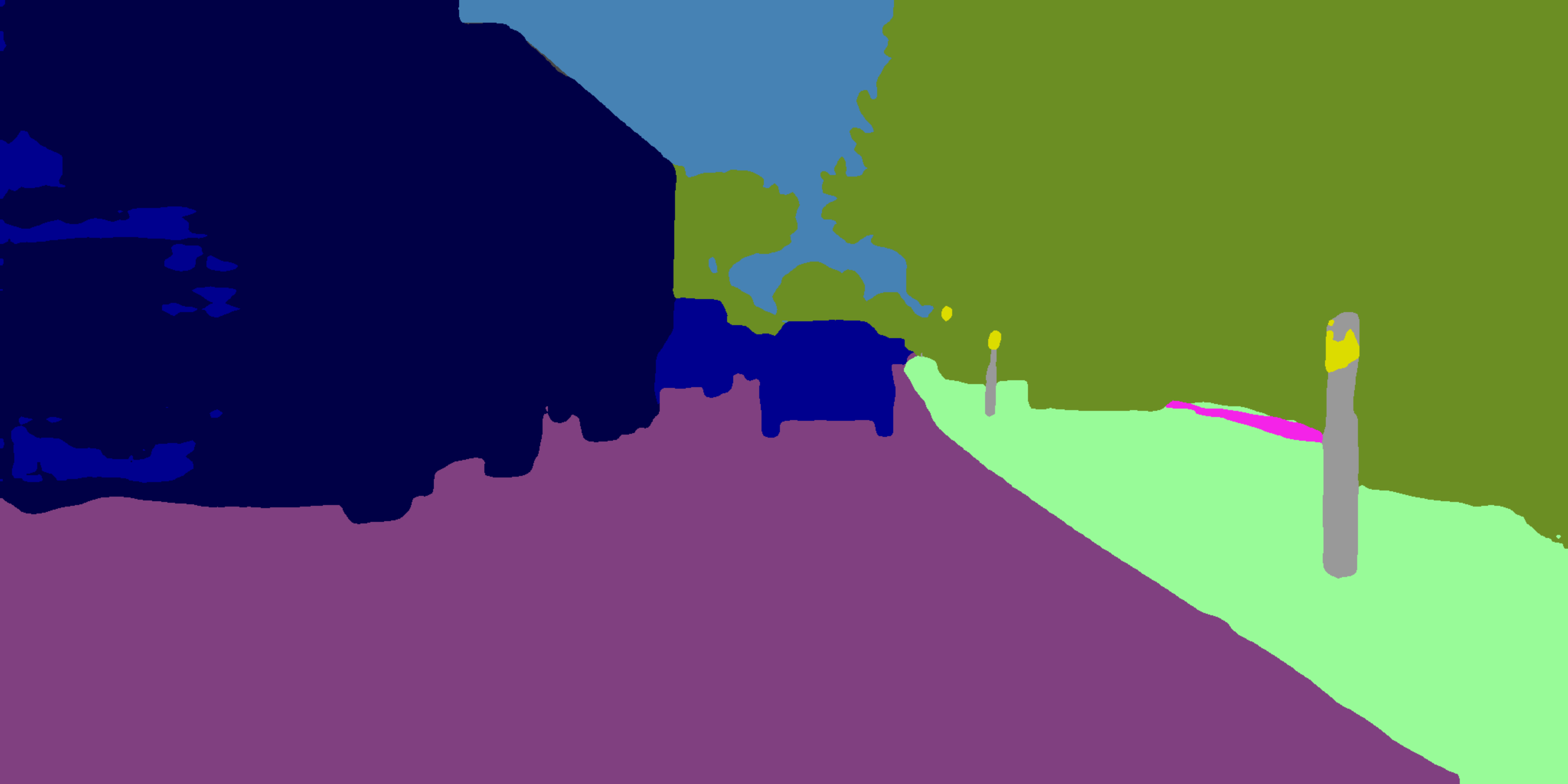}
\end{minipage}
\begin{minipage}{1.6in}
\includegraphics[width=1.6in]{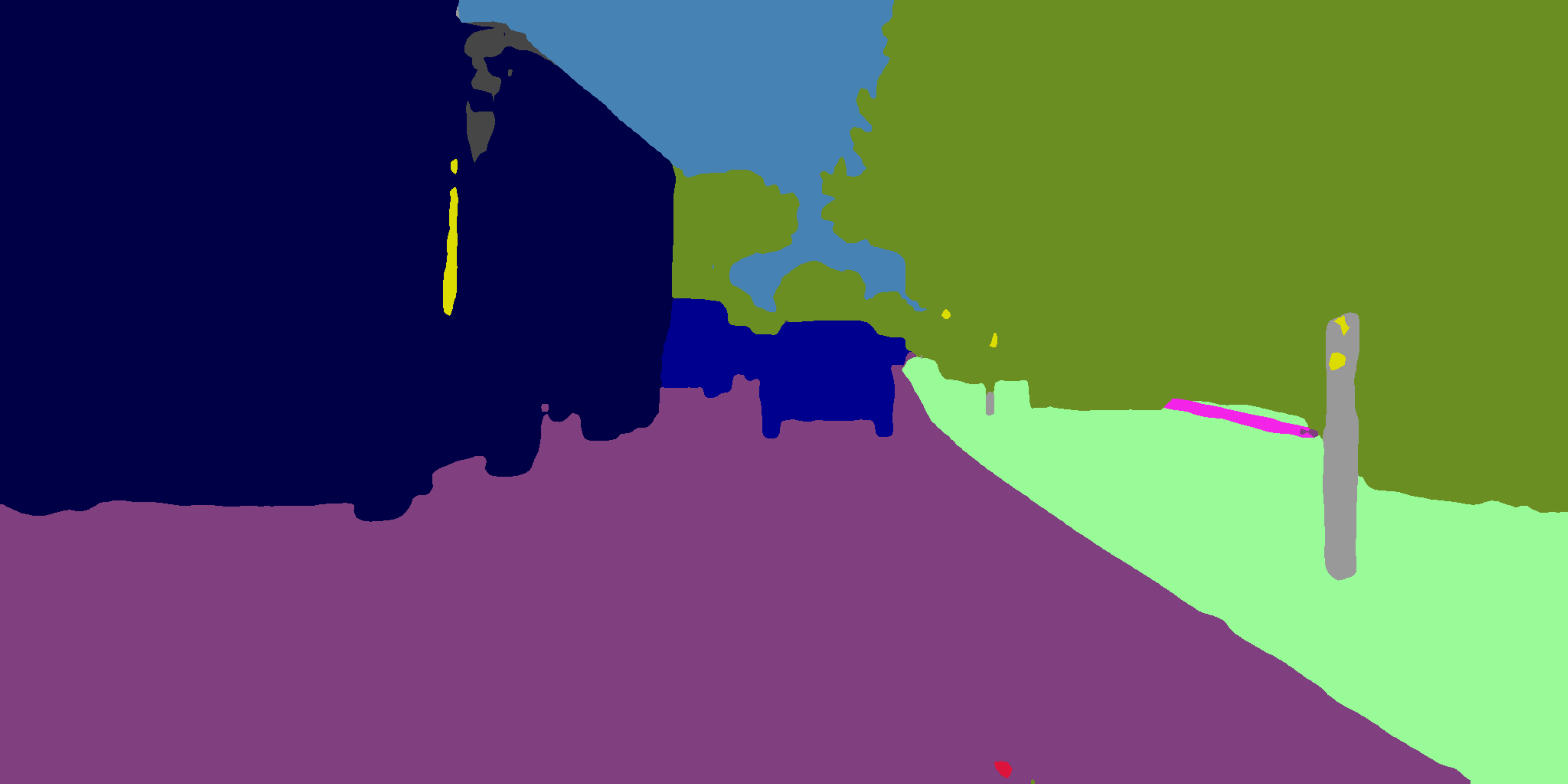}
\end{minipage}
\vspace{0.3em}
\begin{minipage}{1.6in}
\includegraphics[width=1.6in]{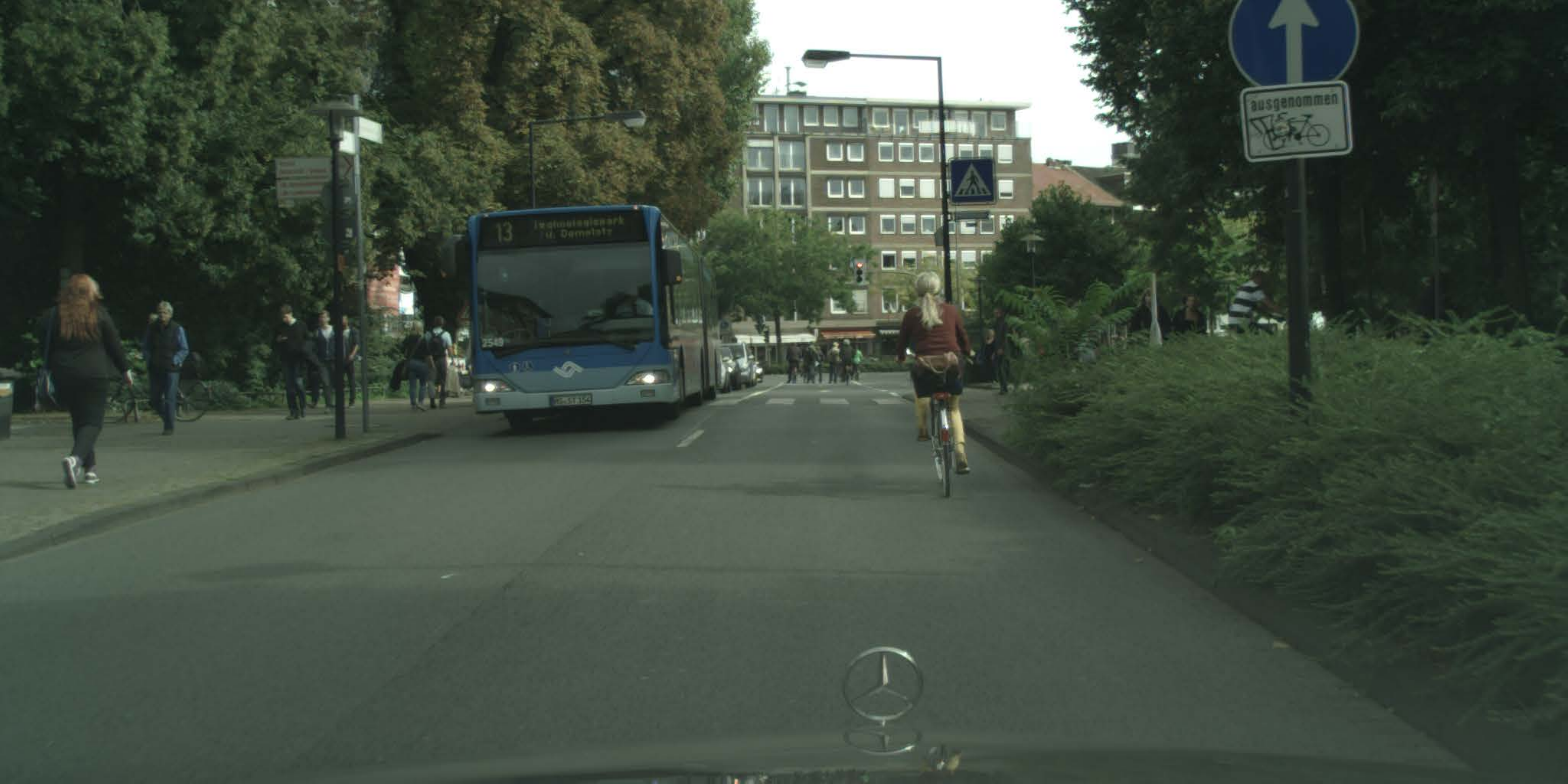}
\end{minipage}
\begin{minipage}{1.6in}
\includegraphics[width=1.6in]{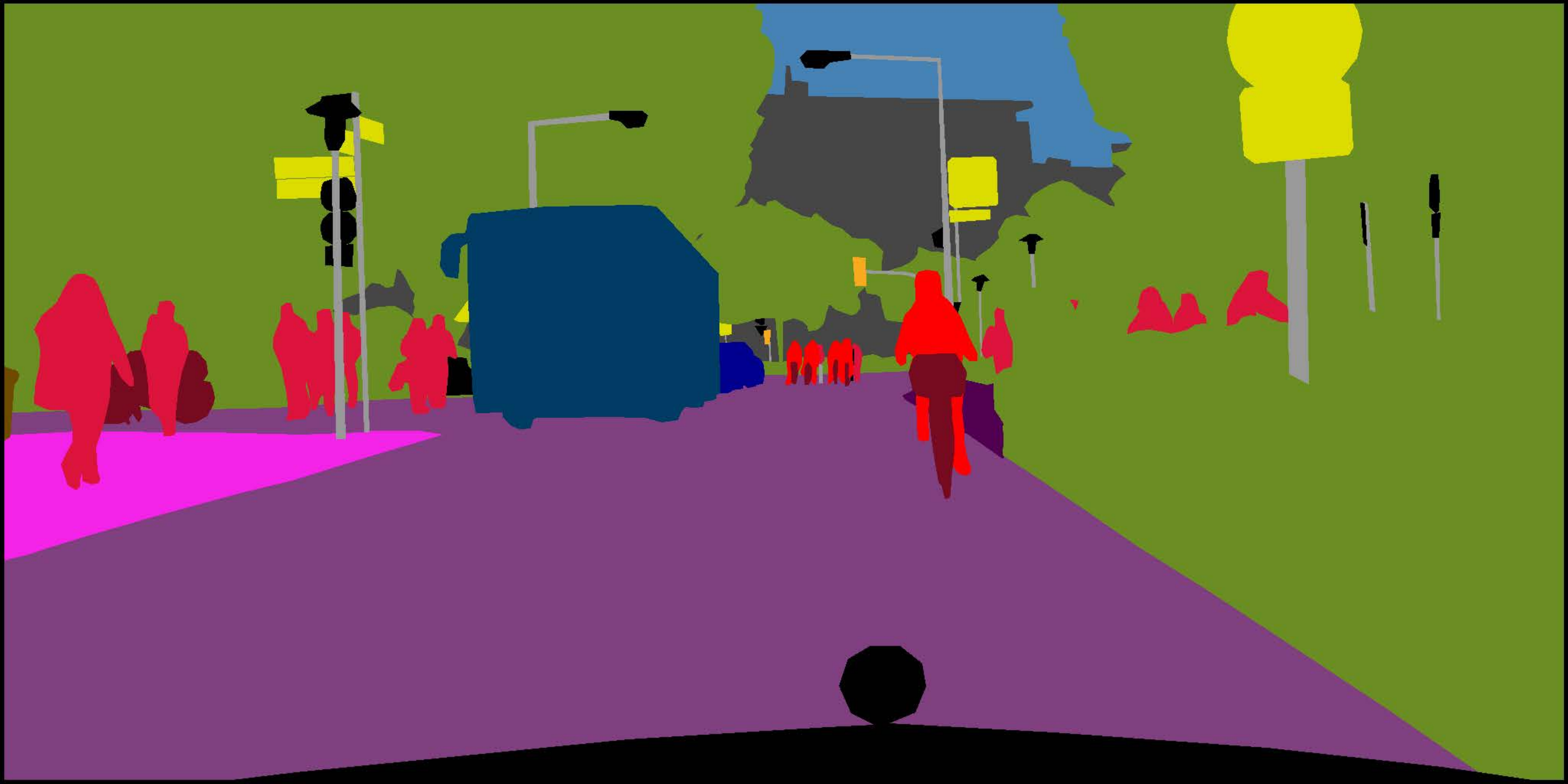}
\end{minipage}
\begin{minipage}{1.6in}
\includegraphics[width=1.6in]{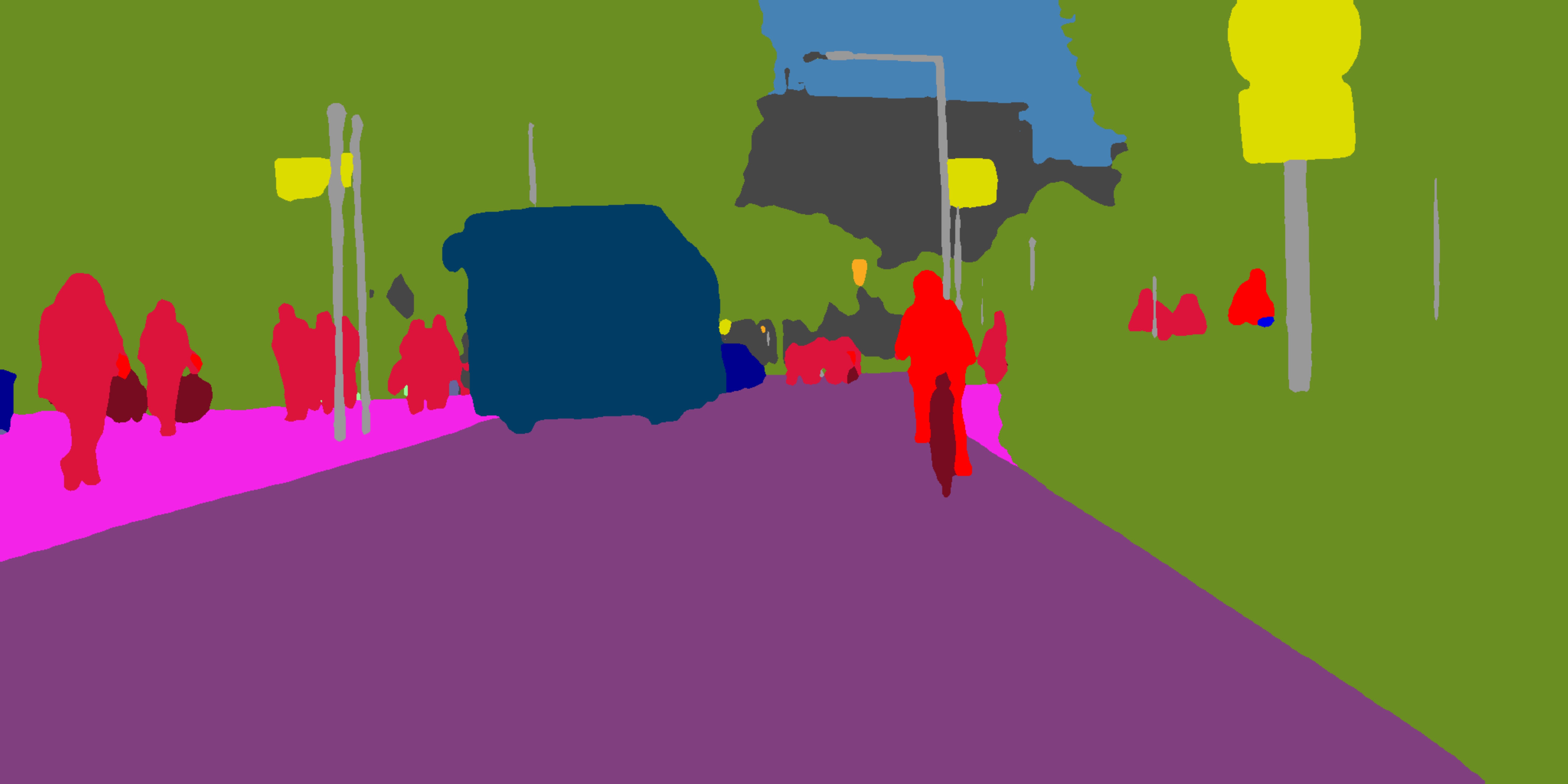}
\end{minipage}
\begin{minipage}{1.6in}
\includegraphics[width=1.6in]{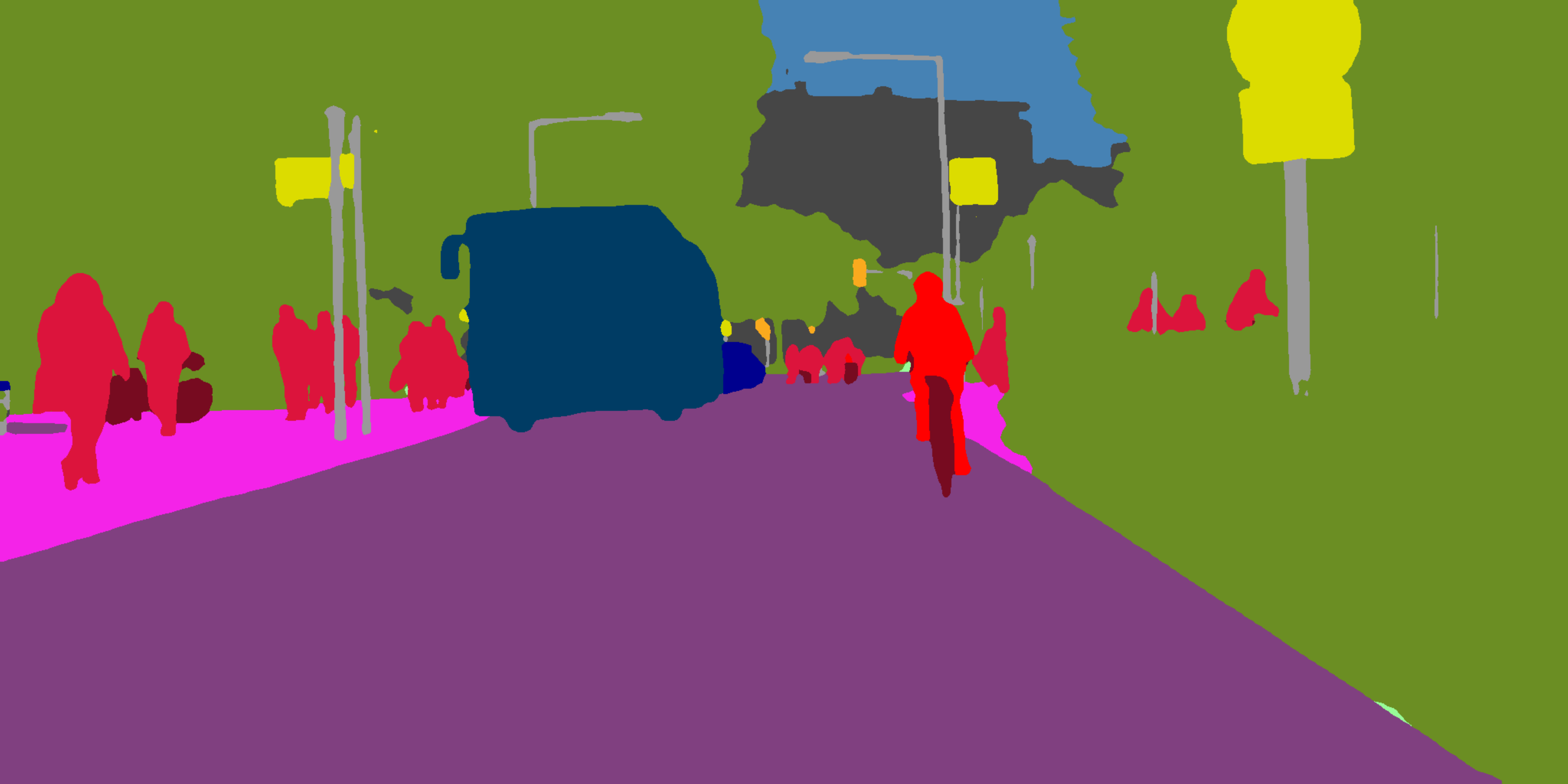}
\end{minipage}
\vspace{0.3em}
\begin{minipage}{1.6in}
\includegraphics[width=1.6in]{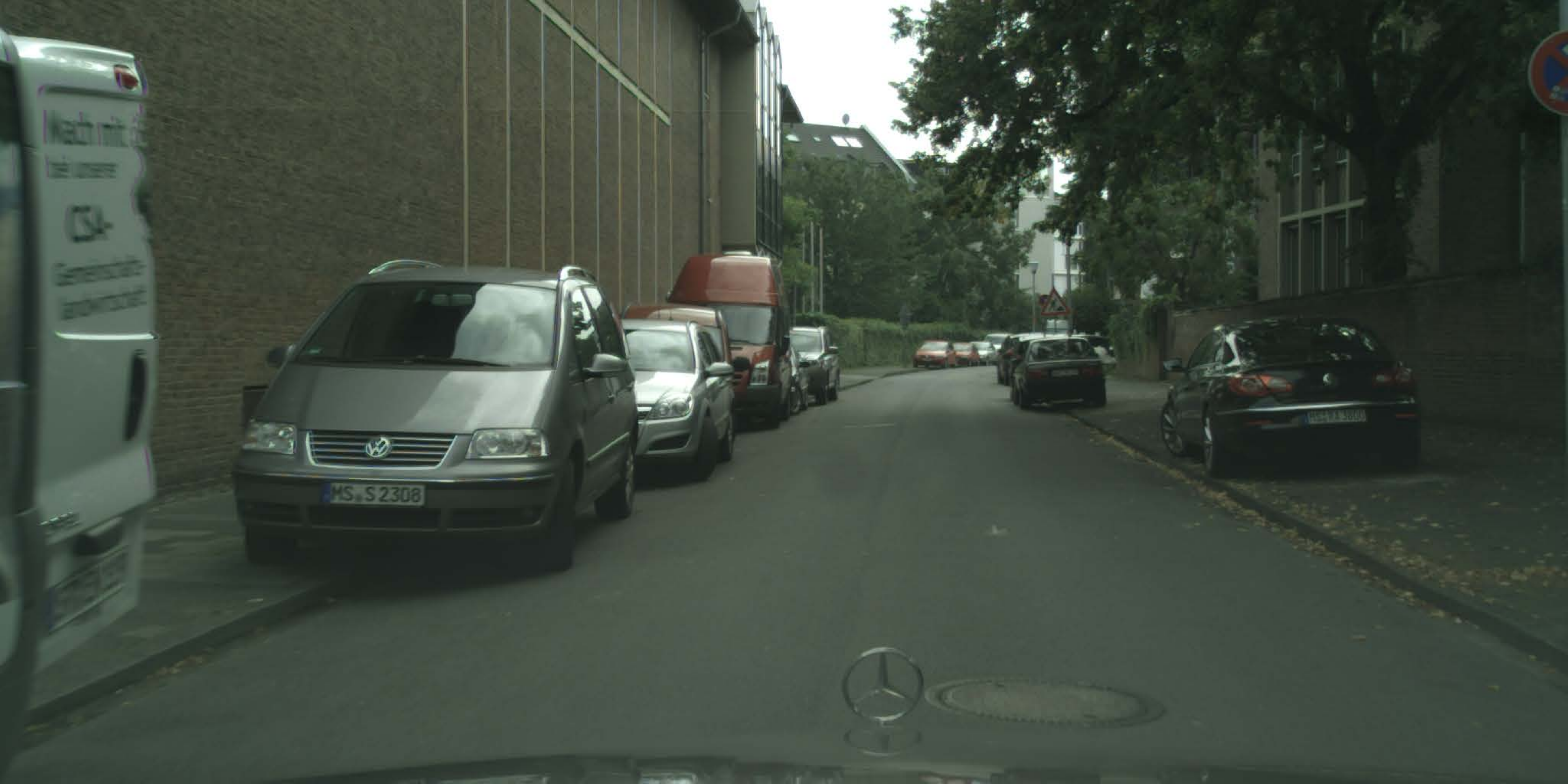}
\end{minipage}
\begin{minipage}{1.6in}
\includegraphics[width=1.6in]{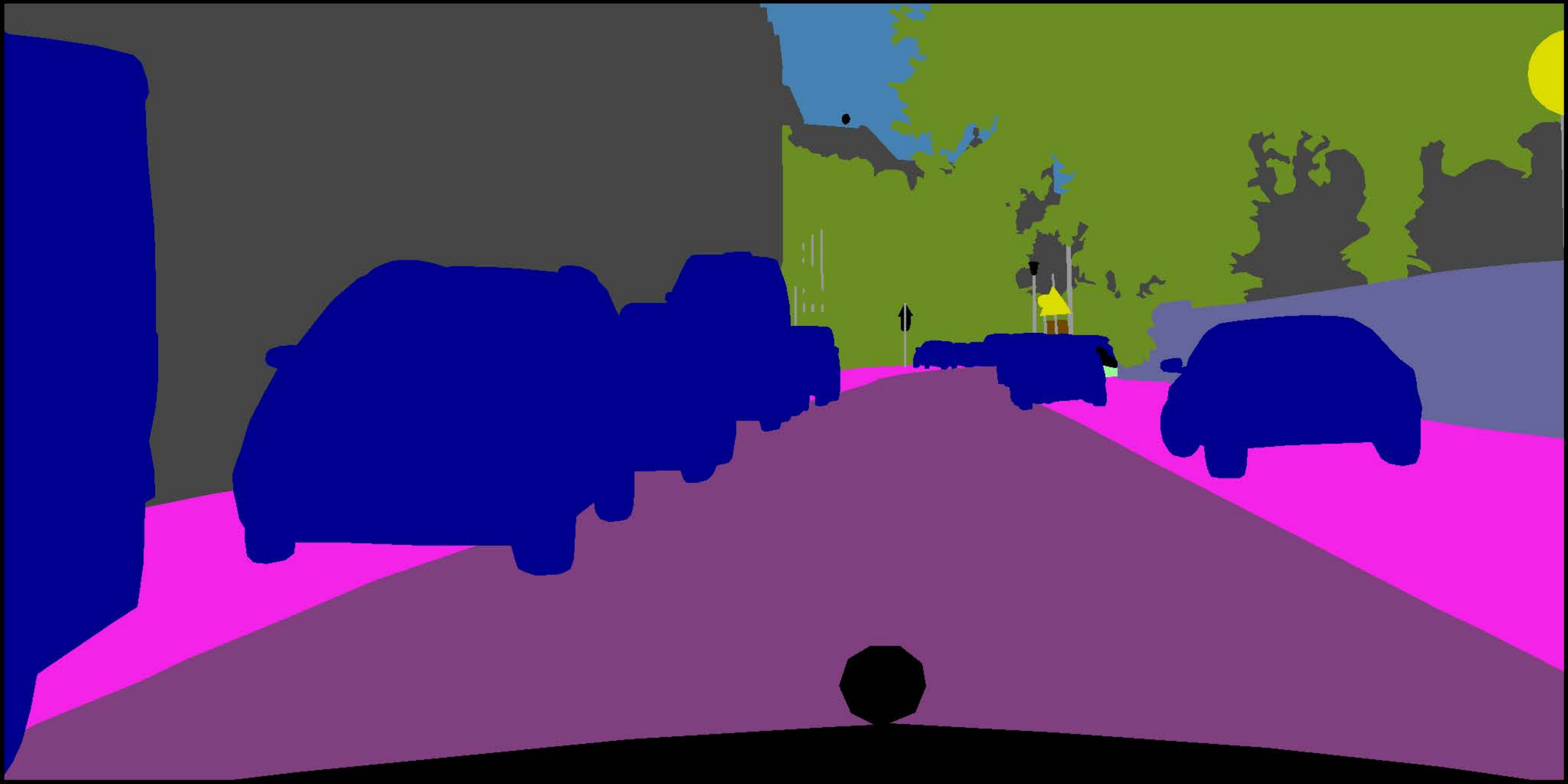}
\end{minipage}
\begin{minipage}{1.6in}
\includegraphics[width=1.6in]{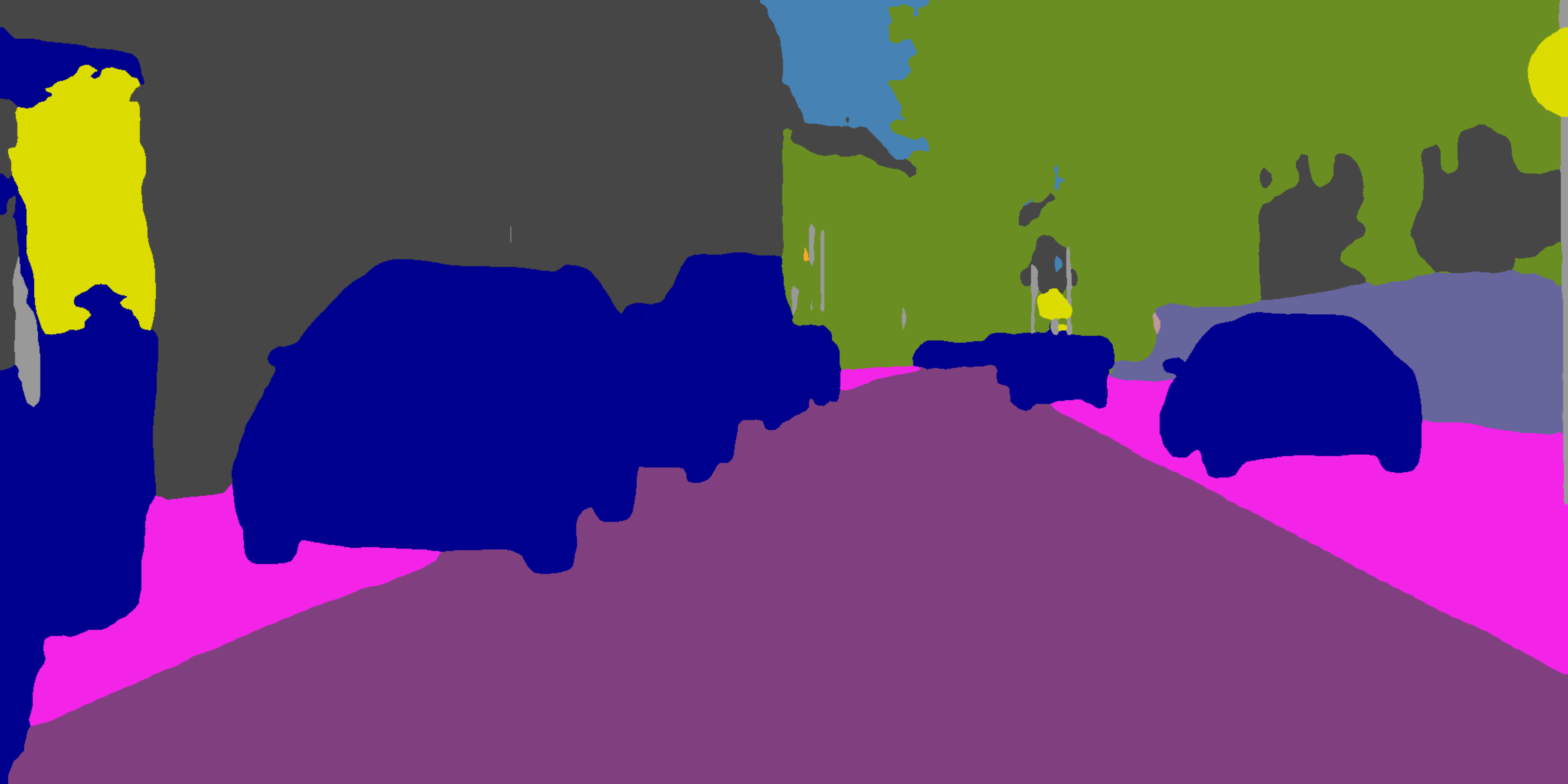}
\end{minipage}
\begin{minipage}{1.6in}
\includegraphics[width=1.6in]{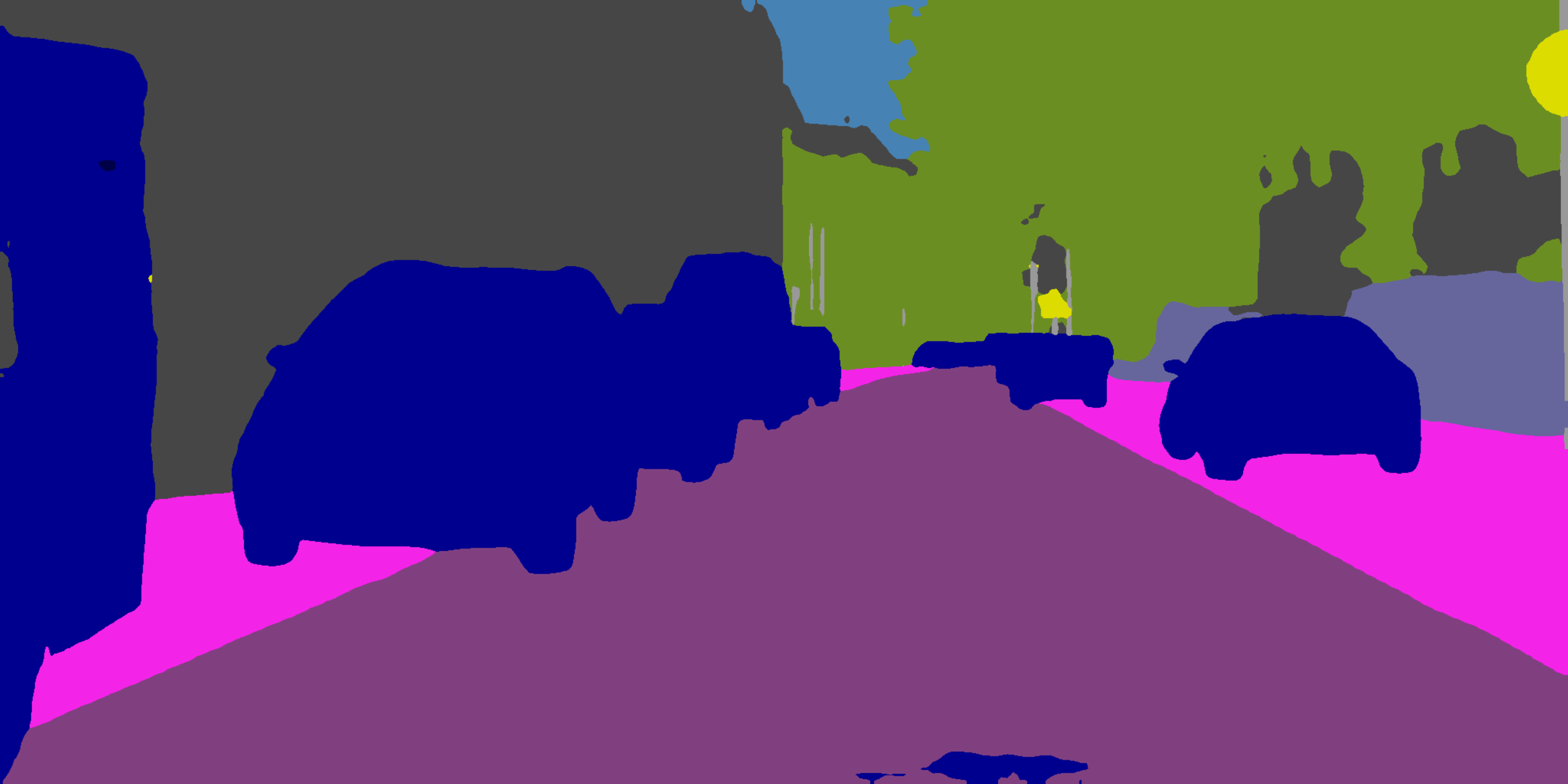}
\end{minipage}
\vspace{0.3em}
\begin{minipage}{1.6in}
\includegraphics[width=1.6in]{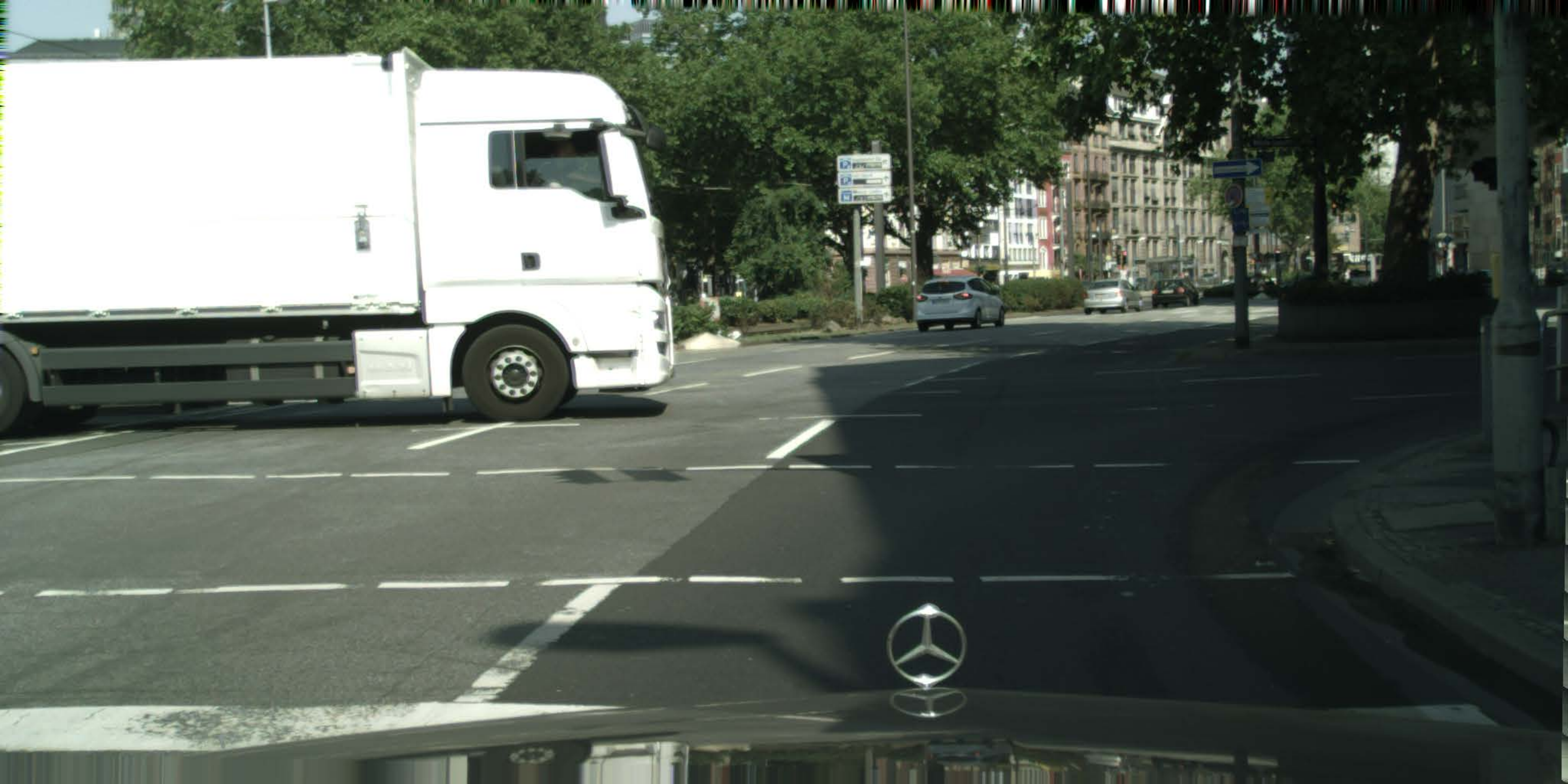}
\end{minipage}
\begin{minipage}{1.6in}
\includegraphics[width=1.6in]{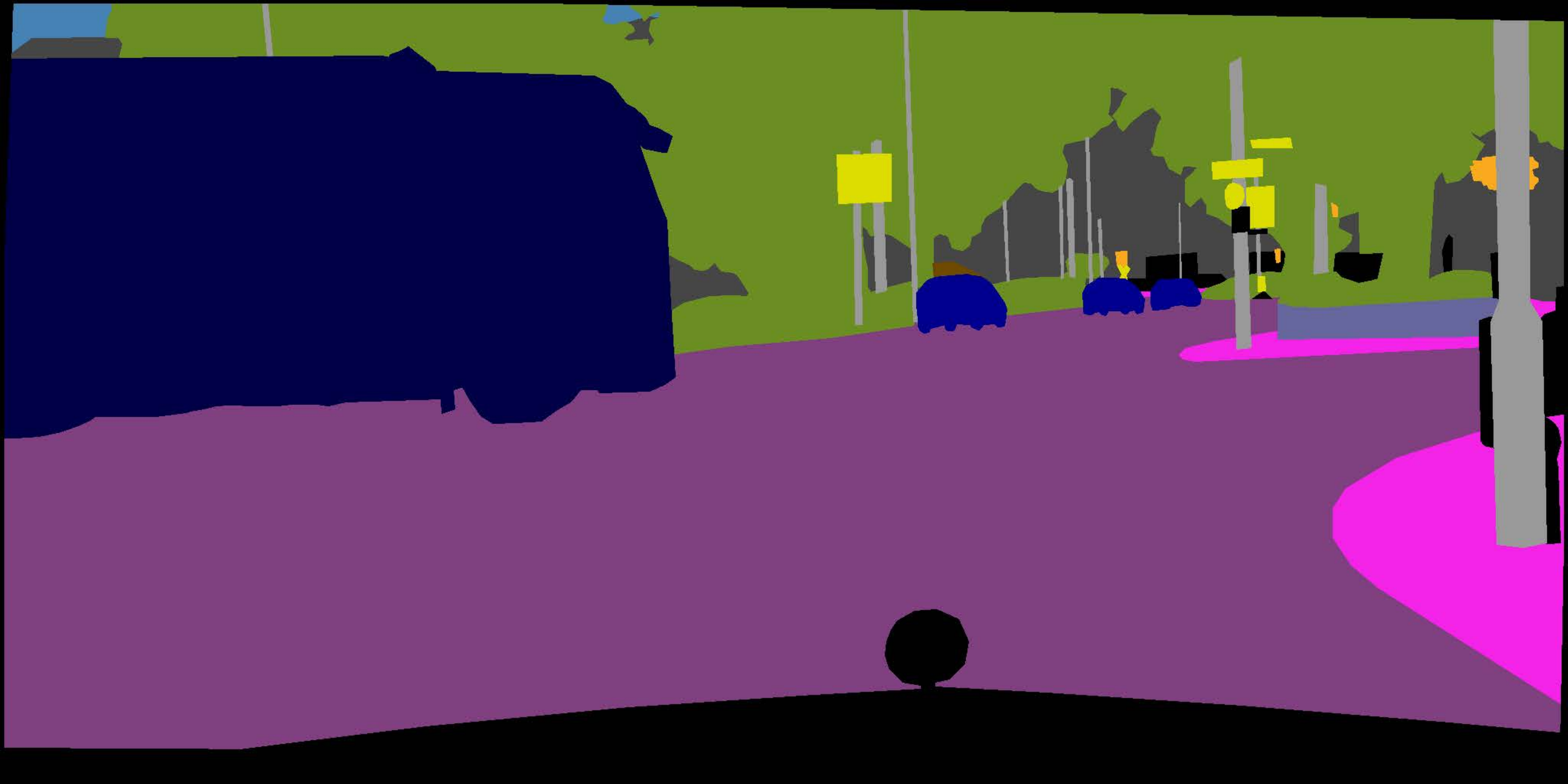}
\end{minipage}
\begin{minipage}{1.6in}
\includegraphics[width=1.6in]{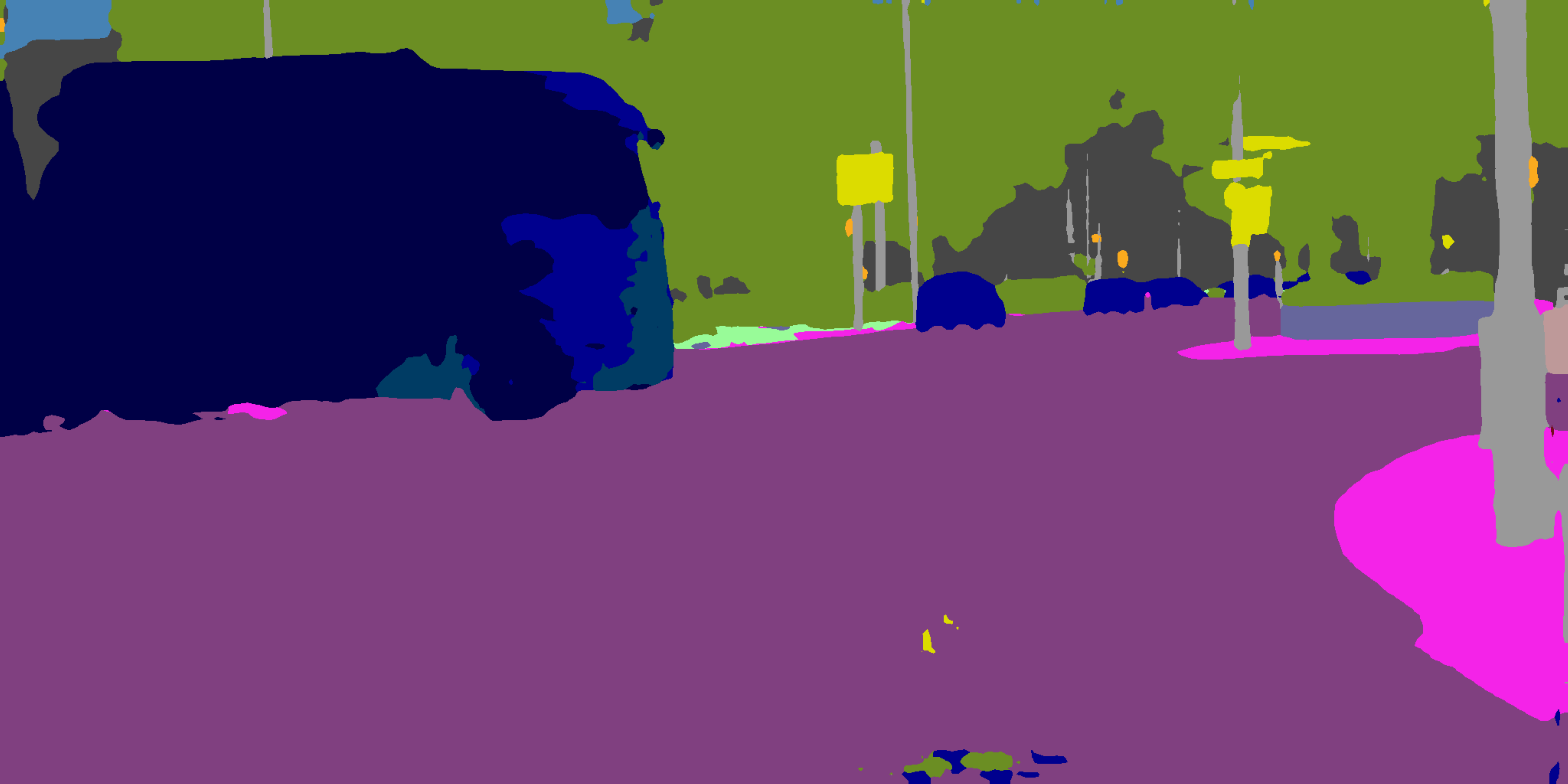}
\end{minipage}
\begin{minipage}{1.6in}
\includegraphics[width=1.6in]{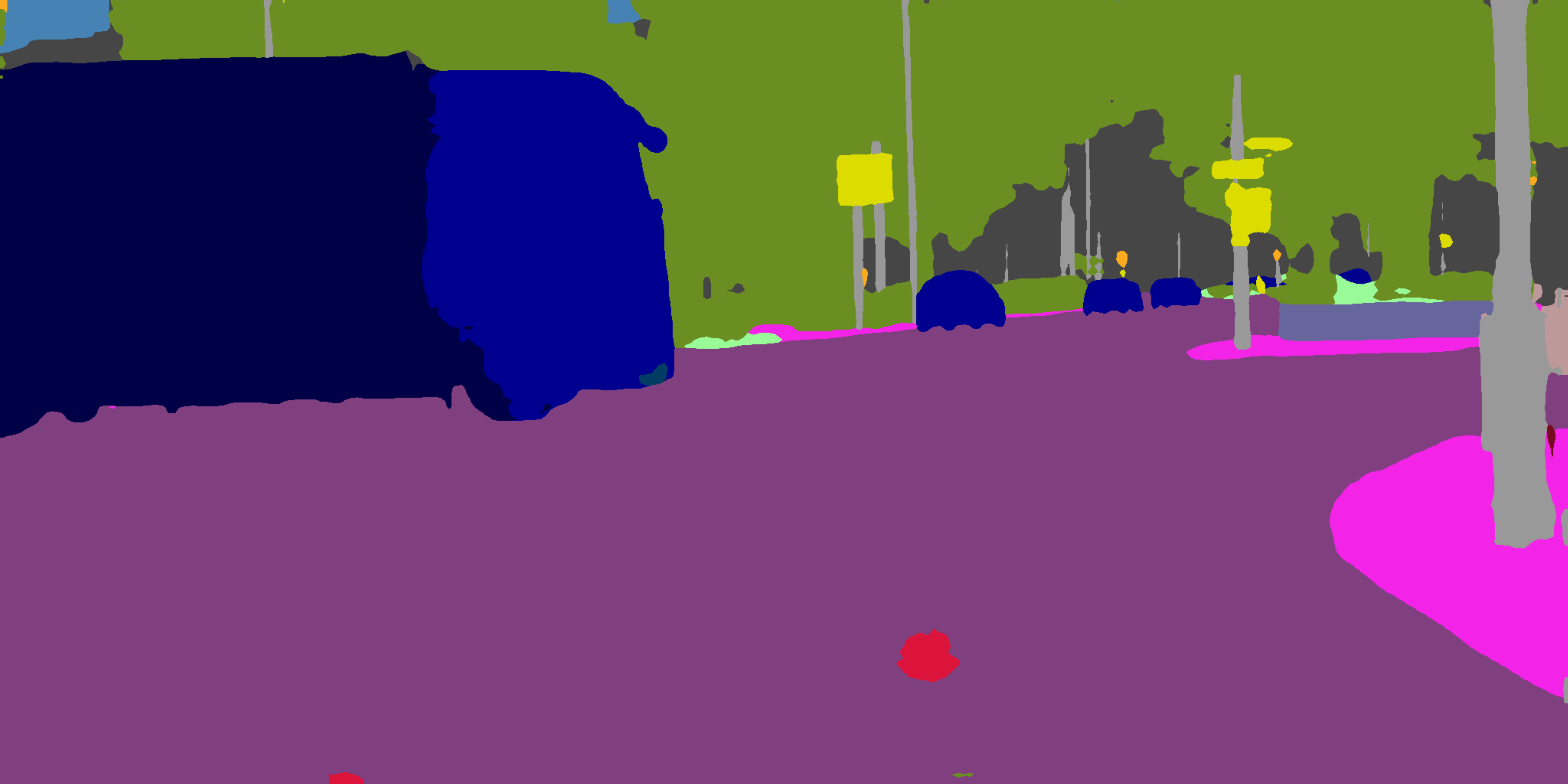}
\end{minipage}
\vspace{0.3em}
\begin{minipage}{1.6in}
\includegraphics[width=1.6in]{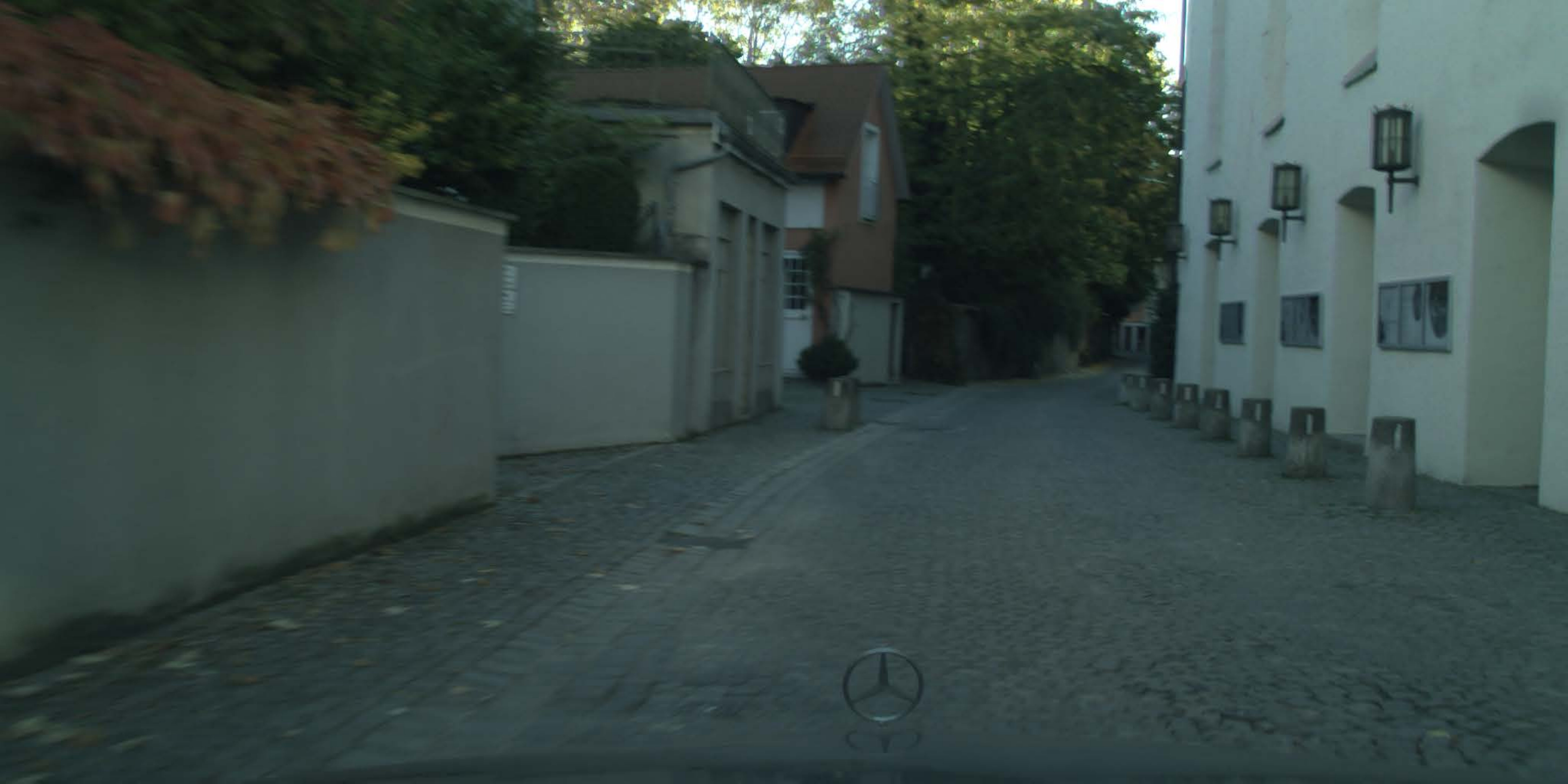}
\end{minipage}
\begin{minipage}{1.6in}
\includegraphics[width=1.6in]{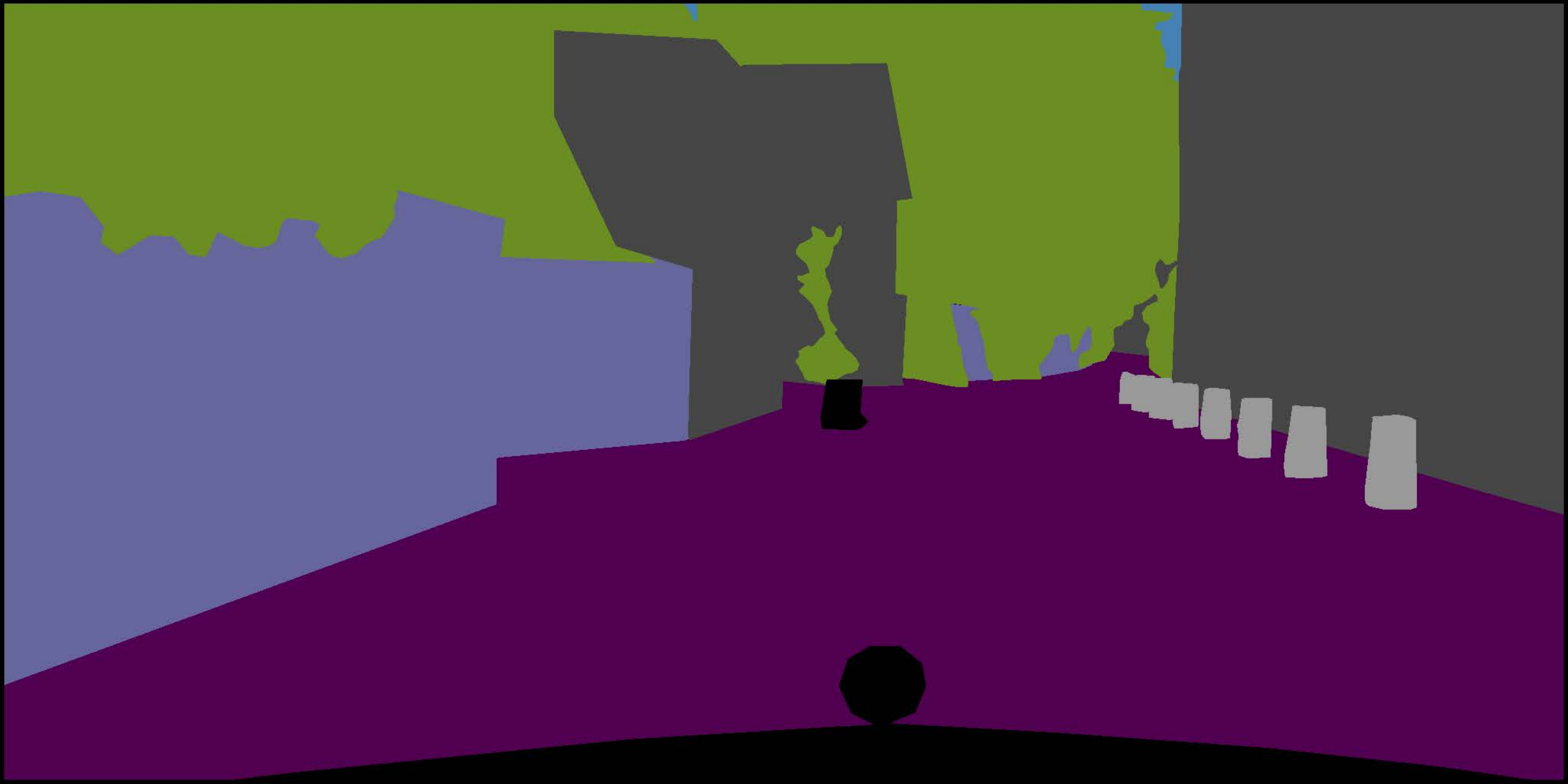}
\end{minipage}
\begin{minipage}{1.6in}
\includegraphics[width=1.6in]{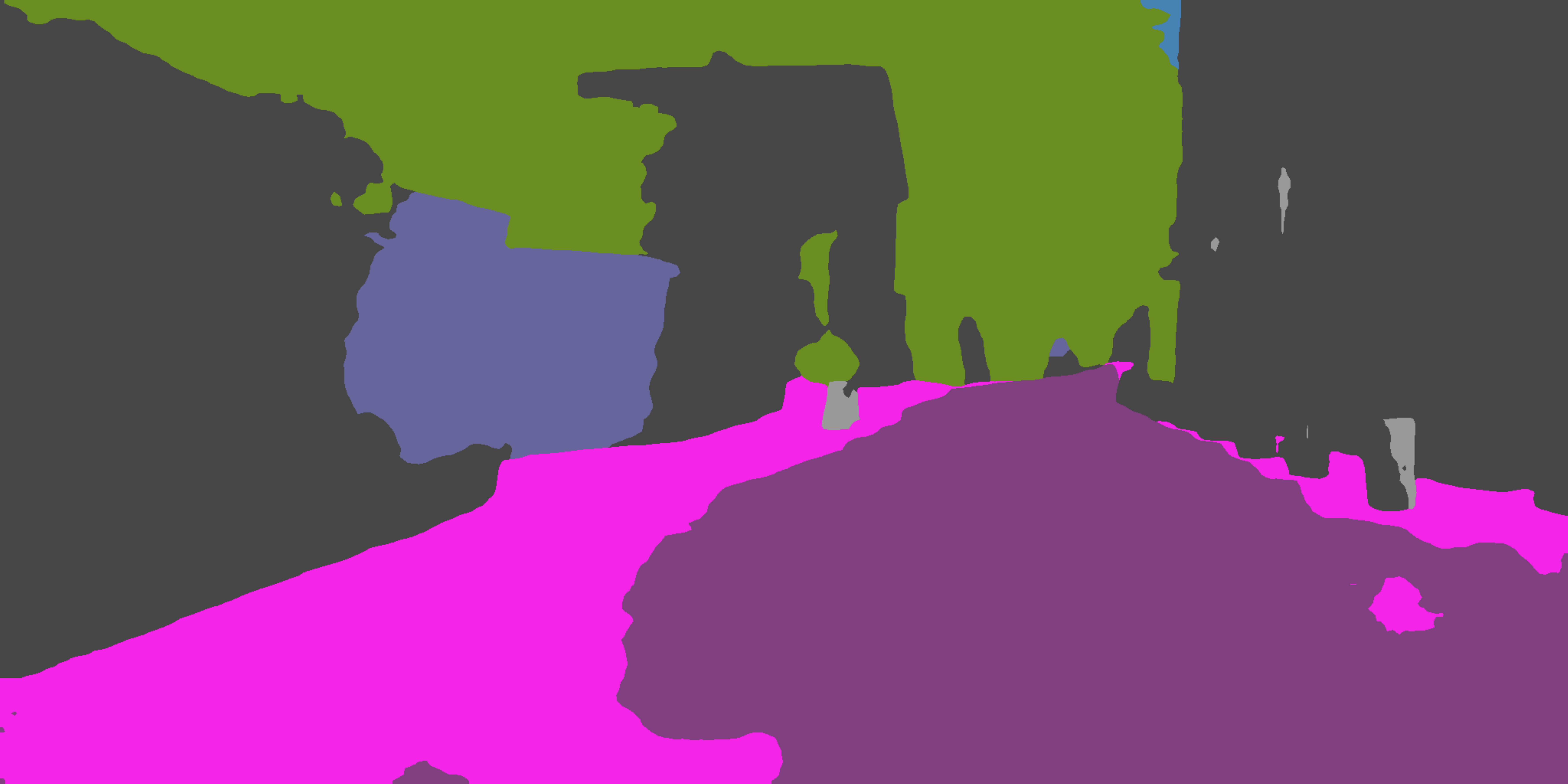}
\end{minipage}
\begin{minipage}{1.6in}
\includegraphics[width=1.6in]{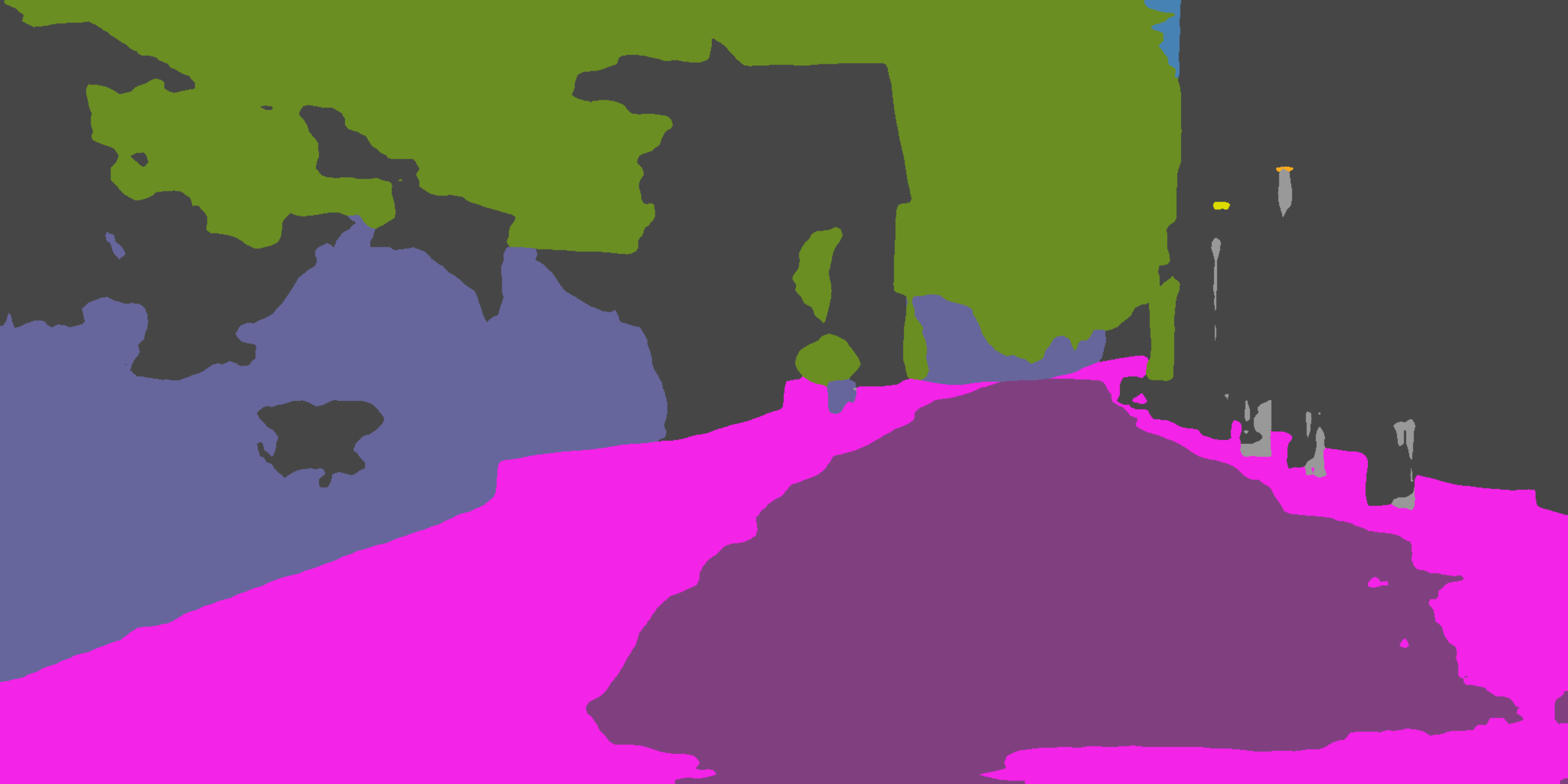}
\end{minipage}
\caption{Visualized segmentation results on Cityscapes val set. The four columns left-to-right refer to the input image, the ground truth, the output of DDRNet-23-slim, and the output of DDRNet-23. The first four rows show the performance of two models while the last two rows represent some segmentation failures.}
\label{fig7}
\end{figure*}

We keep going deeper with DDRNets and achieve 80.4$\%$ mIoU on the Cityscapes test server at 22 FPS, only using finely annotated data. If benefitting from pre-training on Mapillary\cite{neuhold2017mapillary} dataset and TensorRT acceleration like \cite{li2020semantic}, our method can establish a skyscraping baseline for real-time semantic segmentation of road scenes. On the Cityscapes val set, DDRNet-23-slim is superior to all published results in Table \ref{tab:1} with 36.3 GFLOPs and 5.7M parameters. And DDRNet-23 achieves a new overall best result of 79.5$\%$ mIoU. Fig. \ref{fig7} shows the visualized results of DDRNet-23-slim and DDRNet-23 under different scenes.

\subsubsection{CamVid}

\begin{table}[]
\caption{Accuracy and speed comparison on CamVid. MSFNet runs at 1024$\times$768 and MSFNet* runs at 768$\times$512 while other methods run at 960$\times$720. The corresponding speed is measured using TensorRT acceleration if the method is marked with \dag }
\label{tab:2}
\begin{tabular}{p{80pt}p{33pt}<{\centering}p{45pt}<{\centering}p{45pt}<{\centering}}
\toprule
Model                               & MIoU                 & Speed (FPS)           & GPU                  \\ \midrule
\multicolumn{4}{c}{w/o Cityscapes pre-training}                                                             \\ \midrule
DFANet A\cite{li2019dfanet}         & 64.7                 & 120                  & TitanX               \\
DFANet B\cite{li2019dfanet}         & 59.3                 & 160                  & TitanX               \\
SwiftNetRN-18 pyr\cite{orsic2019defense} & 73.9            & -                    & GTX 1080Ti           \\
SwiftNetRN-18\cite{orsic2019defense}     & 72.6            & -                    & GTX 1080Ti           \\
BiSeNet1\cite{yu2018bisenet}        & 65.6                 & 175                  & GTX 1080Ti           \\
BiSeNet2\cite{yu2018bisenet}        & 68.7                 & 116                  & GTX 1080Ti           \\
BiSeNetV2\dag\cite{yu2020bisenet}   & 72.4                 & 124                  & GTX 1080Ti           \\
BiSeNetV2-L\dag\cite{yu2020bisenet} & 73.2                 & 33                   & GTX 1080Ti           \\
CAS\cite{zhang2019customizable}     & 71.2                 & 169                  & TitanXp               \\
GAS\cite{lin2020graph}              & 72.8                 & 153                  & TitanXp               \\
SFNet(DF2)\cite{li2020semantic}     & 70.4                 & 134                  & GTX 1080Ti           \\
SFNet(ResNet-18)\cite{li2020semantic}& 73.8                & 36                   & GTX 1080Ti           \\
MSFNet*\cite{si2019real}            & 72.7                 & 160                  & GTX 2080Ti           \\
MSFNet\cite{si2019real}             & 75.4                 & 91                   & GTX 2080Ti           \\ \midrule
DDRNet-23-slim                      & 74.7(74.3$\pm$0.3)      & $\textbf{230}$    & GTX 2080Ti         \\
DDRNet-23                           & $\textbf{76.3}$(76.0$\pm$0.3)& 94      & GTX 2080Ti         \\ \midrule
\multicolumn{4}{c}{w/ Cityscapes pre-training}                                                              \\ \midrule
VideoGCRF\cite{chandra2018deep}     & 75.2                 & -                    & -                   \\
CCNet3D\cite{9133304}               & 79.1                 & -                    & -                   \\
BiSeNetV2\dag\cite{yu2020bisenet}   & 76.7                 & 124                  & GTX 1080Ti           \\
BiSeNetV2-L\dag\cite{yu2020bisenet} & 78.5                 & 33                   & GTX 1080Ti           \\ \midrule
DDRNet-23-slim                      & 78.6(78.2$\pm$0.3)     & $\textbf{230}$    & GTX 2080Ti           \\
DDRNet-23                           & $\textbf{80.6}$(80.1$\pm$0.4)& 94           & GTX 2080Ti           \\ \bottomrule
\end{tabular}
\end{table}

\begin{figure}
\centering
\begin{minipage}{1.1in}
\includegraphics[width=1.1in]{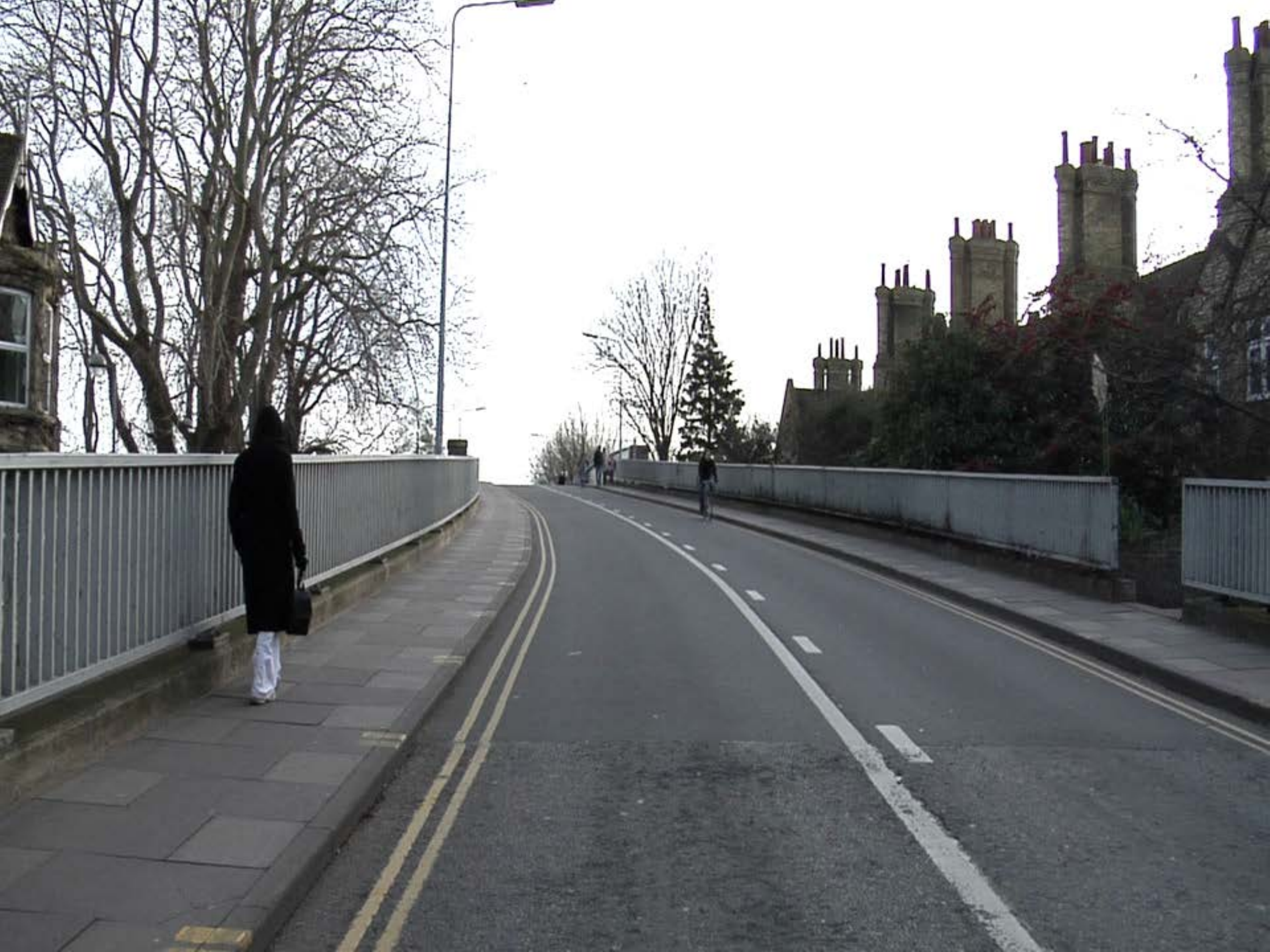}
\end{minipage}
\begin{minipage}{1.1in}
\includegraphics[width=1.1in]{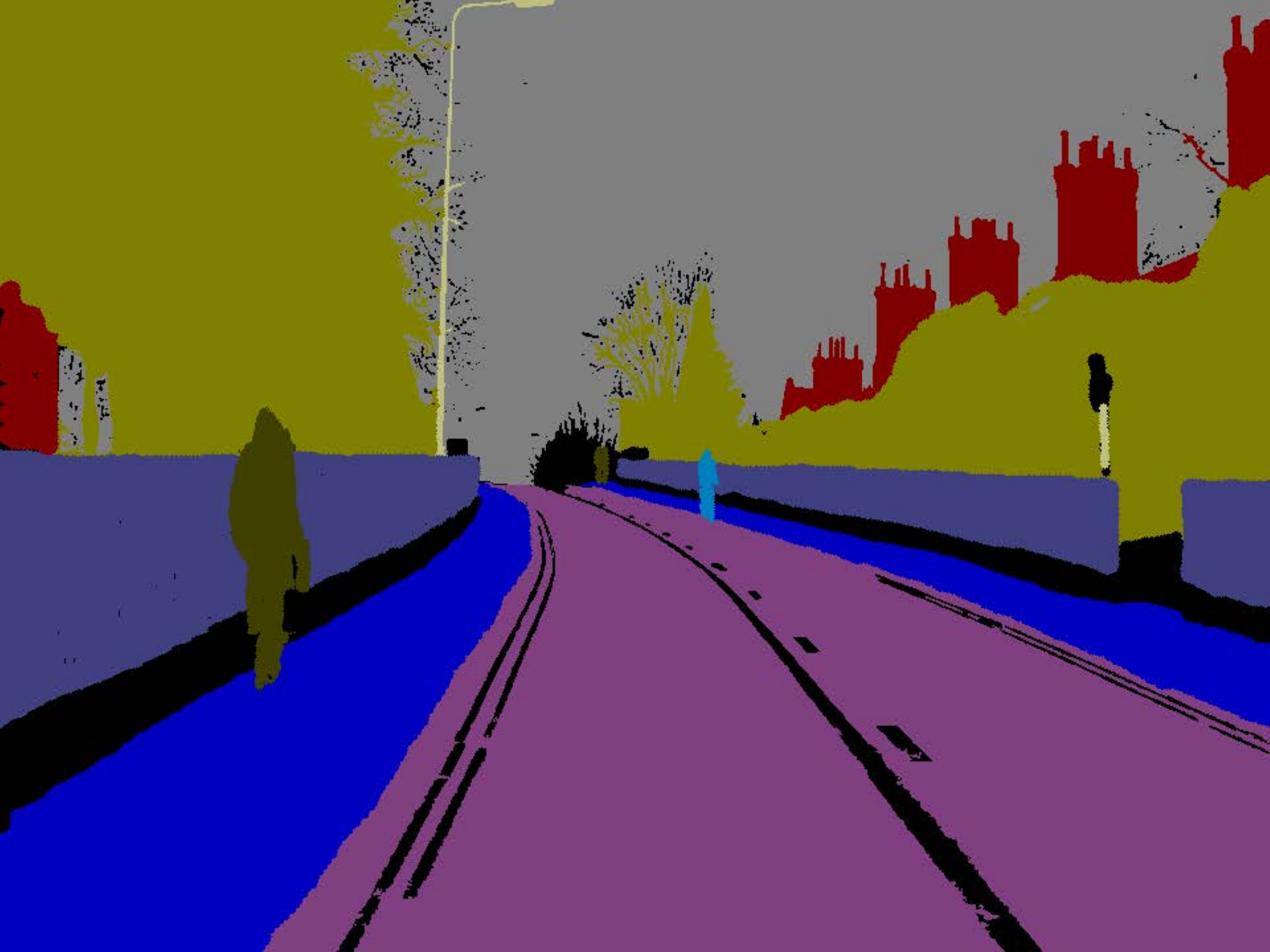}
\end{minipage}
\vspace{0.3em}
\begin{minipage}{1.1in}
\includegraphics[width=1.1in]{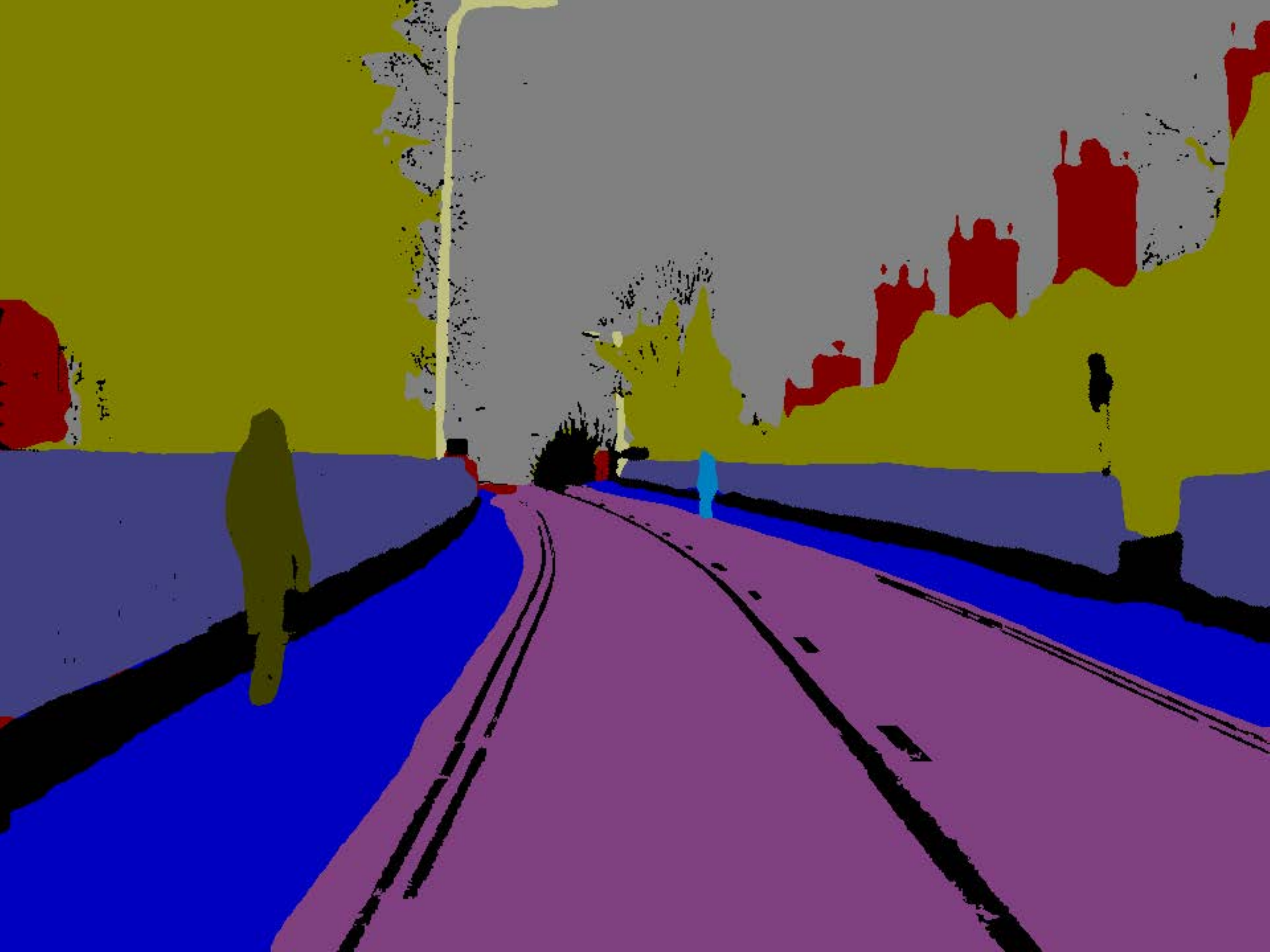}
\end{minipage}
\vspace{0.3em}
\begin{minipage}{1.1in}
\includegraphics[width=1.1in]{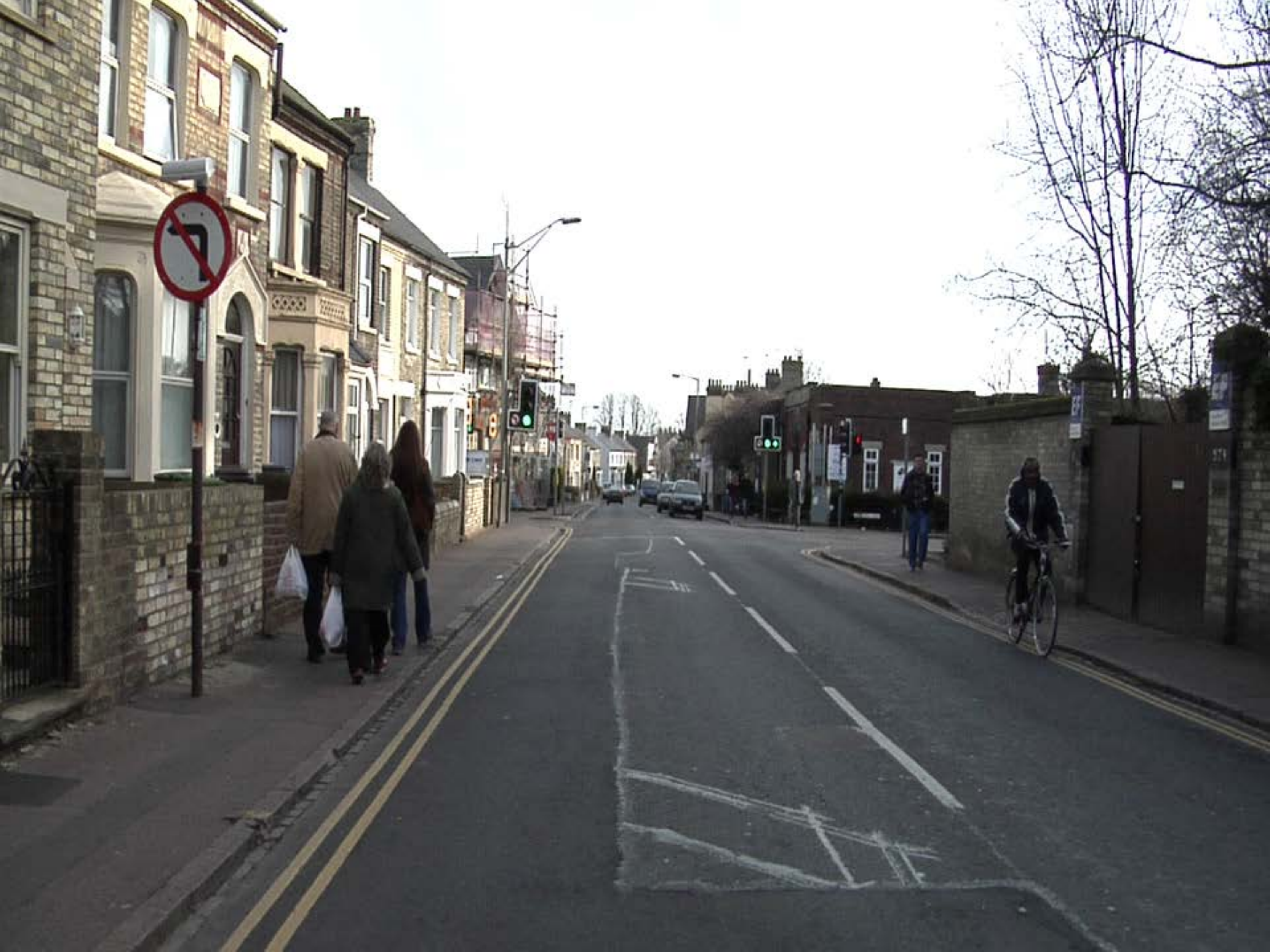}
\end{minipage}
\begin{minipage}{1.1in}
\includegraphics[width=1.1in]{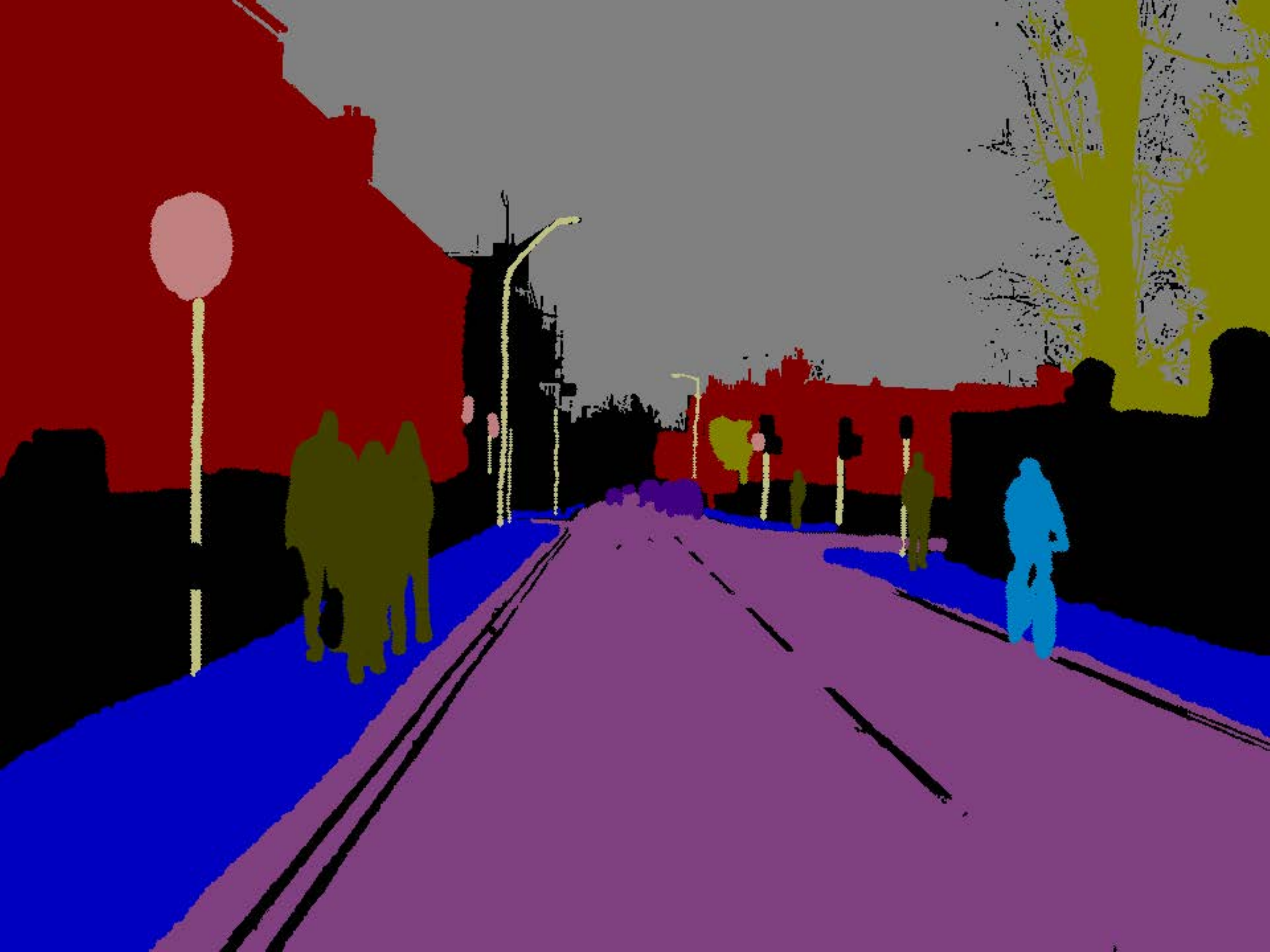}
\end{minipage}
\begin{minipage}{1.1in}
\includegraphics[width=1.1in]{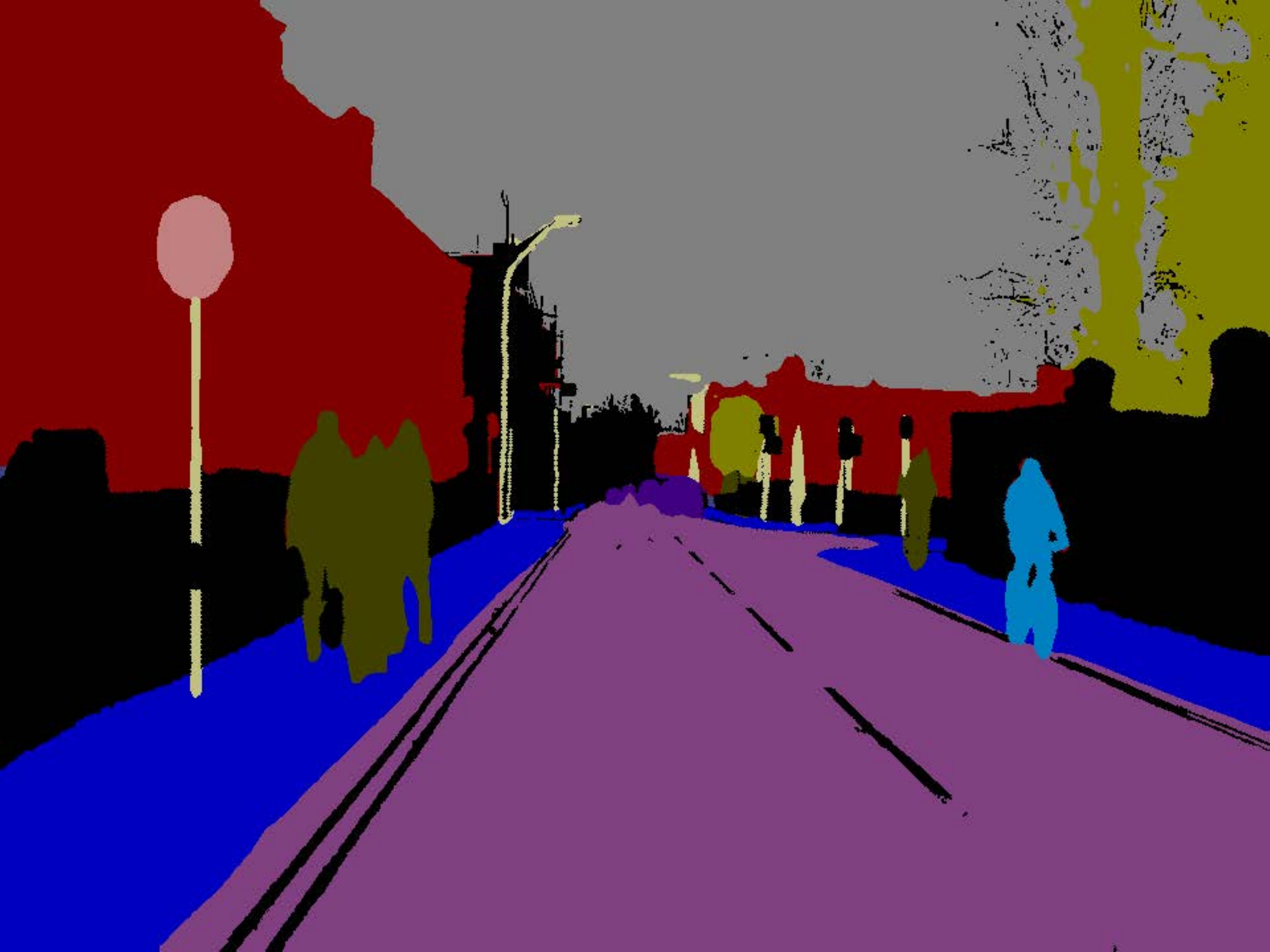}
\end{minipage}
\vspace{0.3em}
\begin{minipage}{1.1in}
\includegraphics[width=1.1in]{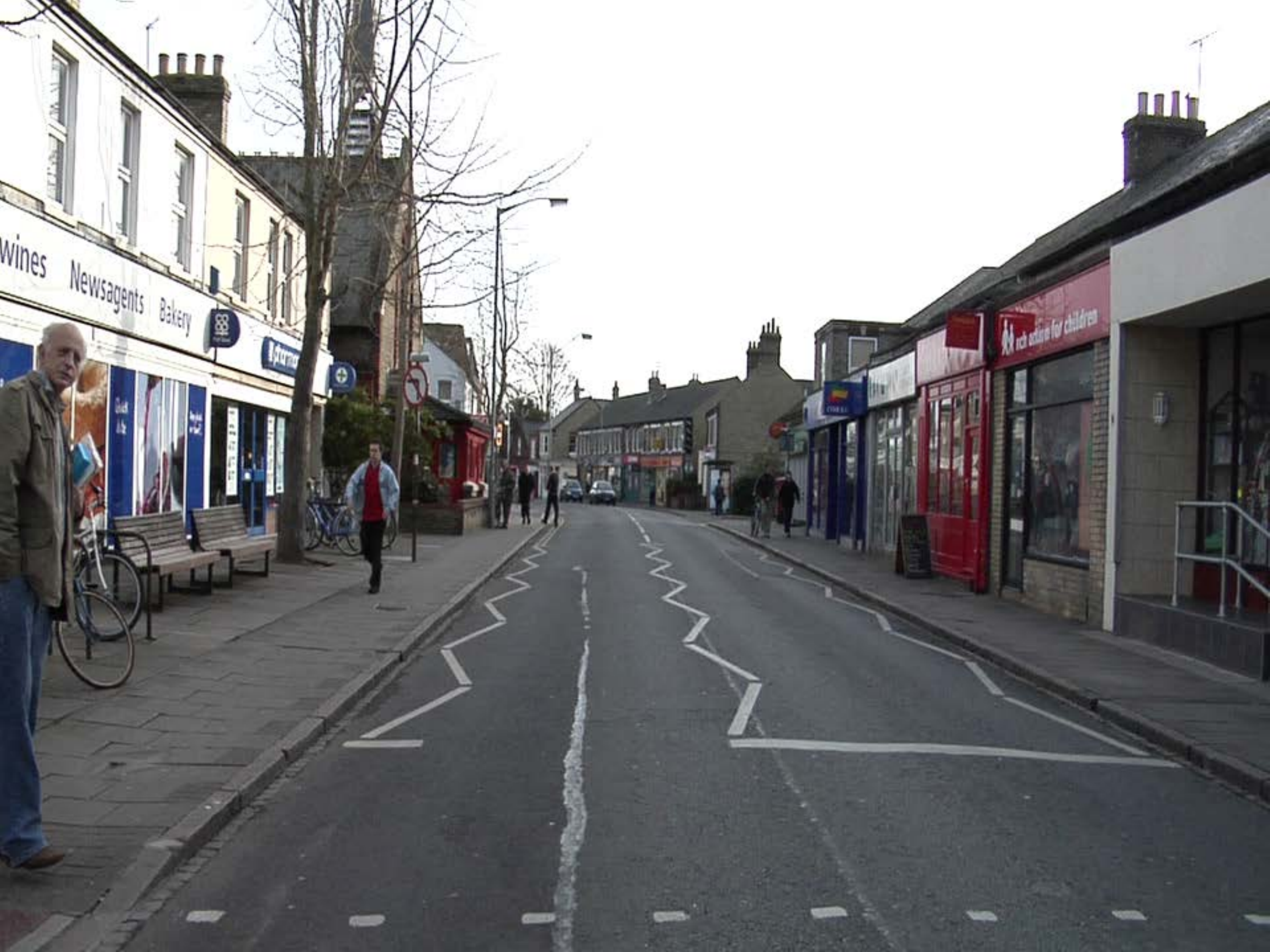}
\end{minipage}
\begin{minipage}{1.1in}
\includegraphics[width=1.1in]{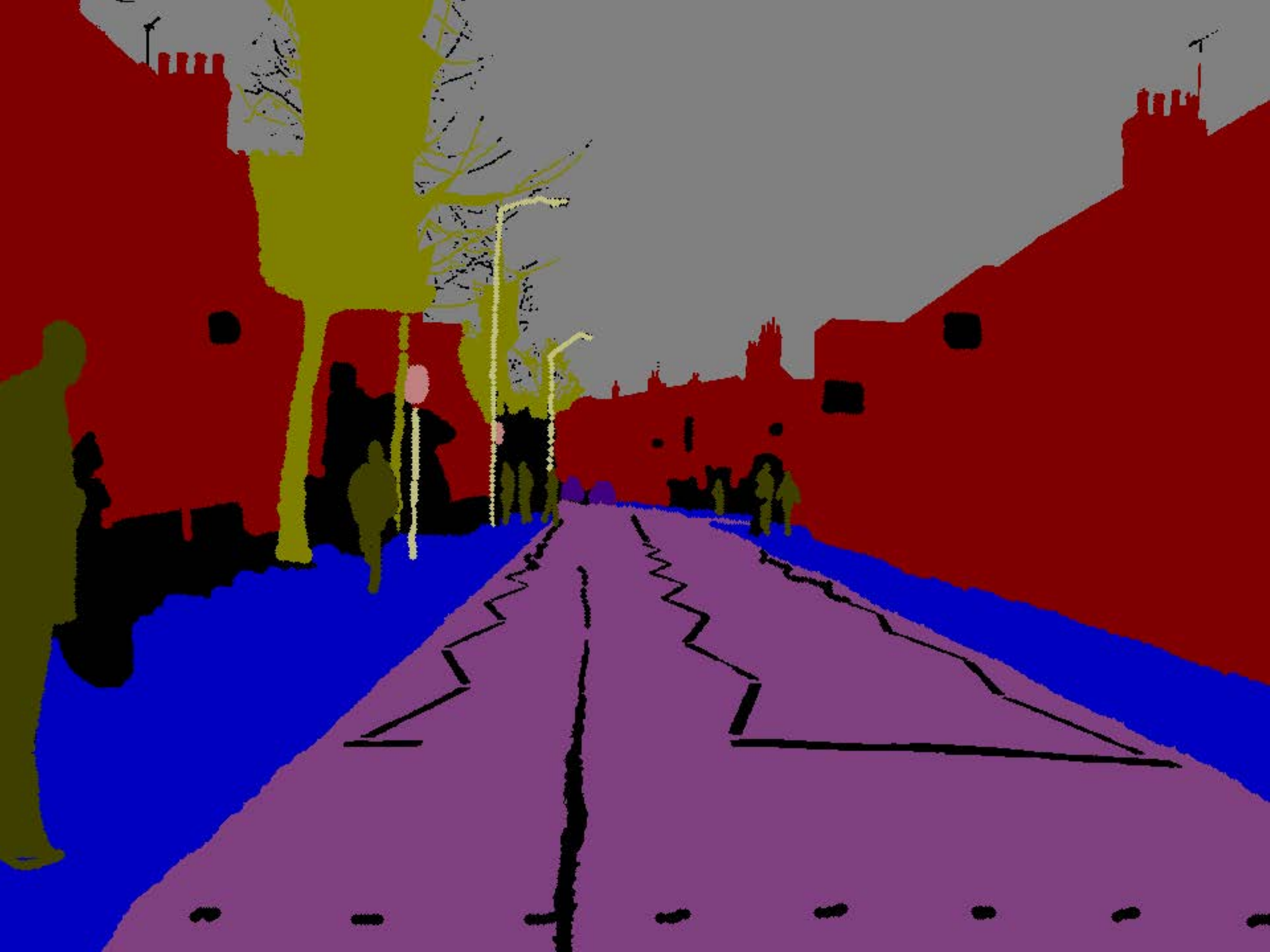}
\end{minipage}
\begin{minipage}{1.1in}
\includegraphics[width=1.1in]{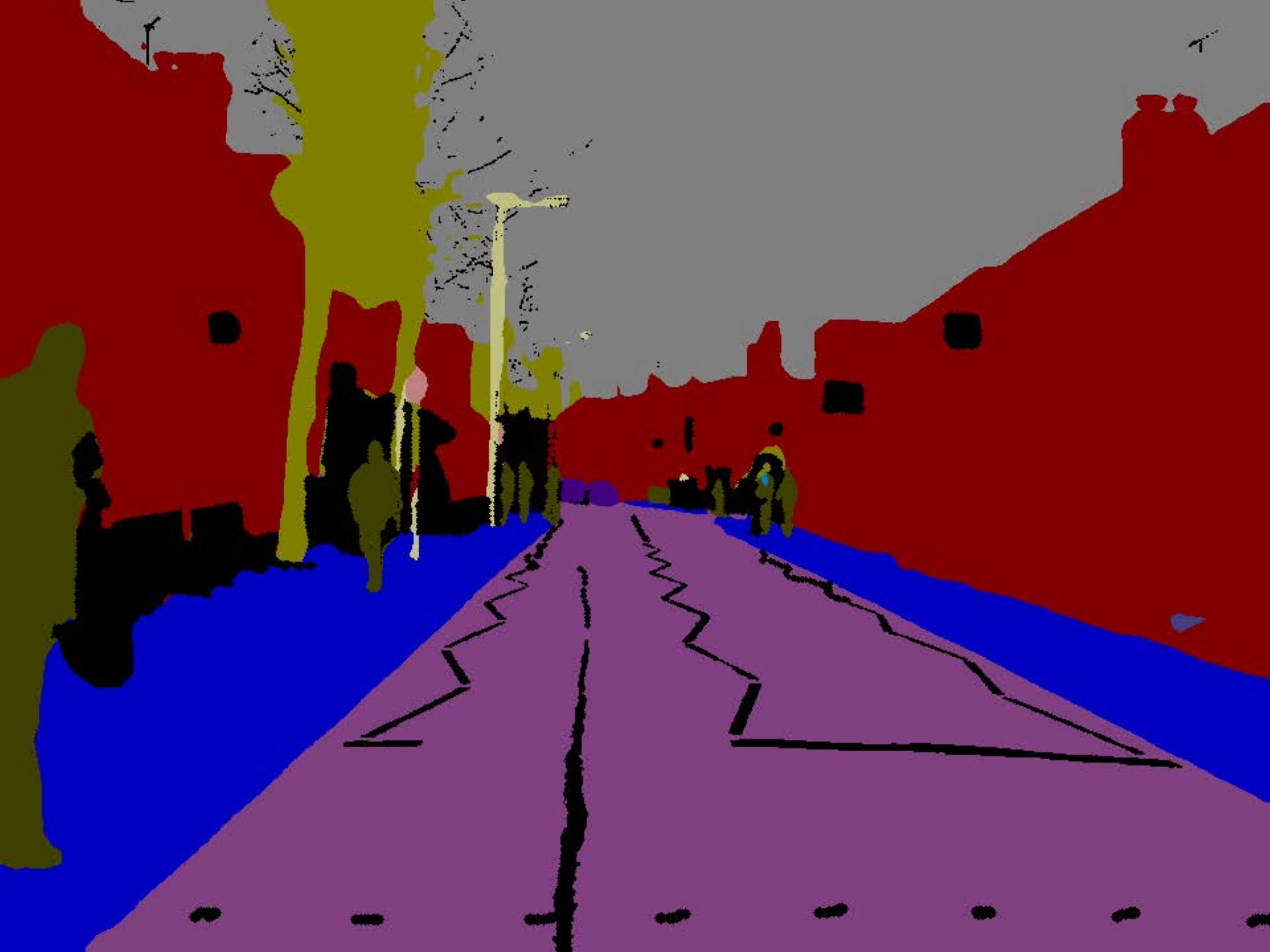}
\end{minipage}
\vspace{0.3em}
\begin{minipage}{1.1in}
\includegraphics[width=1.1in]{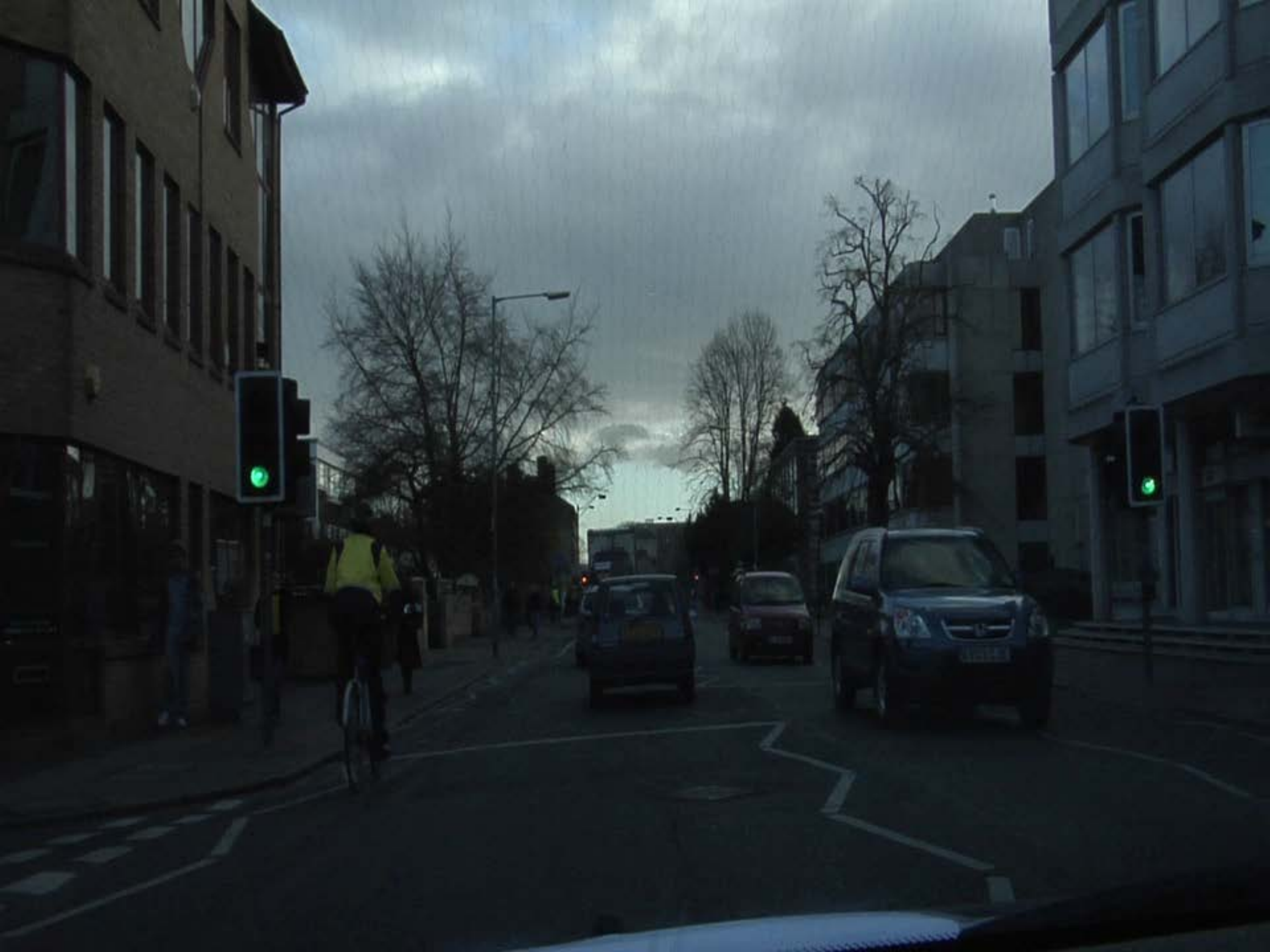}
\end{minipage}
\begin{minipage}{1.1in}
\includegraphics[width=1.1in]{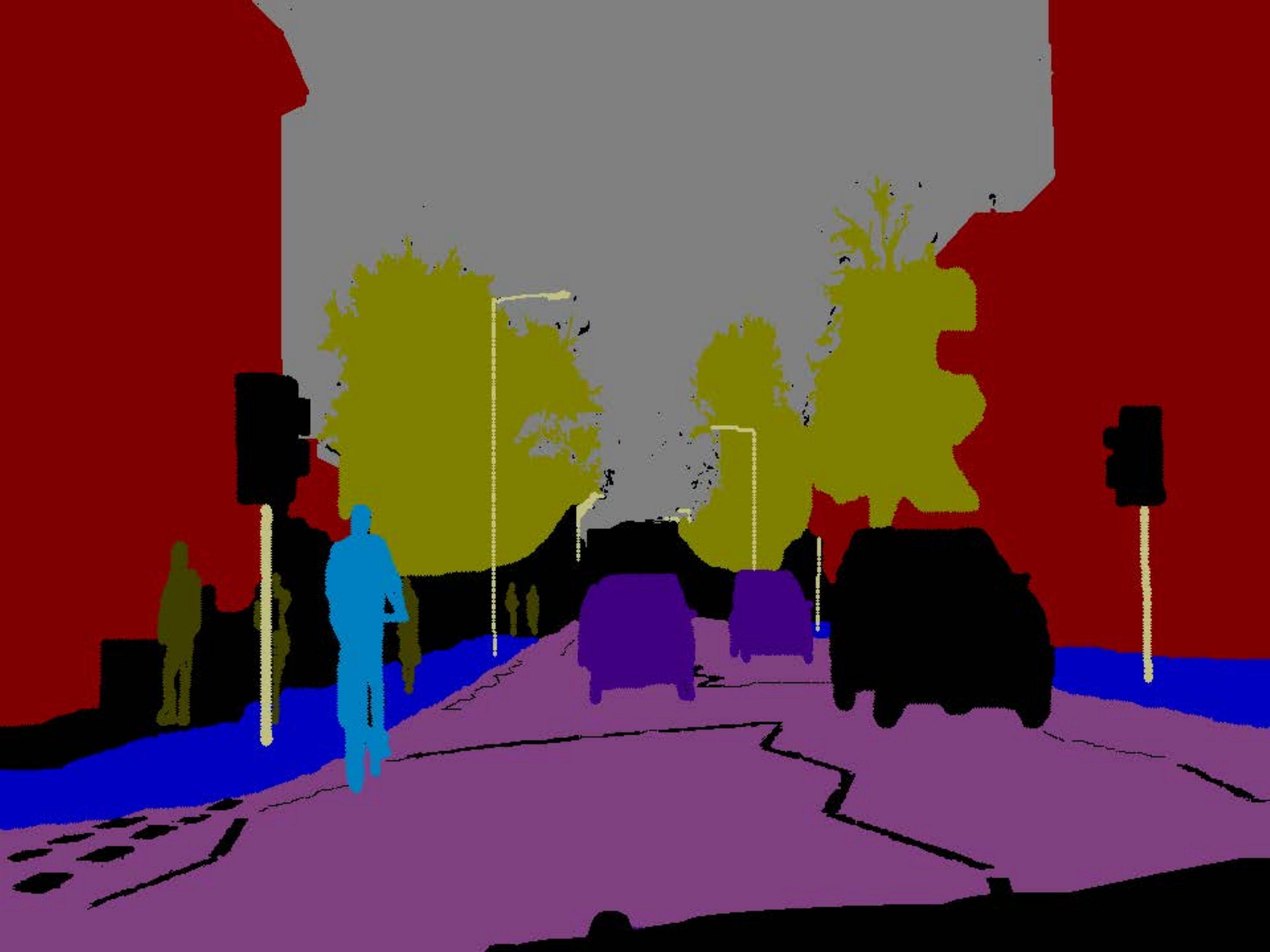}
\end{minipage}
\begin{minipage}{1.1in}
\includegraphics[width=1.1in]{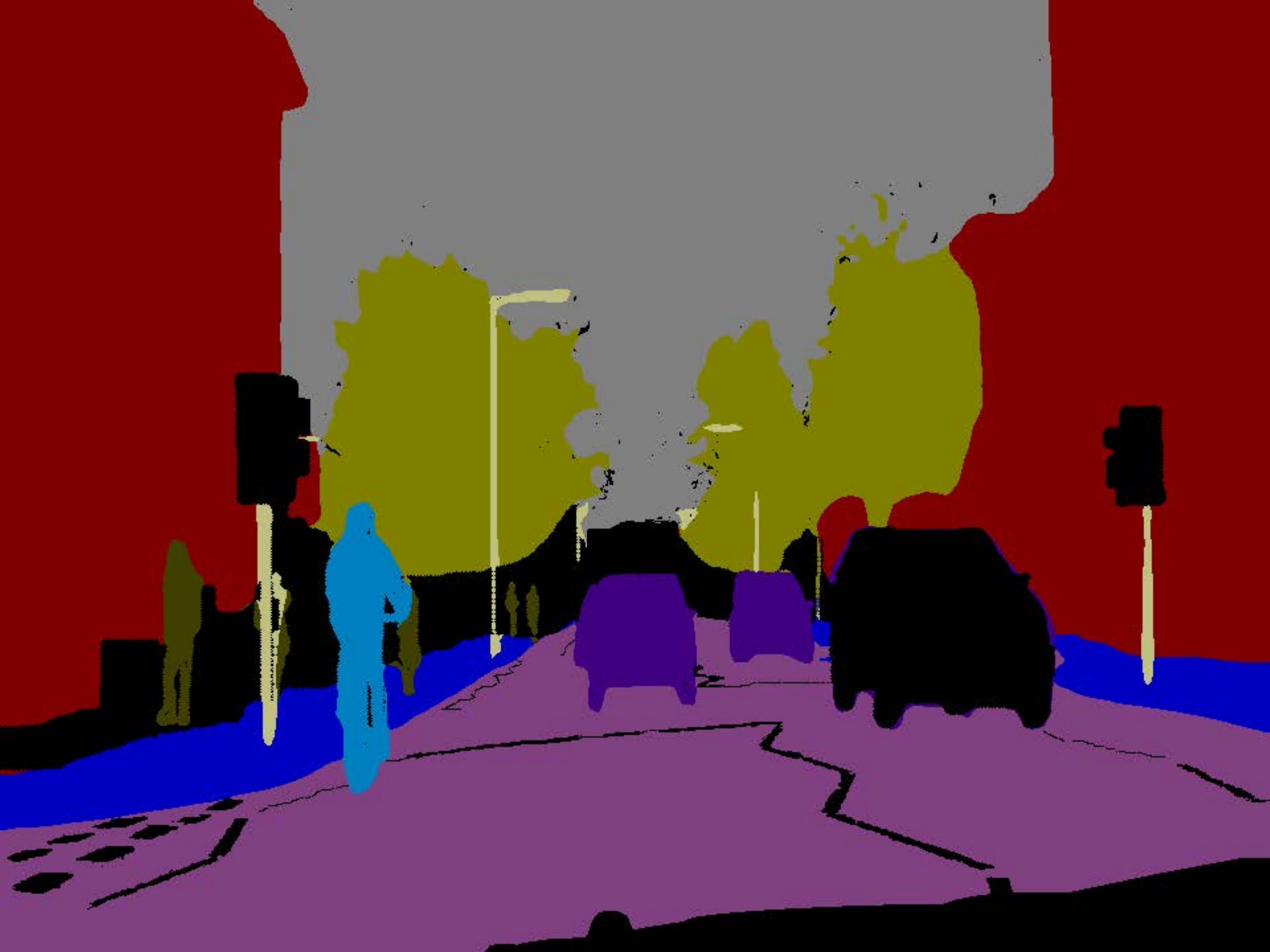}
\end{minipage}
\vspace{0.3em}
\begin{minipage}{1.1in}
\includegraphics[width=1.1in]{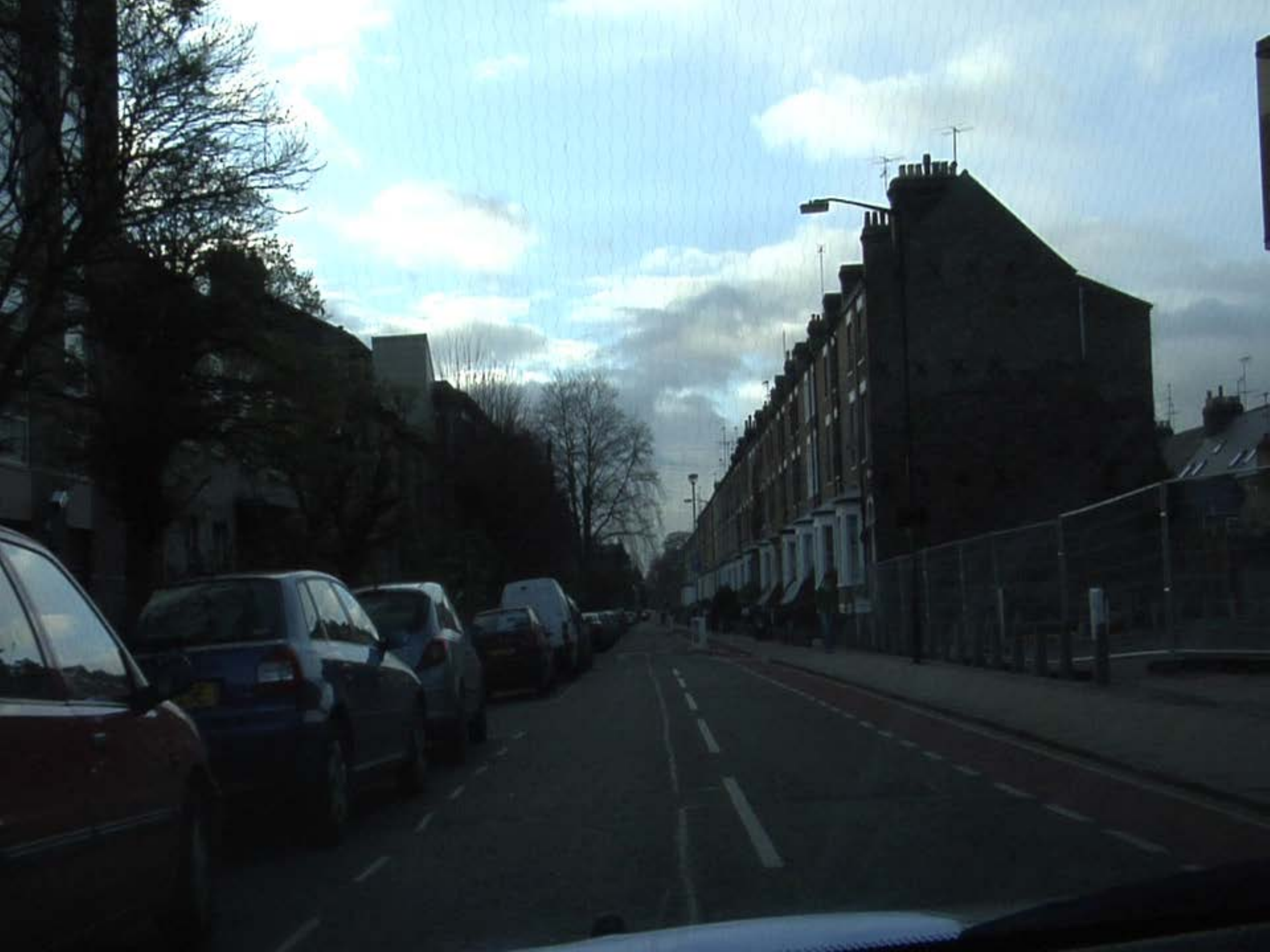}
\end{minipage}
\begin{minipage}{1.1in}
\includegraphics[width=1.1in]{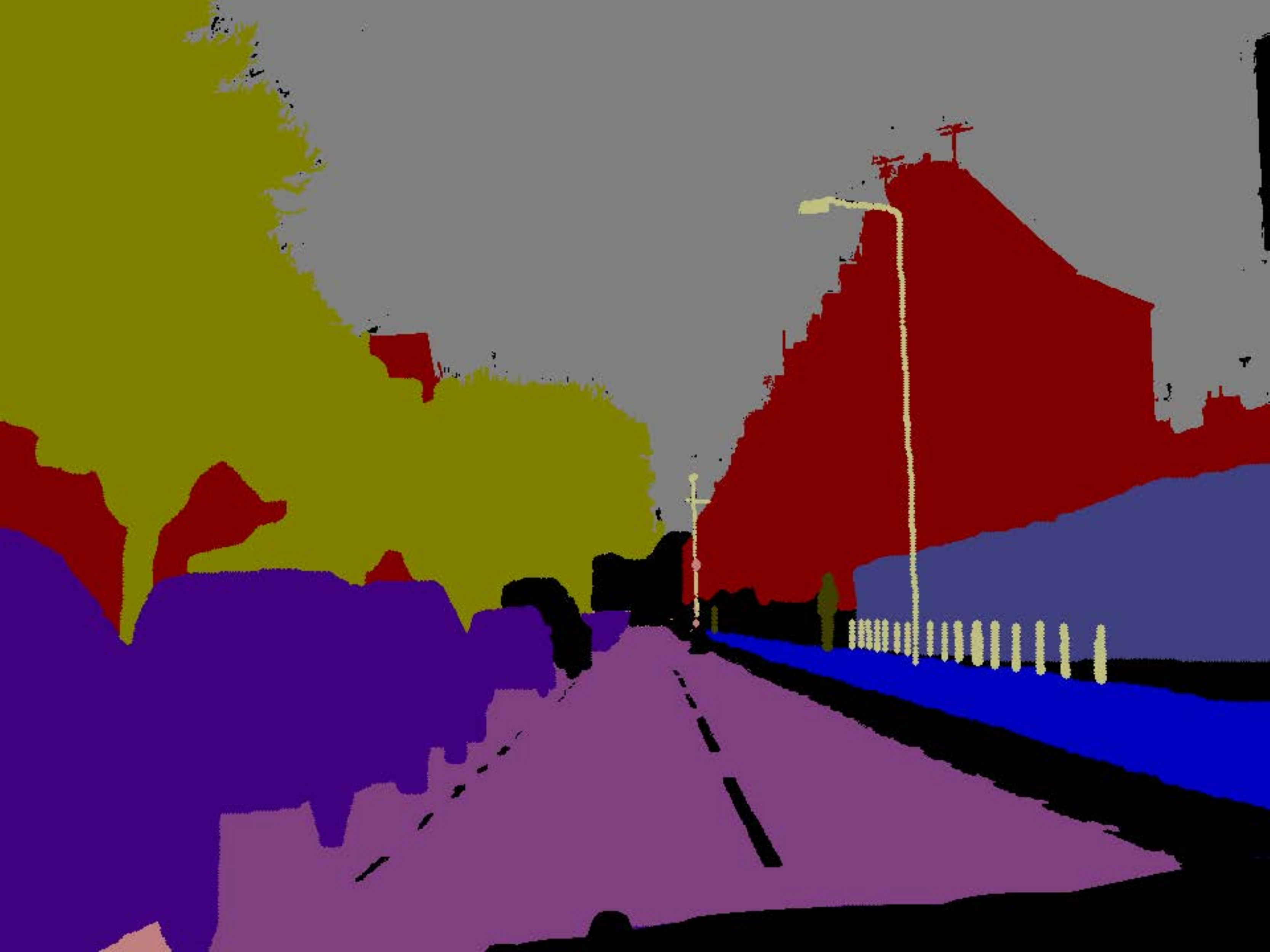}
\end{minipage}
\begin{minipage}{1.1in}
\includegraphics[width=1.1in]{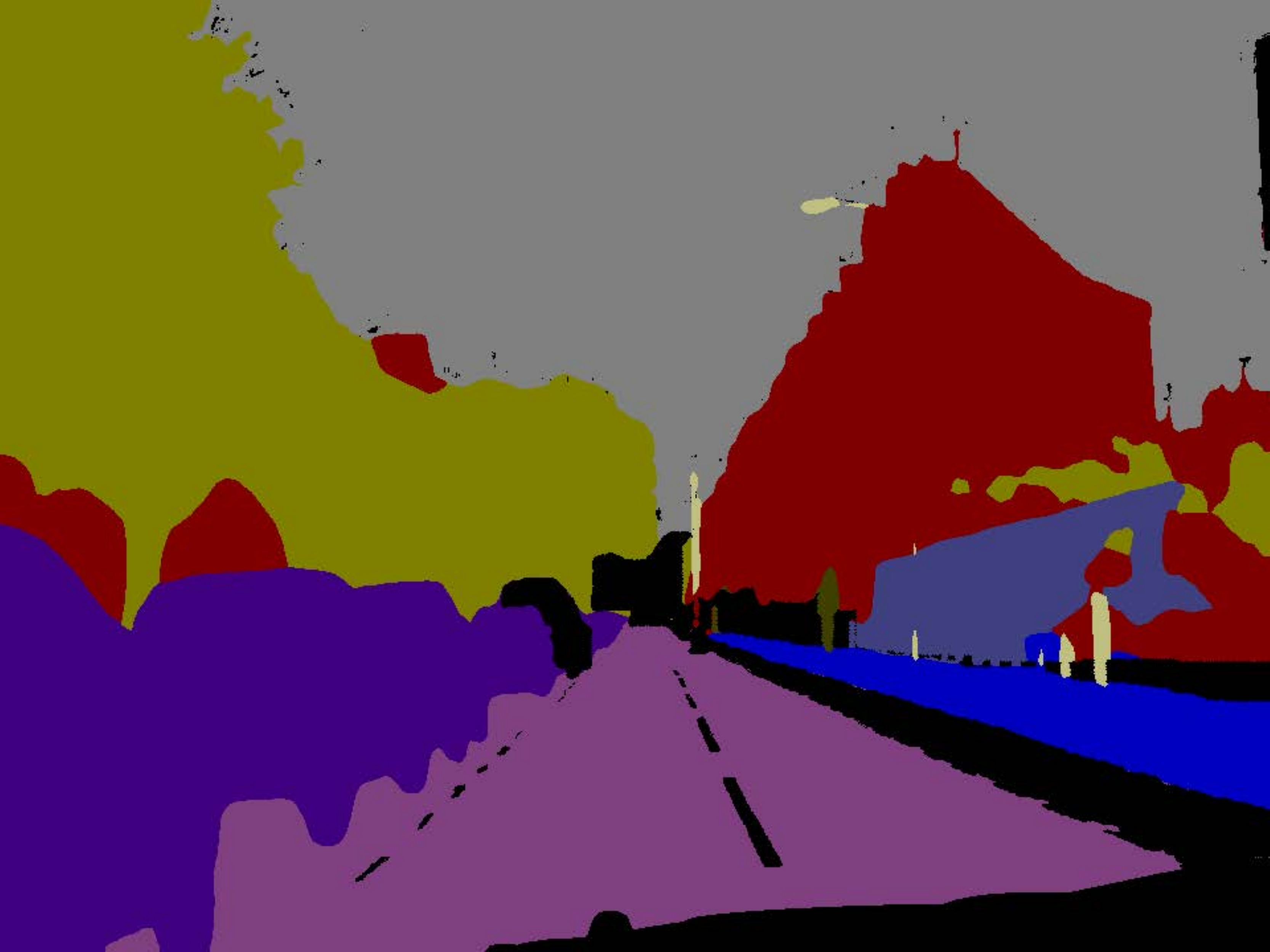}
\end{minipage}
\vspace{0.3em}
\begin{minipage}{1.1in}
\includegraphics[width=1.1in]{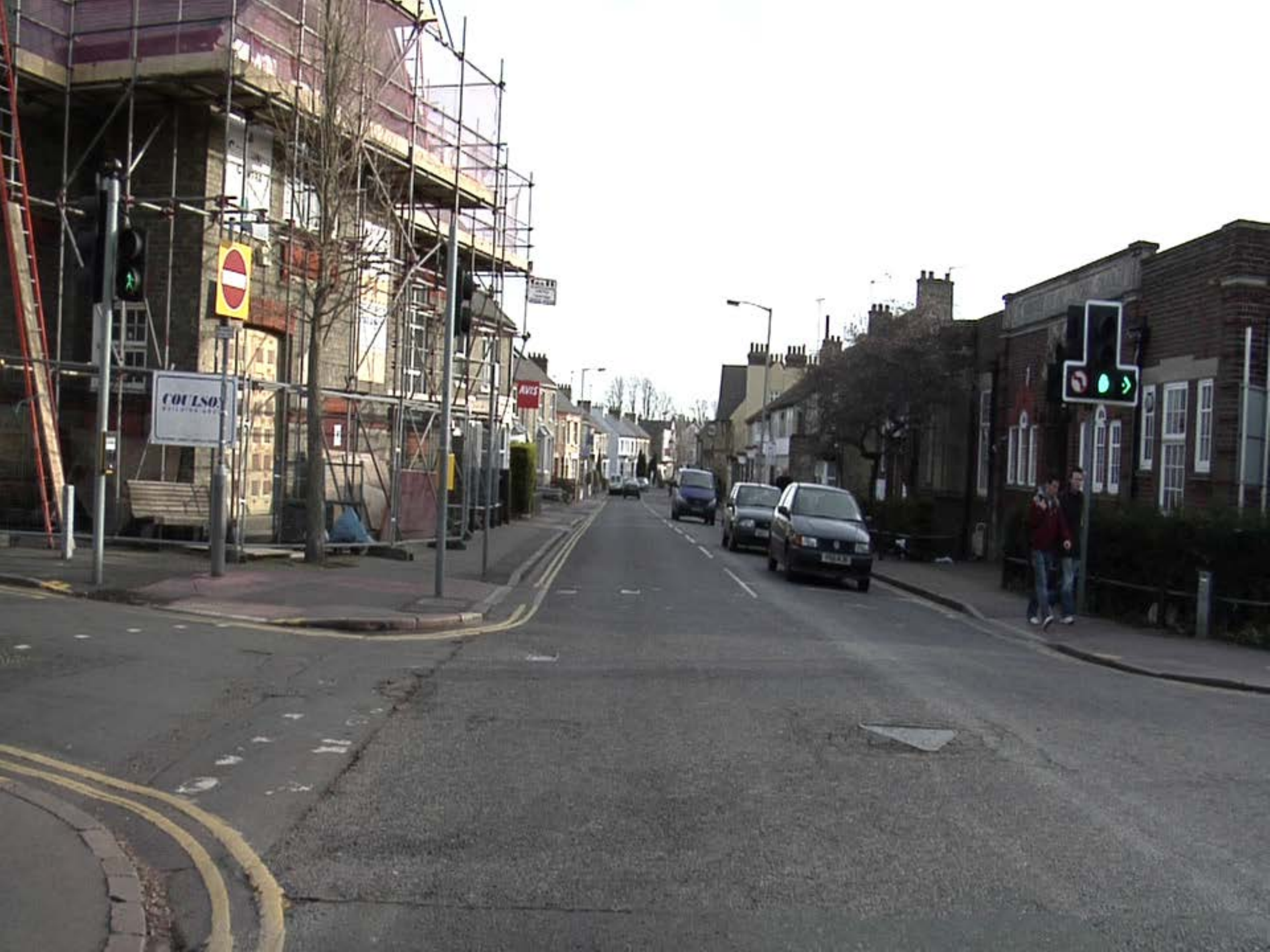}
\end{minipage}
\begin{minipage}{1.1in}
\includegraphics[width=1.1in]{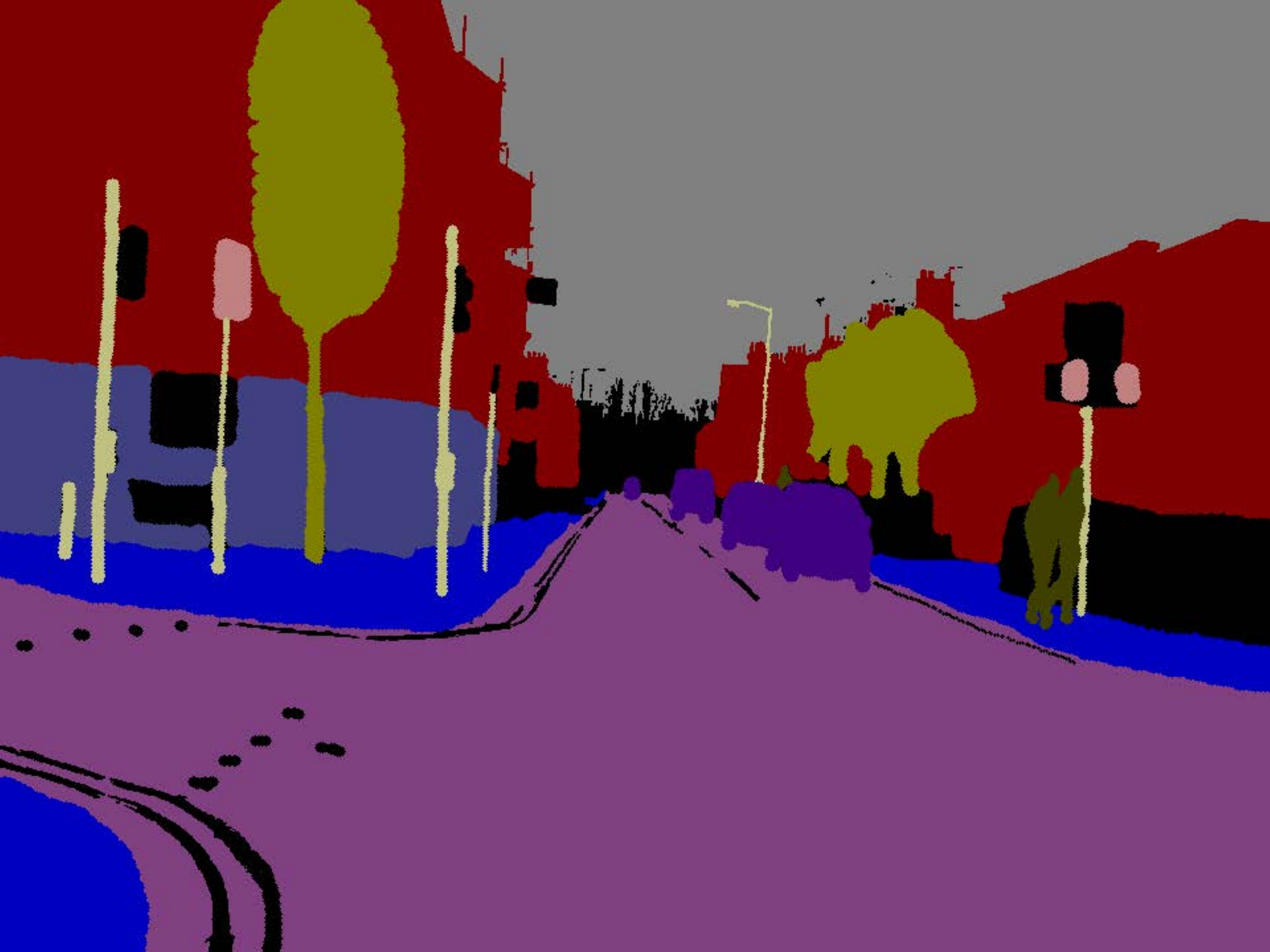}
\end{minipage}
\begin{minipage}{1.1in}
\includegraphics[width=1.1in]{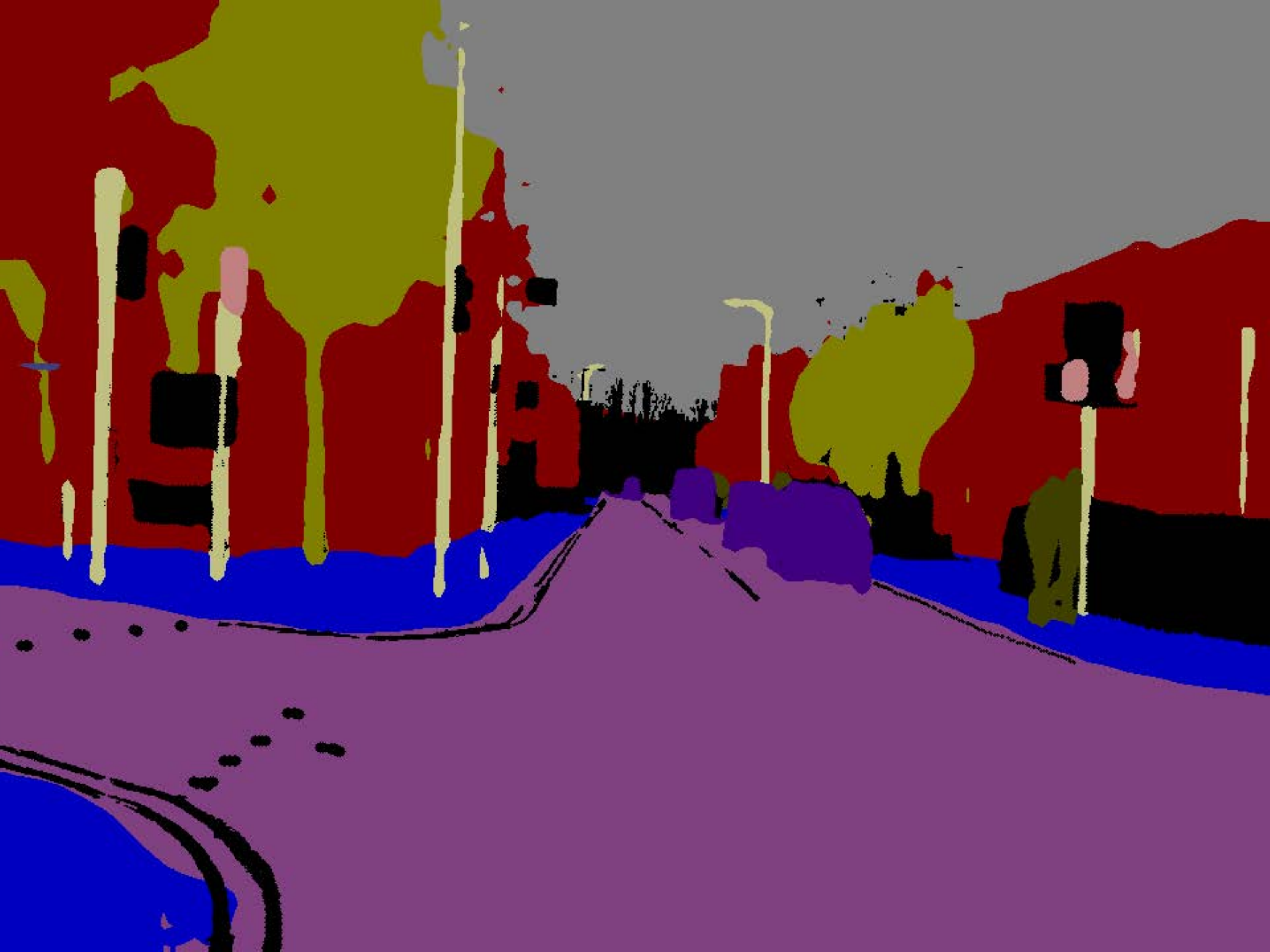}
\end{minipage}
\caption{Visualized segmentation results on CamVid test set. The color of ignored labels during testing is set to black. The three columns left-to-right refer to the input image, the ground truth, and the output of DDRNet-23. The first four rows show the successful samples while the last two rows represent some segmentation failures.}
\label{fig8}
\end{figure}

As shown in Table \ref{tab:2}, DDRNet-23-slim achieves 74.7$\%$ mIoU on the CamVid test set at 230 FPS without Cityscapes pre-training. It obtains the second-highest accuracy and runs faster than all the other methods. In particular, the performance of DDRNet-23 is superior to MSFNet, the previous state-of-the-art method. DDRNet-23 also has a big performance gain over BiSeNetV2-L and SFNet (ResNet-18) and runs about two times faster than them. Given that the training pixels of CamVid are much less than that of Cityscapes, we believe that the outstanding performances of DDRNets partly attribute to adequate ImageNet pre-training. In addition, our Cityscapes pre-trained models achieve new state-of-the-art accuracy with the real-time speed. Specially, Cityscapes pre-trained DDRNet-23 realizes 80.6$\%$ mIoU with 94 FPS, stronger and much faster than BiSeNetV2-L. The corresponding visualized results are shown in Fig. \ref{fig8}.

\subsubsection{COCOStuff}

We also validate our method on COCOStuff which is a more challenging dataset for real-time semantic segmentation due to the plentiful categories. The stride of RBB in the low-resolution branch is set to 1, for the image resolution is smaller than the other two datasets. Time to reshape images and predicted masks is not counted. Table \ref{tab:12} demonstrates that our method outperforms BiSeNetV2 by a substantial degree under very challenging scenarios. Our DDRNet-23 achieves a similar accuracy with PSPNet50 while running 20 times as fast as it.

\begin{table}[]
\caption{Accuracy and speed comparison on COCOStuff. The input resolution is 640$\times$640 and the result of PSPNet50 comes from \cite{yu2020bisenet}. The corresponding speed is measured using TensorRT acceleration if the method is marked with \dag} 
\label{tab:12}
\begin{tabular}{p{65pt}p{60pt}<{\centering}p{30pt}<{\centering}p{48pt}<{\centering}}
\toprule
Model                               & MIoU              & PixAcc               & Speed (FPS)             \\ \midrule
PSPNet50\cite{zhao2017pyramid}      & 32.6              & -                    & 6.6                         \\ \midrule
ICNet\cite{zhao2018icnet}           & 29.1              & -                    & 35.7                    \\ \midrule
BiSeNetV2\dag\cite{yu2020bisenet}   & 25.2              & 60.5                 & 87.9                    \\
BiSeNetV2-L\dag\cite{yu2020bisenet} & 28.7              & 63.5                 & 42.5                  \\ \midrule
DDRNet-23                           & 32.1(31.8$\pm$0.2)& 64.7(64.7$\pm$0.1)   & $\textbf{129.2}$          \\
DDRNet-39                           & $\textbf{34.8}$(34.6$\pm$0.1)& $\textbf{66.6}$(66.7$\pm$0.2)   &83.8        \\  \bottomrule
\end{tabular}
\end{table}

\subsection{Comparisons with State-of-the-art Results}

\begin{table}[]
\caption{State-of-the-art models on Cityscapes test set. OS denotes the final output stride. All the methods train models on both train and val set except PSPNet marked with \dag only using train set. GFLOPs calculation adopts 1024$\times$1024 image as input and most of results about GFLOPs and Params can be found in \cite{li2020semantic}}
\label{tab:9}
\begin{tabular}{p{80pt}p{15pt}<{\centering}p{32pt}<{\centering}p{32pt}<{\centering}p{32pt}<{\centering}}
\toprule
Model                               & OS                & mIoU                 & GFLOPs            & Params.     \\ \midrule
SAC\cite{zhang2017scale}            & 8                 & 78.1                 & -                 & -           \\
DepthSeg\cite{kong2018recurrent}    & 8                 & 78.2                 & -                 & -           \\
PSPNet\dag\cite{zhao2017pyramid}    & 8                 & 78.4                 & 1065.4            & 65.7M  \\
ResNet38\cite{wu2019wider}          & 8                 & 78.4                 & -                 & -      \\
BiSeNet\cite{yu2018bisenet}         & 8                 & 78.9                 & 219.1             & 51.0M  \\
DFN\cite{yu2018learning}            & 8                 & 79.3                 & 1121.0            & 90.7M  \\
PSANet\cite{zhao2018psanet}         & 8                 & 80.1                 & 1182.6            & 85.6M  \\
DenseASPP\cite{yang2018denseaspp}   & 8                 & 80.6                 & 632.9             & 35.7M  \\
CCNet\cite{huang2019ccnet}          & 8                 & 81.4                 & 1153.9            & 66.5M  \\
DANet\cite{fu2019dual}              & 8                 & 81.5                 & 1298.8            & 66.6M  \\
OCNet\cite{yuan2018ocnet}           & 8                 & 81.7                 & -                 & -      \\
OCRNet\cite{yuan2019object}         & 8                 & 81.8                 & -                 & -      \\
HRNetV2-W48\cite{9052469}           & 4                 & 81.6                 & 348.1             & 65.9M  \\
SFNet\cite{li2020semantic}          & 4                 & 81.8                 & 417.5             & 50.3M  \\ \midrule
DDRNet-39                           & 8                 & 81.9                 & $\textbf{140.6}$  & $\textbf{32.3M}$\\
DDRNet-39 1.5$\times$               & 8                 & $\textbf{82.4}$      & 303.0             & 70.2M  \\ \bottomrule
\end{tabular}
\end{table}

In this part, we further demonstrate the capacity of DDRNet for semantic segmentation by comparing it to state-of-the-art models on the Cityscapes test set. Such methods frequently employ multi-scale and horizontal flip inference to achieve better results regardless of time cost. For a fair comparison with them, we also apply multiple scales including 0.50$\times$, 0.75$\times$, 1$\times$, 1.25$\times$, 1.5$\times$, 1.75$\times$, 2$\times$ with left-right flipping during test. As is shown in Table \ref{tab:9}, standard test augmentation improves the accuracy of DDRNet-39 from 80.4$\%$ to 81.9$\%$. Our DDRNet-39 outperforms numerous powerful models which are integrated with self-attention modules such as CCNet, DANet, and OCNet. It is noteworthy that our method only requires 11$\%$ computation of DANet. DDRNet-39 also gets ahead of SFNet (based on ResNet-101 backbone) which is a state-of-the-art method for real-time semantic segmentation, only requiring 34$\%$ computation of it. DDRNet-39 1.5$\times$ of which size is closer to other models in Table \ref{tab:9} achieves a very competitive performance (82.4$\%$ mIoU).

\subsection{Comparisons with HRNet}

The major difference between DDRNet and HRNet is the number of parallel branches. Besides, we append the multi-scale context extraction module to the end of the low-resolution branch. Experimental results in Table \ref{tab:10} demonstrate the improvement of DDRNet over HRNet in both inference time and training memory usage. We get the val results of two small HRNets from the official implementation. Training memory is measured on a single 2080Ti with a batch size of 2 and a crop size of 1024$\times$512, excluding the auxiliary segmentation head.

\begin{table}[]
\caption{Comparative experiments between DDRNet and HRNet in terms of mIoU, FPS and train memory}
\label{tab:10}
\begin{tabular}{p{100pt}p{30pt}<{\centering}p{30pt}<{\centering}p{43pt}<{\centering}}
\toprule
Model                                & mIoU              & FPS               & Train mem.  \\ \midrule
HRNetV2-W18-Small-v1\cite{9052469}   & 70.3              & 67.2              & 1989MiB     \\
HRNetV2-W18-Small-v2\cite{9052469}   & 76.2              & 31.1              & 2745MiB     \\
DDRNet-23-slim                       & $\textbf{76.9}$   & $\textbf{101.6}$  & $\textbf{1629MiB}$     \\ \bottomrule
\end{tabular}
\end{table}

\subsection{Ablative Experiments on Cityscapes}

\subsubsection{Standard Bells and Whistles}

\begin{table}[]
\caption{Influences of standard bells and whistles, including deep supervision (DS), OHEM and training at a crop size of 1024$\times$1024 (the default is 1024$\times$512)}
\label{tab:8}
\begin{tabular}{p{60pt}p{33pt}<{\centering}p{33pt}<{\centering}p{33pt}<{\centering}p{33pt}<{\centering}}
\toprule
Model                    & DS                          & OHEM          & 1024$\times$1024 & mIoU       \\ \midrule
DDRNet-23-slim           &                             &               &                  & 76.1       \\
DDRNet-23-slim           & \checkmark                  &               &                  & 76.1       \\
DDRNet-23-slim           & \checkmark                  &\checkmark     &                  & 76.9       \\
DDRNet-23-slim           & \checkmark                  &\checkmark     & \checkmark       & 77.8       \\ \bottomrule
\end{tabular}
\end{table}

We analyze the effect of some basic training tricks which are also adopted by recent advanced method SFNet\cite{li2020semantic}. As shown in Table \ref{tab:8}, the accuracy is raised from 76.1 to 77.8 with deep supervision, OHEM, and training at a larger crop size.

\subsubsection{DAPPM}

\begin{table}[]
\caption{Comparison of DAPPM and other context extraction modules. RES2 denotes the Res2Net module and Base-OC is the object context module proposed in \cite{yuan2018ocnet}}
\label{tab:7}
\begin{tabular}{p{30pt}<{\centering}p{30pt}<{\centering}p{30pt}<{\centering}p{30pt}<{\centering}p{30pt}<{\centering}p{30pt}<{\centering}}
\toprule
 PPM                  & RES2        &Base-OC     & DAPPM      & mIoU        & Speed\\ \midrule
                      &             &            &            & 74.1        & 107.9 \\
 \checkmark           &             &            &            & 76.8        & 104.9 \\
                      &\checkmark   &            &            & 76.8        & 103.6 \\
                      &             &\checkmark  &            & 75.6        & 104.9 \\
                      &             &            &\checkmark  & 77.8        & 101.6 \\ \bottomrule
\end{tabular}
\end{table}

We compare the DAPPM with the pyramid pooling based methods (PPM), self-attention based modules (Base-OC), and the res2net module. The results in Table \ref{tab:7} shows that the proposed module improves the performance of scene parsing from 74.1$\%$ mIoU to 77.8$\%$ mIoU while the inference speed is hardly affected. Compared to the PPM and RES2, DAPPM also achieves 1$\%$ mIoU gain while the Base-OC, another state-of-the-art method, gets a relatively poor performance with low-resolution feature maps.

\subsubsection{Dual-resolution Networks}

For faster experiments, we train all the bilateral networks from scratch with a initial learning rate of 0.05, a crop size of 1024$\times$512, 600 epochs in total, and without using OHEM. As shown in Table \ref{tab:11}, using thinner detail branch results in 1.3$\%$ accuracy decrease and running much faster than the baseline. Appending the detail branch to the middle layer of the network contributes to the deep high-resolution representation and also improves the inference speed because it avoids computing with higher resolution. The bottleneck expands the feature dimension, which generates richer features for the DAPPM and the final segmentation head. The bilateral fusions further improve the segmentation accuracy at a small time cost. Finally, our dual-resolution network takes less computation resources and time than the baseline while achieving better performance. 

\begin{table}[]
\caption{Ablation study of dual-resolution networks. The baseline is adapted from BiSeNetV2 by replacing the complicated semantic branch with our low-solution branch. `+Thiner detail branch' represents cutting the dimension of the detail branch in half. '+Conv3' represents appending the detail branch to the end of conv3 stage. `+Residual' denotes replacing the 3$\times$3 convolutions with the residual basic blocks. `+Bottleneck' denotes adding a bottleneck block to the end of each branch. `+Low-to-high fusion' or `+Bilateral fusion' denotes performing the multiple low-to-high fusion or bilateral fusion}
\label{tab:11}
\begin{tabular}{p{72pt}p{30pt}<{\centering}p{30pt}<{\centering}p{30pt}<{\centering}p{30pt}<{\centering}}
\toprule
Model                     & mIoU                & Params.              & GFLOPs      & Speed\\ \midrule
Baseline                  & 72.2                & 4.3M                & 70.2        & 60.2  \\
+Thiner detail branch     & 70.9                & 3.8M                & 34.0        & 103.7 \\
+Conv3                    & 71.4                & 4.0M                & 31.7        & 128.7 \\
+Residual                 & 71.2                & 4.0M                & 31.7        & 125.2 \\
+Bottleneck               & 73.3                & 5.2M                & 34.4        & 110.2 \\
+Low-to-high fusion       & 74.0                & 5.3M                & 34.5        & 107.6 \\
+Bilateral fusion         & 74.6                & 5.7M                & 36.3        & 101.6 \\ \bottomrule
\end{tabular}
\end{table}

\section{Conclusion}
In this paper, we are devoted to the real-time and accurate semantic segmentation of road scenes and present a simple solution for it without using extra bells or whistles. In particular, novel deep dual-resolution networks are proposed as efficient backbones for real-time semantic segmentation. And a new module is designed for extracting multi-scale contextual information from low-resolution feature maps. To our best knowledge, we are the first to introduce deep high-resolution representation into real-time semantic segmentation and our simple strategy outperforms all previous real-time models on three popular benchmarks. DDRNets mainly consist of residual basic blocks and bottleneck blocks, providing a wide range of speed and accuracy trade-off by scaling model width and depth. Due to the simplicity and efficiency of our method, it can be seen as a strong baseline for unifying real-time and high-accuracy semantic segmentation. Further studies will focus on improving the baseline and transferring the backbones to other downstream tasks.


%
\bibliographystyle{ieeetr}
\bibliography{reference}

\begin{thebibliography}{10}

\bibitem{8006236}
Z.~{Liu}, X.~{Li}, P.~{Luo}, C.~C. {Loy}, and X.~{Tang}, ``Deep learning markov
  random field for semantic segmentation,'' {\em IEEE Transactions on Pattern
  Analysis and Machine Intelligence}, vol.~40, no.~8, pp.~1814--1828, 2018.

\bibitem{8760555}
L.~{Jing}, Y.~{Chen}, and Y.~{Tian}, ``Coarse-to-fine semantic segmentation
  from image-level labels,'' {\em IEEE Transactions on Image Processing},
  vol.~29, pp.~225--236, 2020.

\bibitem{9126262}
X.~{Ren}, S.~{Ahmad}, L.~{Zhang}, L.~{Xiang}, D.~{Nie}, F.~{Yang}, Q.~{Wang},
  and D.~{Shen}, ``Task decomposition and synchronization for semantic
  biomedical image segmentation,'' {\em IEEE Transactions on Image Processing},
  vol.~29, pp.~7497--7510, 2020.

\bibitem{8264783}
M.~{Saha} and C.~{Chakraborty}, ``Her2net: A deep framework for semantic
  segmentation and classification of cell membranes and nuclei in breast cancer
  evaluation,'' {\em IEEE Transactions on Image Processing}, vol.~27, no.~5,
  pp.~2189--2200, 2018.

\bibitem{romera2017erfnet}
E.~Romera, J.~M. Alvarez, L.~M. Bergasa, and R.~Arroyo, ``Erfnet: Efficient
  residual factorized convnet for real-time semantic segmentation,'' {\em IEEE
  Transactions on Intelligent Transportation Systems}, vol.~19, no.~1,
  pp.~263--272, 2017.

\bibitem{long2015fully}
J.~Long, E.~Shelhamer, and T.~Darrell, ``Fully convolutional networks for
  semantic segmentation,'' in {\em Proceedings of the IEEE Conference on
  Computer Vision and Pattern Recognition}, pp.~3431--3440, 2015.

\bibitem{chen2014semantic}
L.-C. Chen, G.~Papandreou, I.~Kokkinos, K.~Murphy, and A.~L. Yuille, ``Semantic
  image segmentation with deep convolutional nets and fully connected crfs,''
  {\em arXiv preprint arXiv:1412.7062}, 2014.

\bibitem{mallat1999wavelet}
S.~Mallat, {\em A wavelet tour of signal processing}.
\newblock Elsevier, 1999.

\bibitem{chen2017deeplab}
L.-C. Chen, G.~Papandreou, I.~Kokkinos, K.~Murphy, and A.~L. Yuille, ``Deeplab:
  Semantic image segmentation with deep convolutional nets, atrous convolution,
  and fully connected crfs,'' {\em IEEE Transactions on Pattern Analysis and
  Machine Intelligence}, vol.~40, no.~4, pp.~834--848, 2017.

\bibitem{chen2017rethinking}
L.-C. Chen, G.~Papandreou, F.~Schroff, and H.~Adam, ``Rethinking atrous
  convolution for semantic image segmentation,'' {\em arXiv preprint
  arXiv:1706.05587}, 2017.

\bibitem{zhao2017pyramid}
H.~Zhao, J.~Shi, X.~Qi, X.~Wang, and J.~Jia, ``Pyramid scene parsing network,''
  in {\em Proceedings of the IEEE Conference on Computer Vision and Pattern
  Recognition}, pp.~2881--2890, 2017.

\bibitem{yang2018denseaspp}
M.~Yang, K.~Yu, C.~Zhang, Z.~Li, and K.~Yang, ``Denseaspp for semantic
  segmentation in street scenes,'' in {\em Proceedings of the IEEE Conference
  on Computer Vision and Pattern Recognition}, pp.~3684--3692, 2018.

\bibitem{paszke2016enet}
A.~Paszke, A.~Chaurasia, S.~Kim, and E.~Culurciello, ``Enet: A deep neural
  network architecture for real-time semantic segmentation,'' {\em arXiv
  preprint arXiv:1606.02147}, 2016.

\bibitem{9040271}
Z.~{Yang}, H.~{Yu}, M.~{Feng}, W.~{Sun}, X.~{Lin}, M.~{Sun}, Z.~H. {Mao}, and
  A.~{Mian}, ``Small object augmentation of urban scenes for real-time semantic
  segmentation,'' {\em IEEE Transactions on Image Processing}, vol.~29,
  pp.~5175--5190, 2020.

\bibitem{zhao2018icnet}
H.~Zhao, X.~Qi, X.~Shen, J.~Shi, and J.~Jia, ``Icnet for real-time semantic
  segmentation on high-resolution images,'' in {\em Proceedings of the European
  Conference on Computer Vision}, pp.~405--420, 2018.

\bibitem{mehta2018espnet}
S.~Mehta, M.~Rastegari, A.~Caspi, L.~Shapiro, and H.~Hajishirzi, ``Espnet:
  Efficient spatial pyramid of dilated convolutions for semantic
  segmentation,'' in {\em Proceedings of the European Conference on Computer
  Vision}, pp.~552--568, 2018.

\bibitem{9032321}
B.~{Jiang}, W.~{Tu}, C.~{Yang}, and J.~{Yuan}, ``Context-integrated and
  feature-refined network for lightweight object parsing,'' {\em IEEE
  Transactions on Image Processing}, vol.~29, pp.~5079--5093, 2020.

\bibitem{li2019dfanet}
H.~Li, P.~Xiong, H.~Fan, and J.~Sun, ``Dfanet: Deep feature aggregation for
  real-time semantic segmentation,'' in {\em Proceedings of the IEEE Conference
  on Computer Vision and Pattern Recognition}, pp.~9522--9531, 2019.

\bibitem{yu2018bisenet}
C.~Yu, J.~Wang, C.~Peng, C.~Gao, G.~Yu, and N.~Sang, ``Bisenet: Bilateral
  segmentation network for real-time semantic segmentation,'' in {\em
  Proceedings of the European Conference on Computer Vision}, pp.~325--341,
  2018.

\bibitem{poudel2019fast}
R.~P. Poudel, S.~Liwicki, and R.~Cipolla, ``Fast-scnn: Fast semantic
  segmentation network,'' {\em arXiv preprint arXiv:1902.04502}, 2019.

\bibitem{orsic2019defense}
M.~Orsic, I.~Kreso, P.~Bevandic, and S.~Segvic, ``In defense of pre-trained
  imagenet architectures for real-time semantic segmentation of road-driving
  images,'' in {\em Proceedings of the IEEE Conference on Computer Vision and
  Pattern Recognition}, pp.~12607--12616, 2019.

\bibitem{hu2020real}
P.~Hu, F.~Perazzi, F.~C. Heilbron, O.~Wang, Z.~Lin, K.~Saenko, and S.~Sclaroff,
  ``Real-time semantic segmentation with fast attention,'' {\em arXiv preprint
  arXiv:2007.03815}, 2020.

\bibitem{li2020semantic}
X.~Li, A.~You, Z.~Zhu, H.~Zhao, M.~Yang, K.~Yang, and Y.~Tong, ``Semantic flow
  for fast and accurate scene parsing,'' {\em arXiv preprint arXiv:2002.10120},
  2020.

\bibitem{yu2020bisenet}
C.~Yu, C.~Gao, J.~Wang, G.~Yu, C.~Shen, and N.~Sang, ``Bisenet v2: Bilateral
  network with guided aggregation for real-time semantic segmentation,'' {\em
  arXiv preprint arXiv:2004.02147}, 2020.

\bibitem{chen2018encoder}
L.-C. Chen, Y.~Zhu, G.~Papandreou, F.~Schroff, and H.~Adam, ``Encoder-decoder
  with atrous separable convolution for semantic image segmentation,'' in {\em
  Proceedings of the European Conference on Computer Vision}, pp.~801--818,
  2018.

\bibitem{sun2019high}
K.~Sun, Y.~Zhao, B.~Jiang, T.~Cheng, B.~Xiao, D.~Liu, Y.~Mu, X.~Wang, W.~Liu,
  and J.~Wang, ``High-resolution representations for labeling pixels and
  regions,'' {\em arXiv preprint arXiv:1904.04514}, 2019.

\bibitem{he2016deep}
K.~He, X.~Zhang, S.~Ren, and J.~Sun, ``Deep residual learning for image
  recognition,'' in {\em Proceedings of the IEEE Conference on Computer Vision
  and Pattern Recognition}, pp.~770--778, 2016.

\bibitem{zeiler2010deconvolutional}
M.~D. Zeiler, D.~Krishnan, G.~W. Taylor, and R.~Fergus, ``Deconvolutional
  networks,'' in {\em 2010 IEEE Computer Society Conference on Computer Vision
  and Pattern Recognition}, pp.~2528--2535.

\bibitem{si2019real}
H.~Si, Z.~Zhang, F.~Lv, G.~Yu, and F.~Lu, ``Real-time semantic segmentation via
  multiply spatial fusion network,'' {\em arXiv preprint arXiv:1911.07217},
  2019.

\bibitem{kumaar2020cabinet}
S.~Kumaar, Y.~Lyu, F.~Nex, and M.~Y. Yang, ``Cabinet: Efficient context
  aggregation network for low-latency semantic segmentation,'' {\em arXiv
  preprint arXiv:2011.00993}, 2020.

\bibitem{howard2019searching}
A.~Howard, M.~Sandler, G.~Chu, L.-C. Chen, B.~Chen, M.~Tan, W.~Wang, Y.~Zhu,
  R.~Pang, V.~Vasudevan, {\em et~al.}, ``Searching for mobilenetv3,'' in {\em
  Proceedings of the IEEE International Conference on Computer Vision},
  pp.~1314--1324, 2019.

\bibitem{howard2017mobilenets}
A.~G. Howard, M.~Zhu, B.~Chen, D.~Kalenichenko, W.~Wang, T.~Weyand,
  M.~Andreetto, and H.~Adam, ``Mobilenets: Efficient convolutional neural
  networks for mobile vision applications,'' {\em arXiv preprint
  arXiv:1704.04861}, 2017.

\bibitem{zhang2018shufflenet}
X.~Zhang, X.~Zhou, M.~Lin, and J.~Sun, ``Shufflenet: An extremely efficient
  convolutional neural network for mobile devices,'' in {\em Proceedings of the
  IEEE Conference on Computer Vision and Pattern Recognition}, pp.~6848--6856,
  2018.

\bibitem{chollet2017xception}
F.~Chollet, ``Xception: Deep learning with depthwise separable convolutions,''
  in {\em Proceedings of the IEEE Conference on Computer Vision and Pattern
  Recognition}, pp.~1251--1258, 2017.

\bibitem{sandler2018mobilenetv2}
M.~Sandler, A.~Howard, M.~Zhu, A.~Zhmoginov, and L.-C. Chen, ``Mobilenetv2:
  Inverted residuals and linear bottlenecks,'' in {\em Proceedings of the IEEE
  Conference on Computer Vision and Pattern Recognition}, pp.~4510--4520, 2018.

\bibitem{fu2019dual}
J.~Fu, J.~Liu, H.~Tian, Y.~Li, Y.~Bao, Z.~Fang, and H.~Lu, ``Dual attention
  network for scene segmentation,'' in {\em Proceedings of the IEEE Conference
  on Computer Vision and Pattern Recognition}, pp.~3146--3154, 2019.

\bibitem{yuan2018ocnet}
Y.~Yuan and J.~Wang, ``Ocnet: Object context network for scene parsing,'' {\em
  arXiv preprint arXiv:1809.00916}, 2018.

\bibitem{huang2019ccnet}
Z.~Huang, X.~Wang, L.~Huang, C.~Huang, Y.~Wei, and W.~Liu, ``Ccnet: Criss-cross
  attention for semantic segmentation,'' in {\em Proceedings of the IEEE
  International Conference on Computer Vision}, pp.~603--612, 2019.

\bibitem{gao2019res2net}
S.~Gao, M.-M. Cheng, K.~Zhao, X.-Y. Zhang, M.-H. Yang, and P.~H. Torr,
  ``Res2net: A new multi-scale backbone architecture,'' {\em IEEE Transactions
  on Pattern Analysis and Machine Intelligence}, 2019.

\bibitem{cordts2016cityscapes}
M.~Cordts, M.~Omran, S.~Ramos, T.~Rehfeld, M.~Enzweiler, R.~Benenson,
  U.~Franke, S.~Roth, and B.~Schiele, ``The cityscapes dataset for semantic
  urban scene understanding,'' in {\em Proceedings of the IEEE Conference on
  Computer Vision and Pattern Recognition}, pp.~3213--3223, 2016.

\bibitem{brostow2009semantic}
G.~J. Brostow, J.~Fauqueur, and R.~Cipolla, ``Semantic object classes in video:
  A high-definition ground truth database,'' {\em Pattern Recognition Letters},
  vol.~30, no.~2, pp.~88--97, 2009.

\bibitem{caesar2018coco}
H.~Caesar, J.~Uijlings, and V.~Ferrari, ``Coco-stuff: Thing and stuff classes
  in context,'' in {\em Proceedings of the IEEE conference on computer vision
  and pattern recognition}, pp.~1209--1218, 2018.

\bibitem{russakovsky2015imagenet}
O.~Russakovsky, J.~Deng, H.~Su, J.~Krause, S.~Satheesh, S.~Ma, Z.~Huang,
  A.~Karpathy, A.~Khosla, M.~Bernstein, {\em et~al.}, ``Imagenet large scale
  visual recognition challenge,'' {\em International Journal of Computer
  Vision}, vol.~115, no.~3, pp.~211--252, 2015.

\bibitem{xie2017aggregated}
S.~Xie, R.~Girshick, P.~Doll{\'a}r, Z.~Tu, and K.~He, ``Aggregated residual
  transformations for deep neural networks,'' in {\em Proceedings of the IEEE
  Conference on Computer Vision and Pattern Recognition}, pp.~1492--1500, 2017.

\bibitem{badrinarayanan2017segnet}
V.~Badrinarayanan, A.~Kendall, and R.~Cipolla, ``Segnet: A deep convolutional
  encoder-decoder architecture for image segmentation,'' {\em IEEE Transactions
  on Pattern Analysis and Machine Intelligence}, vol.~39, no.~12,
  pp.~2481--2495, 2017.

\bibitem{treml2016speeding}
M.~Treml, J.~Arjona-Medina, T.~Unterthiner, R.~Durgesh, F.~Friedmann,
  P.~Schuberth, A.~Mayr, M.~Heusel, M.~Hofmarcher, M.~Widrich, {\em et~al.},
  ``Speeding up semantic segmentation for autonomous driving,'' in {\em MLITS,
  NIPS Workshop}, vol.~2, 2016.

\bibitem{zhang2019customizable}
Y.~Zhang, Z.~Qiu, J.~Liu, T.~Yao, D.~Liu, and T.~Mei, ``Customizable
  architecture search for semantic segmentation,'' in {\em Proceedings of the
  IEEE Conference on Computer Vision and Pattern Recognition},
  pp.~11641--11650, 2019.

\bibitem{lin2020graph}
P.~Lin, P.~Sun, G.~Cheng, S.~Xie, X.~Li, and J.~Shi, ``Graph-guided
  architecture search for real-time semantic segmentation,'' in {\em
  Proceedings of the IEEE/CVF Conference on Computer Vision and Pattern
  Recognition}, pp.~4203--4212, 2020.

\bibitem{9052469}
J.~{Wang}, K.~{Sun}, T.~{Cheng}, B.~{Jiang}, C.~{Deng}, Y.~{Zhao}, D.~{Liu},
  Y.~{Mu}, M.~{Tan}, X.~{Wang}, W.~{Liu}, and B.~{Xiao}, ``Deep high-resolution
  representation learning for visual recognition,'' {\em IEEE Transactions on
  Pattern Analysis and Machine Intelligence}, pp.~1--1, 2020.

\bibitem{shrivastava2016training}
A.~Shrivastava, A.~Gupta, and R.~Girshick, ``Training region-based object
  detectors with online hard example mining,'' in {\em Proceedings of the IEEE
  Conference on Computer Vision and Pattern Recognition}, pp.~761--769, 2016.

\bibitem{chen2019fasterseg}
W.~Chen, X.~Gong, X.~Liu, Q.~Zhang, Y.~Li, and Z.~Wang, ``Fasterseg: Searching
  for faster real-time semantic segmentation,'' {\em arXiv preprint
  arXiv:1912.10917}, 2019.

\bibitem{neuhold2017mapillary}
G.~Neuhold, T.~Ollmann, S.~Rota~Bulo, and P.~Kontschieder, ``The mapillary
  vistas dataset for semantic understanding of street scenes,'' in {\em
  Proceedings of the IEEE international conference on computer vision},
  pp.~4990--4999, 2017.

\bibitem{chandra2018deep}
S.~Chandra, C.~Couprie, and I.~Kokkinos, ``Deep spatio-temporal random fields
  for efficient video segmentation,'' in {\em Proceedings of the IEEE
  Conference on Computer Vision and Pattern Recognition}, pp.~8915--8924, 2018.

\bibitem{9133304}
Z.~{Huang}, X.~{Wang}, Y.~{Wei}, L.~{Huang}, H.~{Shi}, W.~{Liu}, and T.~S.
  {Huang}, ``Ccnet: Criss-cross attention for semantic segmentation,'' {\em
  IEEE Transactions on Pattern Analysis and Machine Intelligence}, pp.~1--1,
  2020.

\bibitem{zhang2017scale}
R.~Zhang, S.~Tang, Y.~Zhang, J.~Li, and S.~Yan, ``Scale-adaptive convolutions
  for scene parsing,'' in {\em Proceedings of the IEEE International Conference
  on Computer Vision}, pp.~2031--2039, 2017.

\bibitem{kong2018recurrent}
S.~Kong and C.~C. Fowlkes, ``Recurrent scene parsing with perspective
  understanding in the loop,'' in {\em Proceedings of the IEEE Conference on
  Computer Vision and Pattern Recognition}, pp.~956--965, 2018.

\bibitem{wu2019wider}
Z.~Wu, C.~Shen, and A.~Van Den~Hengel, ``Wider or deeper: Revisiting the resnet
  model for visual recognition,'' {\em Pattern Recognition}, vol.~90,
  pp.~119--133, 2019.

\bibitem{yu2018learning}
C.~Yu, J.~Wang, C.~Peng, C.~Gao, G.~Yu, and N.~Sang, ``Learning a
  discriminative feature network for semantic segmentation,'' in {\em
  Proceedings of the IEEE Conference on Computer Vision and Pattern
  Recognition}, pp.~1857--1866, 2018.

\bibitem{zhao2018psanet}
H.~Zhao, Y.~Zhang, S.~Liu, J.~Shi, C.~Change~Loy, D.~Lin, and J.~Jia, ``Psanet:
  Point-wise spatial attention network for scene parsing,'' in {\em Proceedings
  of the European Conference on Computer Vision (ECCV)}, pp.~267--283, 2018.

\bibitem{yuan2019object}
Y.~Yuan, X.~Chen, and J.~Wang, ``Object-contextual representations for semantic
  segmentation,'' {\em arXiv preprint arXiv:1909.11065}, 2019.

\end{thebibliography}

\end{document}